\pdfoutput=1

\documentclass[11pt]{article}

\usepackage[final]{acl}

\usepackage{times}
\usepackage{latexsym}

\usepackage[T1]{fontenc}

\usepackage[utf8]{inputenc}

\usepackage{microtype}

\usepackage{inconsolata}

\usepackage{graphicx}

\usepackage{multirow}

\usepackage[utf8]{inputenc} 
\usepackage[T1]{fontenc}    
\usepackage{hyperref}       
\usepackage{url}            
\usepackage{booktabs}       
\usepackage{amsfonts}       
\usepackage{nicefrac}       
\usepackage{microtype}      

\usepackage{graphicx}
\usepackage{enumitem}
\usepackage{wrapfig, blindtext, booktabs}
\usepackage{longtable}
\usepackage{tabularx}
\usepackage{listings}
\usepackage[ruled]{algorithm2e}
\usepackage{algpseudocode}
\usepackage{amsmath}
\usepackage{float}
\usepackage{color,xcolor,colortbl}
\usepackage{multirow}
\usepackage{longtable}

\title{MedQA-CS: Objective Structured Clinical Examination (OSCE)-Style Benchmark for Evaluating LLM Clinical Skills}

\author{Zonghai Yao $^1$,
Zihao Zhang $^{2}$,
Chaolong Tang $^1$, 
Xingyu Bian $^1$\\ 
\bf{Youxia Zhao} $^1$,
\bf{Zhichao Yang} $^1$,
\bf{Junda Wang} $^1$,
\bf{Huixue Zhou} $^3$\\
\bf{Won Seok Jang} $^4$,
\bf{Feiyun Ouyang} $^4$,
\bf{Hong Yu}  $^{1, 4, 5}$\\
University of Massachusetts, Amherst$^1$, Emory University$^2$, University of Minnesota$^3$\\
University of Massachusetts, Lowell$^4$,
UMass Chan Medical School$^5$\\
{\{zonghaiyao, chaolongtang, xingyubian, youxiazhao\}@umass.edu}
}

\begin{document}
\maketitle
\begin{abstract}
    Artificial intelligence (AI) and large language models (LLMs) in healthcare require advanced clinical skills (CS), yet current benchmarks fail to evaluate these comprehensively. 
We introduce MedQA-CS, an AI-SCE framework inspired by medical education's Objective Structured Clinical Examinations (OSCEs), to address this gap. 
MedQA-CS evaluates LLMs through two instruction-following tasks—LLM-as-medical-student and LLM-as-CS-examiner—designed to reflect real clinical scenarios. 
Our contributions include developing MedQA-CS, a comprehensive evaluation framework with publicly available data and expert annotations, and providing the quantitative and qualitative assessment of LLMs as reliable judges in CS evaluation. 
Our experiments show that MedQA-CS is a more challenging benchmark for evaluating clinical skills than traditional multiple-choice QA benchmarks (e.g., MedQA). 
Combined with existing benchmarks, MedQA-CS enables a more comprehensive evaluation of LLMs' clinical capabilities for both open- and closed-source LLMs
~\footnote{Our data, prompts, codes, and annotations are public at GitHub ~\url{https://github.com/bio-nlp/MedQA-CS} and Hugging Face Hub \url{https://huggingface.co/datasets/bio-nlp-umass/MedQA-CS-Student} and \url{https://huggingface.co/datasets/bio-nlp-umass/MedQA-CS-Exam} with CC-BY-NC-4.0 License. }
.
\end{abstract}

\section{Introduction}
\label{sec:intro}
Artificial intelligence (AI) and large language models (LLMs) are increasingly adopted in healthcare, resulting in many clinical NLP applications that require expert-level clinical skills such as diagnosis and clinical documentation~\cite{achiam2023gpt,mcduff2023towards,tu2024towards,yang2024advancing}.
Current clinical LLM benchmarks, such as MMLU-Med~\cite{hendrycks2020measuring}, MedQA-US~\cite{jin2021disease}, MedMCQA~\cite{pal2022medmcqa}, PubMedQA~\cite{jin2019pubmedqa}, JAMA Clinical Challenge, Medbullets~\cite{chen2024benchmarking}, and AMBOSS~\cite{gilson2023does}, mainly measure clinical knowledge through multiple-choice questions (MCQ). 

\begin{figure}
  \begin{center}
    \includegraphics[width=\linewidth]{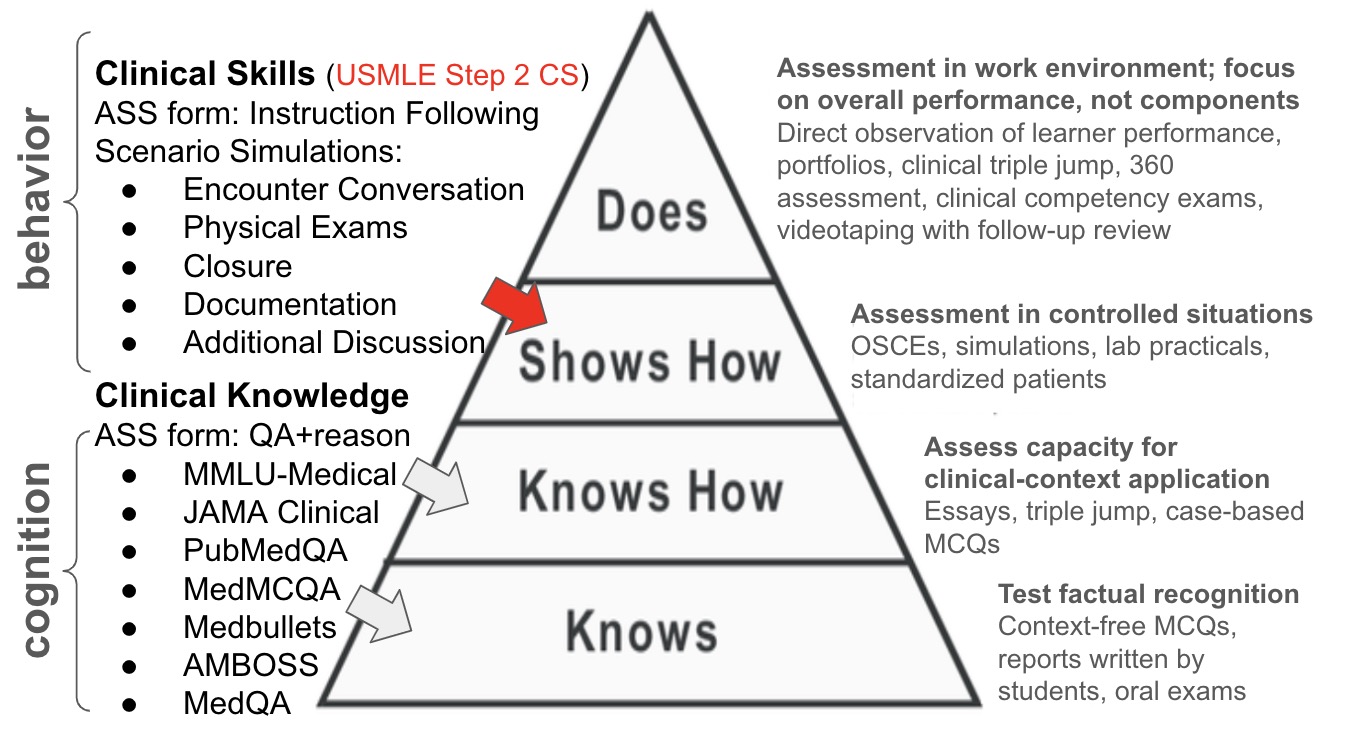}
  \end{center}
  \caption{Miller's pyramid of clinical competence matched with an appropriate level of assessment. Figure adapted from \cite{miller1990assessment}.
  }
  \label{fig:miller_pyramid_assessment}
\end{figure}

\begin{figure*}[!ht]
    \centering
\includegraphics[width=\textwidth]{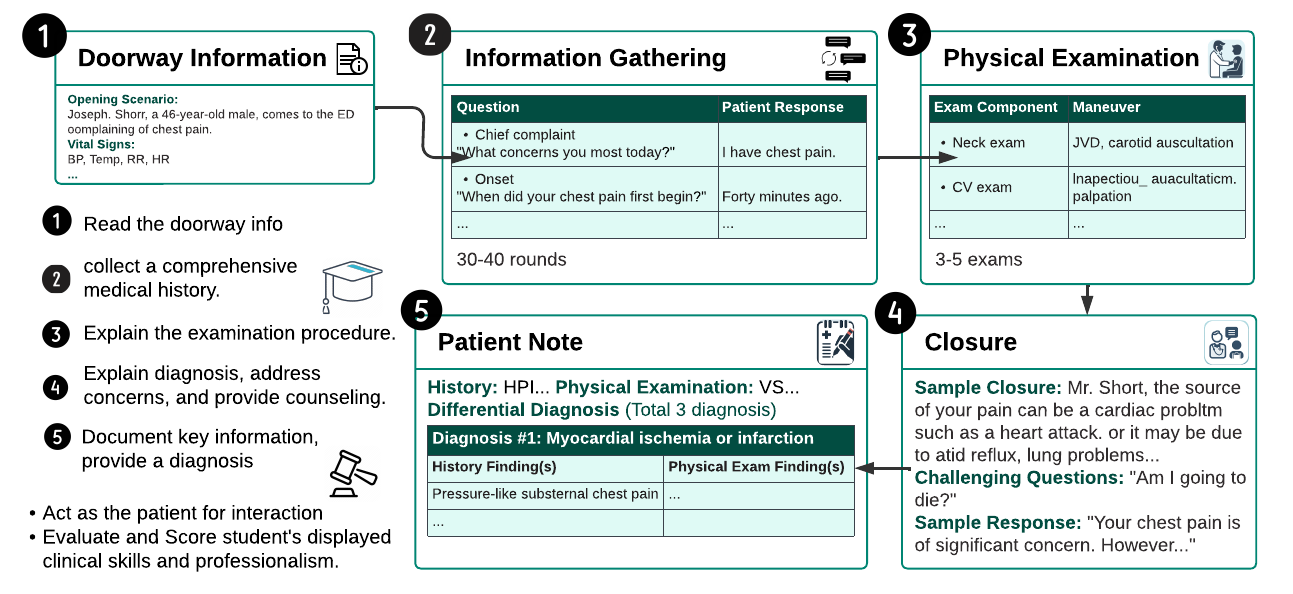}
    \caption{Overview of the United States Medical Licensing Examination (USMLE) Step2 Clinical Skills (CS).
    The medical student begins by reviewing the doorway information (Phase \textcircled{\textbf{1}}),
    then gathers the patient's history \textcircled{\textbf{2}}, 
    performs a physical examination \textcircled{\textbf{3}},
    concludes with the closure phase \textcircled{\textbf{4}},
    and documents the encounter in a patient note with a differential diagnosis \textcircled{\textbf{5}}. 
    Throughout these phases, the \textbf{Clinical Skills Examiner} plays the role of the patient, interacting with the \textbf{Medical Student} to simulate a real clinical encounter and assess their clinical skills. 
    The examiner provides feedback and scores the student’s performance based on predefined criteria. This OSCE structured approach ensures a comprehensive assessment of the student's ability to conduct patient encounters effectively and professionally.
    Our main objective is to transform this OSCE into an AISCE for LLM Clinical Skills benchmarking. 
    Therefore, throughout the process, there will be tasks for both \textbf{MedStuLLM} (LLM-as-student) and \textbf{MedExamLLM} (LLM-as-examiner) that the LLM needs to complete. 
    The goal for MedStuLLM is to achieve a better AI-SCE score to demonstrate its clinical skills, while the goal for MedExamLLM is to have a high correlation with the expert examiner’s scoring results to prove its capability as a judge in the clinical domain.
    More details about USMLE STEP2 CS can be found in appendix~\ref{apx:sec:overview_usmle} and one example in appendix~\ref{apx:sec:example_usmle}.
    }
    \label{fig:OSCE_Flowchart}
\end{figure*}

However, identifying robust clinical guidelines and what constitutes a successful interaction for healthcare LLMs will be crucial towards fulfilling the long-term goals of patients, providers, and other clinical stakeholders~\cite{mehandru2024evaluating}.
In medical education, there has been a shift from assessing students using standardized testing, which evaluates clinical knowledge through MCQs, to modern curricula, which increasingly use Objective Structured Clinical Examination (OSCE)~\cite{zayyan2011objective,harden2015definitive}.
As shown in Figure~\ref{fig:miller_pyramid_assessment}, Miller's Pyramid~\cite{miller1990assessment} provides a comprehensive framework for evaluating the competence of medical students, from knowledge acquisition to real-world performance~\cite{norcini2003work,albino2008assessing}.
Early medical exams have typically evaluated students on the "knows" and "knows how" levels of Miller's Pyramid, while OSCEs primarily evaluate students' practical skills (e.g., the "shows how" level) in clinical settings, including patient examination, clinical history recording, effective communication, and handling unexpected situations. 
As depicted in Figure~\ref{fig:OSCE_Flowchart}, the USMLE Step2-CS exam~\cite{scott2019testing} exemplifies this comprehensive approach by emphasizing real-time interaction with standardized patients and assessing candidates' abilities in these critical areas. 
Similar to the transition in medical education, there is a growing call~\cite{mehandru2024evaluating} to develop AI-Structured Clinical Examinations (AI-SCEs) to assess LLMs' ability to assist in real-world clinical workflows.

Previous MCQ benchmarks have notable shortcomings: 
1) 
MCQ benchmarks primarily focus
on the "knows" and "knows how" levels of Miller's Pyramid, neglecting the practical skills essential in medical education. 
2) The MCQ format limits LLMs to making choices rather than engaging in open-ended queries, failing to capture the nuanced abilities required in real-world clinical encounters, such as patient information gathering.
3) Despite achieving performance levels comparable to or exceeding those of experts in previous MCQ benchmarks, LLMs' scores on MedQA-CS are significantly lower, highlighting the discrepancy between clinical knowledge-based assessments and our practical clinical skills-based assessments.

To address these issues, we propose the MedStuLLM (as shown and defined in Figure~\ref{fig:OSCE_Flowchart}) evaluation framework in MedQA-CS, which incorporates instruction-following to evaluate LLMs at the "shows how" level of Miller's Pyramid.
Unlike previous benchmarks, our framework meticulously follows the USMLE Step 2-CS guidelines to assess four critical aspects of clinical encounters: information gathering, physical examination, closure, and diagnosis. This comprehensive approach ensures that LLMs provide interpretability in their instruction-following process rather than just a final score, offering a more fine-grained evaluation.
Our findings indicate that performance on knowledge-based benchmarks does not equate to clinical skills performance, underscoring the necessity for a specially designed AI-SCE benchmark.

For evaluation metric design, transforming the OSCE into AI-SCE involves creating a subjective yet professional automated evaluation metric to replace human clinical skills examiners. 
OSCEs typically consist of long lists of processes or diagnoses students are graded on. 
Although there are standard answers, the output need not match the reference exactly as long as it is reasonable. 
This subjectivity makes traditional generative metrics like ROUGE~\cite{lin2004rouge}, BERTSCORE~\cite{zhang2019bertscore}, and some medical-concepts-based metrics~\cite{abacha2023investigation}, which are often based on exact word or term matching and do not account for semantic meaning, less effective for AI-SCE benchmarks.
Recent research on LLM-as-Judge has shown a higher correlation with human evaluations in such scenarios~\cite{chiang2023can,kocmi2023large,zheng2024judging,zhang2024comprehensive,kim2023prometheus,kim2024prometheus,lan2024criticbench}.
However, such metrics are not well-studied in clinical or other similar expert-domain NLP tasks. 
Major concerns arise from uncertainties about the ability of LLMs to possess sufficient expertise~\cite{li2024leveraging}, and the need for complex prompt engineering and pipeline design, which should be created by interdisciplinary teams of clinicians, computer scientists, and medical researchers~\cite{mehandru2024evaluating}.
Therefore, carefully designed prompts for MedExamLLM are essential to ensure consistency with expert evaluations, making them reliable automated metrics for MedQA-CS during clinical skills benchmarking.

In summary, our MedQA-CS is meticulously designed with two components: MedStuLLM and MedExamLLM to assess LLMs' ability to assist in real-world clinical workflows and the reliability of LLM-as-Judge as an automated evaluation metric in expert domains. Our main contributions are summarized as follows:

\begin{itemize}
[leftmargin=.2in,topsep=2pt]
\setlength\itemsep{0.01em}
\vspace{-0.2em}
    \item We developed MedQA-CS, an AI-SCE evaluation framework inspired by OSCEs used in medical education to assess LLM clinical skills at the "shows how" level of Miller's Pyramid. 
    We benchmarked a wide range of mainstream open- and closed-source LLMs, demonstrating the necessity and challenge of the MedQA-CS benchmark for current state-of-the-art LLMs. 
    \item To our knowledge, we are the first to quantitatively and qualitatively assess the reliability of the LLM-as-Judge framework in complex clinical NLP scenarios that require subjective yet professional evaluations.
    Our experimental results demonstrate that well-designed prompts aligned with OSCE guidelines and detailed expert-designed evaluation criteria can achieve high CS evaluation agreement between LLMs and human experts, showcasing the significant potential of LLM-as-Judge framework in future clinical skill or relevant downstream tasks assessments.
\end{itemize}

\section{MedQA-Clinical Skills Benchmark}
\label{sec:method}

The official USMLE website\footnote{\url{https://www.usmle.org/about-usmle}} provides publicly available study materials for the USMLE exam, including Step 1~\footnote{\url{https://www.usmle.org/sites/default/files/2021-10/Step_1_Sample_Items.pdf}}, Step 2 CK~\footnote{\url{https://www.usmle.org/sites/default/files/2021-10/Step2_CK_Sample_Questions.pdf}}, and Step 3~\footnote{\url{https://www.usmle.org/sites/default/files/2021-10/Step3_Sample_Items.pdf}}, which are all in the multiple-choice format and have been integrated into the MedQA dataset~\cite{jin2021disease}.
For the Step 2 CS component, we sourced guidelines and clinical case materials from a publicly accessible online resource~\footnote{\url{https://www.doc88.com/p-21161955650573.html}}, containing 44 structured clinical encounter cases. We manually transcribed the original content and processed it into (instruction, input, output) triples following the format described in Appendix~\ref{apx:sec:method}.

To ensure full compliance with U.S. copyright law and best practices for dataset creation, we adopted a two-step, transformative pipeline that combines automated paraphrasing with rigorous expert review (see Appendix~\ref{apx:sec:data_processing_details} for all details). Briefly, a portion of the original content was systematically paraphrased using ChatGPT at the sentence and dialogue levels, and all transformed cases were then independently verified by licensed medical experts for semantic fidelity and clinical accuracy. The final dataset, MedQA-CS, comprises 1,667 instruction-based data points spanning information gathering, physical exams, closures, and differential diagnoses. All data are released under a CC BY-NC 4.0 license for non-commercial research and educational use.

While our dataset consists of 44 carefully curated USMLE Step 2 CS cases, resulting in 1,667 data points, this scale is consistent with real-world clinical assessment standards: the USMLE Step 2 CS exam determines pass/fail based on only 12 cases per student. We acknowledge that further expansion to additional specialties and modalities is important future work.

In the rest of this section, we introduce each part of MedQA-CS and our detailed designs for MedStuLLM and MedExamLLM, including examples (instruction, input, output) for two components. Due to space limitations, additional design and implementation details for each section are provided in Appendix~\ref{apx:sec:method}. At the end of this section, we also validate the data quality of MedQA-CS through human evaluation.
    
    \subsection{MedStuLLM and MedExamLLM}
    \label{sec:method_section_description}
    
        \subsubsection{InfoGatherQA}
        \label{sec:method_InfoGatherQA}
        The InfoGatherQA section simulates patient encounters, requiring the MedStuLLM to ask focused questions based on initial doorway information and prior conversation history to \textbf{gather relevant details about the patient's condition}~\cite{peterson1992contributions,miller1990assessment}. 
As the example shown in Table~\ref{tab:InfoGatherQA_workflow}, given the doorway information of ``Joseph Shorr, a 46-year-old male with chest pain and vital signs (BP: 165/85 mm Hg, Temp: 98.6°F, RR: 22/minute, HR: 90/minute, regular)'',
the MedStuLLM might ask, ``Can you describe where exactly you feel the pain and whether it spreads to any other parts of your body?''
Each generated question is assessed independently to ensure it aligns with the ground truth questions~\cite{amboss2018step2cs}.

The MedExamLLM evaluates these questions based on two criteria: \textbf{consistency} and \textbf{originality}. Consistency ensures the question aligns with the ground truth question bank, while originality checks that the question is novel and not repetitive within the same conversation. 
For example, the MedExamLLM might determine that the above MedStuLLM-generated question is correct because it matches ground truth questions in the question bank.
Each evaluation is done in a structured manner, with the examiner's output in JSON format containing a reasoning section and a score (0 or 1) for each question, ensuring a detailed and transparent evaluation process.  
In the InfoGatherQA section, the performance score for the MedStuLLM reflects the proportion of questions that meet both criteria, indicating the LLM's proficiency in information gathering.
The MedExamLLM's score is used to measure how closely the evaluations align with expert assessments. 
These scores are essential for benchmarking different LLMs clinical skills.

        \subsubsection{Physical Exams}
        \label{sec:method_PhysicalExams}
        The Physical Exams section of the MedQA-CS benchmark assesses the ability of the MedStuLLM to \textbf{document and justify physical examinations during a patient encounter}.
After completing the initial patient interaction, the MedStuLLM is required to write down a detailed physical exam based on the doorway information and the chat history~\cite{zayyan2011objective,harden2015misconceptions}.
As the example shown in Table~\ref{tab:Physical_Exam_workflow}, if a patient presents with chest pain, the MedStuLLM might document ``Heart: Auscultation for possible abnormal heart sounds and rhythm such as murmurs, gallop sound, or arrhythmias'' 
and explain that ``The patient is presenting with chest pain, shortness of breath, and sweating which can indicate a cardiac issue like angina or a heart attack.''
This documentation is grounded in the patient’s symptoms and medical history~\cite{rao2025large}.
The MedExamLLM in this section evaluates these documented examinations by comparing them to a ground truth answer using a specified rubric. This evaluation includes three main criteria: \textbf{Exam Coverage}, \textbf{Reason Relevance and Accuracy}, and \textbf{Extra Exams Penalty}. 
For example, if the MedStuLLM documented heart and chest exams but missed necessary components like the neck or abdominal exams, the MedExamLLM would score the response lower for Exam Coverage.
The MedExamLLM output is detailed in JSON format, and the final overall score range is from 0 to 100.
Scores for the MedStuLLM reflect its ability to cover necessary examinations, provide accurate and relevant justifications, and avoid unnecessary tests.
The MedExamLLM's score indicates how closely its evaluations align with those of human experts.

        \subsubsection{Closure}
        \label{sec:method_Closure}
        
The Closure section evaluates the MedStuLLM's ability to \textbf{effectively conclude patient encounters}. 
This involves \textbf{summarizing} the patient’s chief complaint, history of present illness (HPI), and findings from physical examinations, as well as \textbf{outlining the next steps} in management and \textbf{addressing any challenging questions the patient may have}~\cite{swygert2009relationship}. 
As the example shown in Table~\ref{tab:Closure_workflow}, after assessing a patient like Mr. Shorr with severe chest pain, the MedStuLLM might summarize: 
``Mr. Shorr, based on your recent episode of severe chest pain that woke you up from sleep, and that it radiated to your neck, upper back, and left arm, alongside your reported hypertension, high cholesterol, and past GERD, coupled with your physical exam that did not show any major abnormal findings, we are tentatively considering a few possibilities.''
The MedExamLLM evaluates these closure summaries generated by the MedStuLLM based on a rubric adapted from the USMLE guidelines. 
This rubric includes five main criteria: \textbf{Diagnostic Impressions}, \textbf{Management Plans}, \textbf{Challenging Questions}, \textbf{Language Accessibility}, and \textbf{Compassion}. 
For instance, if MedStuLLM misses key components or uses inaccessible medical jargon, it would receive lower scores for certain criteria (e.g., Diagnostic Impressions) with an explanation like, ``The student's diagnostic impressions align with those in the sample, including the possibility of a heart condition or GERD. However, the student omitted other potential causes such as lung problems or disorders related to the large blood vessels in the chest.''
In addition to summarizing the encounter, the MedStuLLM must address any challenging questions the patient poses. 
For instance, in response to a patient asking, ``Is it a heart attack? Am I going to die?''
The MedStuLLM might answer: ``I understand that you're scared. Right now, we suspect a heart condition given your symptoms, but further tests like an EKG and blood work will help us determine the exact cause.''

The MedExamLLM evaluates responses based on two key criteria: adequacy, ensuring alignment with the ground truth answers, and compassion, assessing the level of empathy demonstrated. This approach reflects the latter three core elements of the SPIKES model, a widely adopted framework in the medical field for delivering bad news~\cite{choe2019communicating}. These elements include providing Knowledge and information to the patient (K), addressing the patient’s Emotions with empathic responses (E), and applying Strategies and Summary (S) to support patient understanding and emotional well-being.
The MedExamLLM output is detailed in JSON format, and the final overall score range is from 0 to 100.
The MedStuLLM's scores reflect its proficiency in summarizing patient encounters, ensuring clarity and completeness in communication, and demonstrating empathy. 
The MedExamLLM's score indicates the accuracy and reliability of its evaluations compared to expert assessments.

        \subsubsection{Differential Diagnosis}
        \label{sec:method_Diagnosis}
        The Differential Diagnosis section in the MedQA-CS benchmark assesses the MedStuLLM's ability to \textbf{formulate and justify potential diagnoses} based on information gathered during the patient encounter. 
This involves synthesizing all previous information from the InfoGatherQA stage, physical examinations, and initial patient notes to propose a list of possible medical conditions. 
As the example shown in Table~\ref{tab:Diagnosis_workflow}, given a patient like Joseph Shorr with chest pain and relevant clinical data, the MedStuLLM must identify differential diagnoses, providing historical and physical findings that support each diagnosis~\cite{olson2020improving}. 
Each diagnosis is evaluated for accuracy and supported by evidence from the patient's history and examination. 
For example, the MedStuLLM might output: ``Diagnosis: Acute Coronary Syndrome. Historical Finding(s): Substernal chest pain radiating to the left arm, upper back, and neck. Associated symptoms of nausea, sweating, and dyspnea. History of hypertension and high cholesterol.''
The MedExamLLM evaluates the differential diagnoses proposed by the MedStuLLM using a detailed rubric. 
The evaluation focuses on three main criteria: \textbf{correctness of the diagnosis name}, \textbf{relevance of historical findings}, and \textbf{relevance of physical findings}. 
The evaluation process involves comparing the generated diagnoses with correct answers and additional potential diagnoses, ensuring a thorough assessment.
Scores for the MedStuLLM in this section reflect the accuracy and completeness of the differential diagnoses. 
The total score is calculated by summing the points for each diagnosis, historical finding, and physical finding, then dividing by the maximum possible points to yield a final score between 0 and 1. 
The MedExamLLM's scores measure the alignment of its evaluations with expert assessments, ensuring the reliability of the evaluation process. 
    
    \subsection{Quality Evaluation}
    \label{sec:method_data_quality}

\begin{table*}[!ht]
\centering
\begin{tabular}{c|cccc}
\hline
\text{Pearson ($p$)} & \text{InfoGatherQA} & \text{Physical Exam} & \text{Closure} & \text{Diagnosis} \\
\hline
\text{E1 vs. 2} & 0.89 (<0.001) & 0.94(<0.001) & 0.89(<0.005) & 0.95(<0.001) \\
\text{E1 vs. 3} & 0.88 (<0.001) & 0.99(<0.001) & 0.87(<0.005) & 0.92(<0.001) \\
\text{E2 vs. 3} & 0.77 (<0.01) & 0.99(<0.001) & 0.86(<0.005) & 0.88(<0.001) \\
\hline
\text{K Tau ($p$)} & \text{InfoGatherQA} & \text{Physical Exam} & \text{Closure} & \text{Diagnosis} \\
\hline
\text{E1 vs. 2} & 0.73 (<0.005) & 0.55(<0.05) & 0.65(<0.05) & 0.73(<0.005) \\
\text{E1 vs. 3} & 0.78 (<0.001) & 0.90(<0.001) & 0.68(<0.05) & 0.82(<0.001) \\
\text{E2 vs. 3} & 0.60 (<0.05) & 0.54(<0.05) & 0.58(<0.05) & 0.73(<0.005) \\
\hline
K W ($p$) & 0.89 (<0.005) & 0.78 (<0.05) & 0.84 (<0.01) & 0.91(<0.005)\\
\hline
\end{tabular}
\caption{Pearson's r, Kendall's $\tau$ and W for three different experts evaluation agreement using MedExamLLM evaluation guideline. We used MedStuLLM (GPT-4) output for experts evaluation.}
\label{tab:human_corr}
\end{table*}

The reliability of the MedQA-CS design was evaluated through the agreement among three experts who assessed the MedStuLLM (GPT-4) results across four sections. 
Detailed information about the human annotation guidelines derived from MedExamLLM prompts, as well as the recruitment and guidance of domain experts for the evaluation, is provided in Appendix \ref{apx:sec:human_annotation_guidelines}. 
Our goal was to validate the MedQA-CS MedStuLLM and MedExamLLM design from the perspective of domain experts. 
If experts can follow each requirement of MedExamLLM to evaluate MedStuLLM's output and achieve highly consistent results, it confirms the soundness of our MedQA-CS design details.
This evaluation employed Pearson's r and Kendall's $\tau$ to measure correlation and consistency between the expert pairs~\cite{shen-etal-2023-large}. 
Pearson's r values ranged from 0.77 to 0.99, indicating strong to very strong correlations in all sections~\cite{moslem2019analysing}, with highly significant p-values (most p < 0.001). 
Kendall's $\tau$ values, ranging from 0.54 to 0.90, further support the consistency of the experts' evaluations. 
The Kendall's W values, representing the overall agreement among the three experts, were all significant, ranging from 0.78 to 0.91 (with p-values < 0.05), indicating substantial agreement. 
The high correlations and consistent evaluations across different sections demonstrate that the experts' assessments of the MedStuLLM outputs are highly reliable, confirming the effectiveness of the MedExamLLM design in providing consistent and accurate evaluations. 
Fine-grained expert evaluation results can be found in appendix~\ref{apx:sec:human_annotation_results}.

\section{Experimental Settings}
\label{sec:experimental_settings}
We focus on the following two research questions (RQ): 
\textbf{RQ1:} Assessing the reliability of LLMs as judges in the MedQA-CS context (for MedExamLLM). This involves benchmarking various LLMs' MedExamLLM capabilities and evaluating AI-expert agreements when reviewing MedStuLLM (GPT-4) results. 
\textbf{RQ2:} Utilizing the most reliable MedExamLLM as an automatic metric to benchmark the clinical skills of various LLMs in critical instruction-following tasks across different sections (for MedStuLLM).

The LLMs includeding the experiments are the GPT series (GPT-3.5-turbo, GPT-4-turbo, GPT-4o)~\cite{achiam2023gpt}, the Claude-3 series (Claude-3-haiku, Claude-3-sonnet, Claude-3-opus, Claude-3.5-sonnet)~\cite{anthropic2024claude}, and some representative open-source general LLMs (LLAMA2~\cite{touvron2023llama}, LLAMA3~\cite{meta2024introducing}, Mistral\&Mixtral~\cite{jiang2024mixtral}, GLM-4~\cite{zeng2022glm}, and Qwen2~\cite{bai2023qwen}). 
In RQ1 settings, we use all default parameters in their official API with temperature=0 for GPT and Claude 3.
For traditional metrics, we use
ROUGE~\cite{lin2004rouge},
METEOR~\cite{banerjee2005meteor},
BERTScore~\cite{zhang2019bertscore},
Exact String Match with lowercase,
and UMLS-F1~\footnote{The UMLS-F1 score evaluates how well medical terms extracted from an LLM-generated response align with medical terms extracted from the reference text, where the terms are identified using Scispacy~\cite{neumann-etal-2019-scispacy} (using \textit{en\_core\_sci\_lg} NER model) and linked to UMLS biomedical concepts~\cite{bodenreider2004unified}.}.
In RQ2 settings, we use all default parameters in their official API with temperature=0.9 for GPT and Claude 3.
For open-source LLMs, we use HuggingFacePipeline (text-generation~\footnote{\url{https://huggingface.co/blog/langchain}}) with parameters: max\_new\_tokens = 2000, top\_k = 50, do\_sample = True, temperature = 0.1, return\_full\_text=False.

More details about experimental settings are detailed in appendix~\ref{apx:sec:experimental_settings}.

\section{Results and Discussion}
\label{sec:experiments}
    
    \subsection{RQ-MedExamLLM: Reliability of LLM-as-Judge in MedQA-CS}
    \label{sec:result_MedExamLLM}

\begin{table*}
\centering
\begin{tabular}{l|cccc}
\hline
\text{Pearson | K $\tau$} & \text{InfoGatherQA} & \text{Physical Exam} & \text{Closure} & \text{Diagnosis} \\
\hline
\text{GPT-4o} & 0.82 | 0.64 & 0.80 | 0.38 & \textcolor{red}{0.76} | 0.37 & 0.71 | 0.56 \\
\textcolor{red}{GPT-4} & \textcolor{red}{0.90} | \textcolor{red}{0.78} & \textcolor{red}{0.92} | \textcolor{red}{0.53} & 0.47 | \textcolor{red}{0.47} & \textcolor{red}{0.78} | \textcolor{red}{0.69} \\
\text{GPT-3.5} & -0.25 | -0.07 & -0.14 | -0.56 & 0.25 | 0.13 & -0.05 | 0.11 \\
\text{Claude3-Opus} & 0.78 | 0.63 & 0.82 | 0.35 & 0.75 | 0.25 & 0.64 | 0.56  \\
\text{Claude3-Sonnet} & 0.52 | 0.33 & 0.75 | 0.40 & -0.09 | -0.12 & 0.41 | 0.29  \\
\text{Claude3-haiku} & 0.05 | 0.05 & 0.36 | 0.12 & -0.02 | 0.23 & 0.43 | 0.29  \\
\hline
\text{ROUGE-1} & 0.67 | 0.56 & 0.52 | 0.18 & 0.16 | 0.14 & 0.02 | -0.07 \\
\text{ROUGE-2} & 0.70 | \textcolor{red}{0.60} & 0.33 | 0.38 & 0.04 | 0.07 & 0.17 | \textcolor{red}{0.24} \\
\text{ROUGE-L} & 0.65 | \textcolor{red}{0.60} & 0.45 | 0.07 & 0.28 | 0.35 & -0.02 | -0.02 \\
\text{METEOR} & 0.62 | 0.47 & \textcolor{red}{0.72} | \textcolor{red}{0.46} & -0.07 | -0.35 & 0.05 | 0.07 \\
\text{BERTScore} & \textcolor{red}{0.86} | 0.56 & 0.28 | 0.44 & 0.23 | 0.05 & 0.03 | 0.02 \\
\text{Exact Match} & - & 0.35 | 0.25 & - & 0.19 | 0.20 \\
\text{UMLS-F} & 0.65 | 0.47 & 0.63 | 0.28 & \textcolor{red}{0.35} | \textcolor{red}{0.54} & \textcolor{red}{0.25} | 0.11 \\
\hline
\end{tabular}
\caption{Pearson correlation and Kendall's Tau between expert evaluation (average) and 1. different LLMs' MedExamLLM output (LLM-as-Judge) 2. some traditional metrics used in clinical generation tasks.
We put the full table with p-value in appendix~\ref{apx:sec:results}. 
We finally chose MedExamLLM (GPT-4) for MedQA-CS clinical skills benchmarking because it best aligns with expert evaluation. "Red" numbers are "best".}
\label{tab:human_ai_corr}
\end{table*}

This RQ serves two primary purposes: 
1) to use the MedExamLLM dataset with experts' evaluation results to benchmark LLMs' ability as LLM-as-Judge in clinical skills-related tasks, and 
2) to assess the reliability of using the best MedExamLLM as the evaluation metric in the follow-up MedStuLLM by examining the correlation between LLMs' assessments and expert evaluations.
Specifically, the alignment of LLMs' outputs with expert evaluations was measured using Pearson r. and Kendall's $\tau$. 
As shown in Table~\ref{tab:human_ai_corr}, GPT-4 exhibited the highest reliability, with Pearson r. of 0.90, 0.92, and 0.78, and Kendall's $\tau$ values of 0.78, 0.53, and 0.69 for the Information Gathering, Physical Exam, and Diagnosis, respectively. 
For the Closure section, MedExamLLM correlations were relatively lower (0.47 for both Pearson and Kendall's $\tau$). 
Discussions with experts suggest that, unlike the other three sections, which have more definitive correct answers (e.g., question banks, target physical exams, diagnosis), the content generation and evaluation criteria for Closure are more subjective. 
This indicates that LLMs need improvement in handling patient-oriented information summarization and communication.
GPT-4o and Claude-3-Opus also demonstrated strong correlations with expert evaluations. 
In contrast, weaker LLMs showed significantly poorer results.
Traditional automatic metrics often showed low correlations in the Diagnosis or Closure sections, indicating poor agreement with human experts. 
While ROUGE, METEOR, BERTScore, and UMLS-F performed much better in the other two sections, they still lagged behind the best-performing MedExamLLM. 
Considering cost factors, certain sections may use traditional metrics instead of best-performing MedExamLLM as evaluation metrics, such as BERTScore for InfoGatherQA. 
Overall, carefully chosen traditional metrics are better choices than weak LLM-as-judge in InfoGatherQA and Physical Exam, but MedExamLLM (GPT4) remains the most reliable option. 
While not all LLMs are reliable for clinical skills evaluation, stronger models like GPT-4, GPT-4o, and Claude-3-Opus show substantial promise by providing evaluations that closely mirror expert assessments. 
Consequently, GPT-4 was selected as the most reliable MedExamLLM for benchmarking clinical skills in MedStuLLM.
Error analysis for detailed MedExamLLM performance can be found in appendix~\ref{apx:sec:case_study}.

However, current options for answer scoring rely on proprietary language models, which pose challenges for future cost and reproducibility.
To address this, we distilled a LLaMA-based judge model using GPT-4 as the reference.
Specifically, we used GPT-4-judge to generate synthetic scores from 10 randomly selected cases out of 44, covering different LLM-student responses.
We then fine-tuned the LLaMA 3.1-8B model on this data and tested it on 10 expert-evaluated cases.
Our results showed that LLaMA-judge achieved high correlations with expert evaluations in InfoGatherQA (0.9289) and Physical Exam (0.93), matching GPT-4-judge.
For Closure, LLaMA-judge performed moderately well, slightly exceeding 0.5, which is comparable to GPT-4-judge’s 0.47.
However, it struggled in the Diagnosis section, consistently scoring below 0.3 in correlation with experts—far behind GPT-4-judge’s 0.78.
We hypothesize that this is due to the complexity of diagnosis, which exceeds what an 8B model can capture through simple distillation.
Future work is needed to enhance the performance of open-source medical judge models, particularly for diagnosis evaluation.

    \subsection{RQ-MedStuLLM: Benchmarking Clinical Skills Ability of LLMs}
    \label{sec:result_MedStuLLM}

\begin{table*}[!t]
\centering
\resizebox{0.7\linewidth}{!}{
\begin{tabular}{r|cccc|c}
\hline
MedStuLLM & \footnotesize{InfoGatherQA} & \footnotesize{Physical Exam} & \footnotesize{Closure} & \footnotesize{Diagnosis} & Avg. \\
\hline
GPT-4o & 62.12 & 52.08 & 78.45 & 55.05 & 61.93 \\
GPT-4 & 62.79 & 48.97 & 77.21 & 50.58 & 59.89 \\
GPT-3.5 & 39.11 & 43.34 & 66.52 & 44.78 & 48.44 \\
Claude3-Opus & 61.28 & 50.34 & 83.26 & 53.68 & 62.14 \\
Claude3-Sonnet & 46.66 & 52.82 & 77.88 & 51.28 & 57.16 \\
Claude3-haiku & 33.47 & 50.86 & 77.40 & 51.04 & 53.19 \\
Claude3.5-Sonnet & 72.04 & 48.95 & 77.55 & 50.84 & 62.35 \\
\hline
Qwen2-72b & 43.07 & 51.7 & 85.77 & 47.15 & 56.92 \\
Qwen2-moe-57b & 46.09 & 45.3 & 81.57 & 46.05 & 54.75 \\
Qwen2-7b & 17.95 & 44.3 & 72.13 & 37.01 & 42.85 \\
Qwen2-1.5b & - & -16.4 & 14.44 & 25.71 & - \\
Qwen2-0.5b & - & -7.4 & - & 11.32 & - \\
\hline
GLM4-9b	& 22.95 & 59.90 & 76.67 & 40.58 & 50.02 \\
\hline
LLAMA3-8b & 21.16 & - & - & 37.75 & - \\
+SimPO & 33.59 & 31.7 & 67.2 & 39.31 & 42.94 \\
+DPO & 17.39 & 37.6 & 60.03 & - & - \\
+IPO & 17.66 & 27.6 & 69.0 & 44.25 & 39.63 \\
+KTO & 12.09 & 16.5 & 72.0 & 47.79 & 37.09 \\
+RDPO & 15.5 & 39.2 & 73.0 & - & - \\
+ORPO & 6.01 & 25.7 & 40.5 & 41.77 & 28.48 \\
OpenBioLLM-8b & 10.56 & - & - & 39.80 & - \\
\hline
Mistral-7b & 23.47 & 49.70 & 78.30 & 38.17 & 47.41 \\
+SimPO & 32.83 & 42.9 & 76.8 & 39.77 & 48.08 \\
+DPO & 26.76 & 42.7 & 77.4 & 45.82 & 48.17 \\
+IPO & 29.59 & 36.3 & 72.3 & 41.52 & 44.92 \\
+KTO & 35.04 & 40.2 & 80.1 & 37.99 & 48.33 \\
+RDPO & 35.96 & 41.9 & 72.2 & 39.96 & 47.51 \\
+ORPO & 21.00 & 27.4 & 64.5 & 45.42 & 39.58 \\
BioMistral-7b & 15.33 & 14.20 & - & 42.04 & - \\
\hline
LLAMA2-70b & 12.05 & - & - & 32.61 & - \\
LLAMA3-70b & 37.86 & - & - & 41.6 & - \\
OpenBioLLM-70B & 24.40 & 39.4 & - & 35.35 & - \\
Mixtral-8x7b & 29.80 & - & - & 42.67 & - \\
\hline
\end{tabular}
}
\caption{Benchmarking results. '-' means that LLM cannot follow instruction to generate valid output. We used MedExamLLM (GPT-4) as metric to evalute different LLMs' output for MedStuLLM tasks.}
\label{tab:cs_student_benchmark}
\end{table*}

As shown in Table~\ref{tab:cs_student_benchmark}, the MedStuLLM average scores are significantly lower than those of previous clinical knowledge-focused benchmarks (e.g., MedQA, LLMs with 90+ scores surpassed human expert level).
This disparity underscores the complexity and unique challenges of CS instruction-following ability for LLMs, highlighting the need for enhanced training strategies to improve LLMs' proficiency in this domain.
Specifically, the state-of-the-art LLMs evaluated in this study achieved avg. scores ranging from 48.44 (GPT-3.5) to 62.35 (Claude-3.5-Sonnet), indicating a substantial opportunity for improvement in their ability to follow complex clinical skill instructions. 

Regarding open-source LLMs, the scaling law of LLM clinical skills ability can be observed in the results of QWen2. The performance of QWen2-72B is comparable to that of some closed-source LLMs.
However, our findings also reveal that open-source LLMs struggle significantly with following complex CS instructions in MedStuLLM to generate valid outputs.
Even 70B version models fail to follow the instructions of the physical exam and closure sections.
This difficulty suggests that future research should focus on developing effective training methodologies to enhance the performance of open-source LLMs in these challenging tasks, thereby promoting their potential to serve as AI agents~\cite{li2024agent,park2023generative} capable of interacting with humans or other agents in clinical settings.
We then explored two potential directions for improvement: domain adaptation training and human preferences alignment. 
Our findings indicate that current domain adaptation training strategy~\cite{OpenBioLLMs,labrak2024biomistral,tran2023bioinstruct}, which has been successful for previous clinical knowledge benchmarks (e.g., MedQA), negatively impacts the LLMs clinical skills instruction-following ability. 
This adverse effect is likely due to catastrophic forgetting, where enhancing domain knowledge leads to losing previously learned abilities to follow clinical instructions. 
This phenomenon aligns with recent studies in the field~\cite{luo2023empirical,ren2024analyzing,Chang2024HowDL}~\footnote{\url{https://ai.meta.com/blog/adapting-large-language-models-llms/}}. 
In contrast, current human preference alignment training, such as DPO~\cite{rafailov2024direct} and its variants~\cite{meng2024simpo,ethayarajh2024kto,azar2024general,hong2024reference,park2024disentangling}, show improving results. 
While these approaches do not try to enhance domain-specific knowledge, they improve the LLMs' ability to follow complex CS instructions that were previously unmanageable, even in the absence of specific adaptations for clinical instructions.
This aligns with recent findings that RLHF helps LLMs generalize more effectively to new inputs, especially when there is a significant distribution shift during inference time~\cite{kirk2023understanding}, such as in our case of complex clinical instruction following.
These observations highlight the necessity of a combined advanced training strategy that integrates both domain knowledge enhancement and complex instruction-following capability~\cite{cheng2023adapting}. 
Future work should continue to refine these strategies to unlock LLMs' full potential in clinical applications.
The case study can be found in appendix~\ref{apx:sec:case_study}.

\section{Related Work}
\label{sec:related_work}
The rapid advancement of AI has enabled LLMs to tackle complex healthcare tasks, including disease diagnosis and medical consultations~\cite{mcduff2023towards, achiam2023gpt}. These models encode clinical knowledge, retrieve relevant medical texts, and perform accurate Q\&A during consultations or discharge~\cite{singhal2023large, hernandez2023we, zakka2024almanac, xiong2024benchmarking, wang2024jmlr, li2024benchmarking}. LLMs have surpassed the average human score on USMLE, achieving up to 91.1\%~\cite{saab2024capabilities}, yet existing evaluations fail to capture real-world clinical performance\cite{mehandru2024evaluating}.
Recently, several medical QA benchmarks have been proposed, including MedHELM~\cite{bedi2025medhelm}, CliMedBench~\cite{ouyang-etal-2024-climedbench}, and MedBench~\cite{cai2024medbench}.
MedHELM and MedBench focus on knowledge recall and diagnostic reasoning through MCQs or single-turn QA, while CliMedBench emphasizes the structure of clinical reasoning.
However, none of these benchmarks simulate OSCE-style workflows or comprehensively evaluate clinical communication skills, gaps that MedQA-CS is specifically designed to address.

Recent research has explored using well-trained language models for natural language generation evaluation across various domains, including but not limited to law~\cite{cui2023chatlaw}, medicine~\cite{singhal2023large}, and finance~\cite{yang2023fingpt}, employing both reference-based and reference-free methods~\cite{bai2024benchmarking,alpaca_eval}.
LLMs have demonstrated a high correlation with human evaluations in generation tasks by scoring candidate texts or comparing two candidates based on specified evaluation aspects~\cite{fabbri2021summeval,chen2023storyer,chiang2023can,kocmi2023large,zheng2024judging,zhang2024comprehensive,kim2023prometheus,kim2024prometheus,lan2024criticbench}.
However, most LLM-as-Judge work is done primarily in general NLP fields. 
There is no previous work that claims the usefulness of LLM-as-Judge in clinical NLP, mainly because the generalizability of LLMs as evaluation tools in specialized fields faces significant challenges due to the lack of domain-specific knowledge and the need for evaluation prompts designed to meet specific domain standards~\cite{li2024leveraging}. 
Our research advances this discussion by exploring the feasibility of LLM-as-CS-Examiner. 

\section{Conclusion}
\label{sec:conclusion}

MedQA-CS offers a novel AI-SCE framework, emphasizing the critical need for clinical skills benchmarks and showcasing the potential of LLMs as reliable CS judges in relevant NLP tasks. This framework introduces a more rigorous evaluation approach compared to traditional benchmarks, ensuring a more accurate assessment of LLMs' clinical capabilities. By integrating real clinical scenarios and expert annotations, MedQA-CS provides a comprehensive and publicly accessible tool for advancing AI-based evaluations in healthcare.

\section{Limitations}
\label{sec:limits}
This study has several limitations.

First, the sample size, derived from the USMLE Step 2 CS, is relatively small and may not comprehensively represent the full spectrum of clinical medicine disciplines or clinical skills, thus limiting the generalizability of our findings. Notably, our dataset predominantly covers common conditions and lacks adequate representation of certain specialties such as pediatrics and psychiatry, as well as rare diseases. Future studies should involve larger and more diverse datasets that encompass a broader range of medical domains, conditions, and specialties to validate and extend these results.

Second, the LLM-as-Judge framework in this paper did not consider the reasoning process of MedStuLLM during evaluation. This is because we found it challenging to produce stable and reliable scores for MedStuLLM’s reasoning without ground truth reasoning. To assess the reliability of MedStuLLM’s reasoning, we conducted a qualitative analysis through human evaluation. Specifically, we randomly selected 25 samples from four sections and had two medical experts independently annotate the reasoning as either Correct or Incorrect. The inter-annotator agreement was 85.1\%, indicating a high level of consistency between experts. Additionally, the agreement between MedStuLLM’s reasoning and the corresponding response, as measured in our previous human evaluation, was 81\%. Overall, the quality of MedStuLLM’s reasoning and response was consistent. In future work, we plan to explore reference-free clinical reasoning evaluation.

Third, it is important to note that clinical skills typically encompass treatment plan formulation. However, due to limitations in the original USMLE Step 2 CS dataset (as illustrated in Figure~\ref{fig:OSCE_Flowchart}), the USMLE only evaluates medical students' clinical skills up to the diagnostic part of clinical note generation, without extending to treatment plans for each diagnosis. This is why our benchmark does not include this aspect. In the future, we aim to explore how to gather suitable treatment plan data from other sources to integrate into MedQA-CS.

Fourth, all MedQA-CS data are presented exclusively in English, which limits the applicability of our findings in non-English-speaking contexts. In addition, the dataset is constrained to a single modality, using only text-based inputs and outputs. Multimodal clinical data—such as speech (for doctor–patient communication analysis), images (e.g., radiology), or physiological waveforms—are not represented. Future work should investigate the inclusion of multimodal data, such as speech, images, or other clinically relevant signals, to better reflect the complexities of real-world clinical interactions.

Moreover, this research exclusively addresses tasks related to medical visits, such as information gathering, question answering, physical examination recommendations, closure, and differential diagnosis. The extension of our findings to other domains and tasks remains unexplored, indicating that further validation and adjustments will be necessary before applying this approach to different fields.

Finally, although we employed three medical experts for human evaluation, increasing the number and diversity of qualified domain experts would improve the statistical significance and robustness of our findings. Additionally, future work should consider issues of fairness, generalizability, and potential biases inherent in LLMs, particularly when expanding the evaluation across specialties, modalities, and patient populations.

\section{Ethical Considerations}
\label{sec:ethical}
\paragraph{Societal Impacts}
The MedQA-CS benchmark is designed to advance the evaluation of large language models (LLMs) in realistic clinical skill scenarios. However, the scores reported do not suggest that LLMs possess the clinical competency required to replace physicians, medical students, or licensed examiners. Although certain models demonstrate promising performance on specific benchmark sections, their real-world clinical skills—including diagnostic reasoning, patient interaction, and context-aware judgment—remain unproven and outside the scope of this work.

The benchmark focuses on the instruction-following and text generation capabilities of LLMs, not on medical decision-making or actual patient care. Thus, LLMs should be viewed as supportive tools to assist healthcare professionals rather than as substitutes. The safe integration of LLMs into medical workflows must be pursued with caution, emphasizing close collaboration with clinicians, rigorous validation, and continuous human oversight. Responsible and ethical development is essential to maximize positive societal impact and avoid overreliance or misuse of AI in healthcare.

\paragraph{Data Sourcing, Copyright, and Legal Compliance}
All clinical case materials used in MedQA-CS were obtained from publicly accessible, non-commercial sources, in particular, USMLE Step 2 CS sample cases available online. To fully respect intellectual property rights and comply with U.S. copyright law, each case was subjected to substantial transformation via automated paraphrasing (using ChatGPT) and expert validation, as described in Section 2. The resulting content is strictly intended for non-commercial, educational, and research purposes, and is released under a CC BY-NC 4.0 license.

The paraphrasing process involved extensive changes at the sentence, dialogue, and contextual levels—far exceeding simple synonym replacement or minor edits. The structure and style of conversations were reworked to support multi-turn, instruction-following evaluation, with deliberate attention to clinical communication flow. Every transformed case underwent review by qualified medical experts, who verified both semantic fidelity and clinical accuracy, ensuring that no factual errors or inappropriate transformations were introduced.

This approach represents a significant extension beyond prior work such as~\cite{jin2021disease}, which primarily focused on reformatting multiple-choice questions (e.g., shuffling answer options). In contrast, our process entails deep, context-aware re-authoring of open-ended, multi-turn interactions, resulting in cases that are both substantially different in form and function from the original texts and suitable for advanced benchmarking.

\paragraph{Ethical Use, Transparency, and Reproducibility}
All data, prompts, and evaluation artifacts associated with MedQA-CS are released for research and educational purposes only, with no commercial applications permitted. The entire data processing and transformation pipeline—including all paraphrasing and validation steps—is thoroughly documented and open to scrutiny, supporting transparency and reproducibility for the research community.

\paragraph{Human Subject Involvement and Privacy}
No identifiable patient data, personal health information (PHI), or real-world medical records were used in constructing MedQA-CS. All expert annotation, verification, and scoring were performed by qualified clinicians or domain experts who participated with informed consent. The benchmark does not involve any real patient interaction or intervention, thereby posing no risk to patient privacy or safety.

\paragraph{Commitment to Responsible Research}
We recognize the profound responsibility associated with developing and deploying LLMs in clinical contexts. The MedQA-CS benchmark is intended as a resource for advancing research in clinical AI evaluation—not as a diagnostic tool or clinical decision support system. We strongly recommend that future development, deployment, and evaluation of AI systems in healthcare continue to adhere to the highest ethical, legal, and professional standards.

\bibliography{custom}

\newpage

\appendix

\onecolumn
\newpage

\section{Overview of the USMLE STEP2 Clinical Skills OSCE}
\label{apx:sec:overview_usmle}
\subsection{Phase 1: Doorway Information (10-20 seconds)}

\textbf{Medical Student:}
\begin{itemize}
    \item Read the doorway information, noting vital signs, age, and chief complaint.
    \item Identify the encounter objectives: history and physical exam versus just history.
    \item Develop a list of likely differential diagnoses.
\end{itemize}

\textbf{Examiner:}
\begin{itemize}
    \item Ensure the doorway information is accurate and complete.
\end{itemize}

\subsection{Phase 2: History Taking (7-8 minutes)}

\textbf{Medical Student:}
\begin{itemize}
    \item Greet the patient, shake hands, and introduce yourself.
    \item Make eye contact and maintain proper posture.
    \item Cover the patient with a drape to ensure modesty.
    \item Start with an open-ended question to elicit the chief complaint.
    \item Avoid using technical terms and show empathy.
    \item Do not interrupt or rush the patient.
    \item Obtain past medical, surgical, family, social, and sexual history, including current medications and allergies.
\end{itemize}

\textbf{Examiner:}
\begin{itemize}
    \item Act as the patient, responding to questions appropriately.
    \item Evaluate the medical student’s ability to gather a complete and accurate history while demonstrating empathy and professionalism.
\end{itemize}

\subsection{Phase 3: Physical Examination (3-5 minutes)}

\textbf{Medical Student:}
\begin{itemize}
    \item Explain each step of the physical exam to the patient.
    \item Ask for permission before starting the examination.
    \item Use respectful draping techniques and never examine through the gown.
\end{itemize}

\textbf{Examiner:}
\begin{itemize}
    \item Act as the patient, providing appropriate physical findings.
    \item Assess the medical student’s technique and thoroughness in performing the physical examination.
\end{itemize}

\subsection{Phase 4: Closure (2-3 minutes)}

\textbf{Medical Student:}
\begin{itemize}
    \item Explain possible diagnoses and required workups to the patient.
    \item Avoid using complicated medical terminology.
    \item Ask if the patient has any concerns and address them.
    \item Be prepared to handle challenging questions.
    \item Provide counseling and say goodbye, thanking the patient.
\end{itemize}

\textbf{Examiner:}
\begin{itemize}
    \item Act as the patient, asking challenging questions if necessary.
    \item Evaluate the medical student’s communication skills, ability to explain medical information clearly, and overall closure of the encounter.
\end{itemize}

\subsection{Phase 5: Patient Note (10 minutes)}

\textbf{Medical Student:}
\begin{itemize}
    \item Document key elements, including chief complaint, history of present illness, review of systems, past medical history, social history, and family history.
    \item Record key physical findings and pertinent positives and negatives.
    \item Provide up to three differential diagnoses with supporting history and physical findings.
    \item Suggest up to eight diagnostic tests.
\end{itemize}

\textbf{Examiner:}
\begin{itemize}
    \item Review the patient note for completeness, accuracy, and coherence.
    \item Score the note based on the quality of documentation and the rationale for differential diagnoses and suggested tests.
\end{itemize}

\subsection{Final: Evaluation}

\textbf{Medical Student:}
\begin{itemize}
    \item Reflect on the encounter, noting areas of strength and areas for improvement.
\end{itemize}

\textbf{Examiner:}
\begin{itemize}
    \item Provide feedback based on the performance during the interaction.
    \item Score each phase based on predefined criteria, focusing on the medical student’s clinical skills, professionalism, and communication abilities.
\end{itemize}

\section{Example of in USML-step2-CS}
\label{apx:sec:example_usmle}
\subsection*{Opening Scenario}
Joseph Short, a 46-year-old male, comes to the ED complaining of chest pain.

\subsection*{Vital Signs}
\begin{itemize}
    \item BP: 165/85 mm Hg
    \item Temp: 98.6°F (37°C)
    \item RR: 22/minute
    \item HR: 90/minute, regular
\end{itemize}

\subsection*{Examinee Tasks}
\begin{enumerate}
    \item Take a focused history.
    \item Perform a focused physical exam (do not perform rectal, genitourinary, or female breast exam).
    \item Explain your clinical impression and workup plan to the patient.
    \item Write the patient note after leaving the room.
\end{enumerate}

\subsection*{Checklist/SP Sheet}

\subsection*{Patient Description}
Patient is a 46 yo M.

\subsection*{Notes for the SP}
\begin{itemize}
    \item Lie on the bed and exhibit pain.
    \item Place your hands in the middle of your chest.
    \item Exhibit difficulty breathing.
    \item If ECG is mentioned by the examinee, ask, “What is an ECG?”
\end{itemize}

\subsection*{Challenging Questions to Ask}
“Is this a heart attack? Am I going to die?”

\subsection*{Sample Examinee Response}
“Your chest pain is of significant concern. However, chest pain can be caused by a large variety of issues. We need to learn more about what’s going on to know if your pain is life threatening.”

\subsection*{Examinee Checklist}
\subsection*{Building the Doctor-Patient Relationship}
\begin{itemize}
    \item Examinee knocked on the door before entering.
    \item Examinee made eye contact with the SP.
    \item Examinee correctly used patient’s name.
    \item Examinee introduced self by name.
    \item Examinee identified his/her role or position.
\end{itemize}

\subsection*{Reflective Listening}
\begin{itemize}
    \item Examinee asked an open-ended question and actively listened to the response.
    \item Examinee asked the SP to list his/her concerns and listened to the response without interrupting.
    \item Examinee summarized the SP’s concerns, often using the SP’s own words.
\end{itemize}

\section*{Information Gathering}
\begin{itemize}
    \item Examinee elicited data efficiently and accurately.
\end{itemize}

\subsection*{Question and Patient Response}
\begin{longtable}{|p{3cm}|p{10cm}|}
\hline
\textbf{Question} & \textbf{Patient Response} \\
\hline
Chief complaint & “What concerns you most today?” \\
& \textbf{I have chest pain.} \\
\hline
Onset & “When did your chest pain first begin?” \\
& \textbf{Forty minutes ago.} \\
\hline
Precipitating events & “Were you doing anything in particular when the pain began?” \\
& \textbf{Nothing; I was asleep and woke up at 5:00 in the morning having this pain.} \\
\hline
Progression & “Has the pain become more or less intense since it first began?” \\
& \textbf{Constant severity.} \\
\hline
Severity on a scale & “On a scale of 0 to 10, with 1 being almost no pain and 10 being the worst pain of your life, what rating would you give your chest pain right now?” \\
& \textbf{7/10.} \\
\hline
Location & “Please show me exactly where you feel the pain.” \\
& \textbf{Middle of the chest. It feels as if it’s right underneath the bone.} \\
\hline
Radiation & “Does the pain move anywhere?” \\
& \textbf{To my neck, upper back, and left arm.} \\
\hline
Quality & “What does the pain feel like?” \\
& \textbf{Pressure. Like something sitting on my chest.} \\
\hline
Alleviating/exacerbating factors & “Does anything make the pain better or worse?” \\
& \textbf{Nothing.} \\
\hline
Shortness of breath & “Have you had any difficulty breathing?” \\
& \textbf{Yes.} \\
\hline
Nausea/vomiting & “Have you had any nausea or vomiting?” \\
& \textbf{I feel nauseated, but I didn’t vomit.} \\
\hline
Sweating & “Have you noticed any increased sweating?” \\
& \textbf{Yes.} \\
\hline
Associated symptoms & “Have you noticed any other symptoms? Cough? Wheezing? Stomach pain?” \\
& \textbf{None.} \\
\hline
Previous episodes of similar pain & “Has anything like this ever happened to you before?” \\
& \textbf{Yes, but not exactly the same.} \\
& “What makes this episode different from previous episodes?” \\
& \textbf{The pain is much worse this time and feels more like pressure than burning.} \\
\hline
Onset of previous episode & “When did you first experience this kind of chest pain?” \\
& \textbf{The past 3 months.} \\
\hline
Severity & “How intense was the pain at that time?” \\
& \textbf{Less severe.} \\
\hline
Frequency & “Since that first episode, how frequently would you experience chest pain?” \\
& \textbf{I have had two to three episodes a week, each lasting 5 to 10 minutes.} \\
\hline
Precipitating events & “Do you associate any events or activities with the onset of the pain?” \\
& \textbf{Walking up the stairs, strenuous work, and heavy meals.} \\
\hline
Alleviating factors & “Has anything helped to relieve your chest pain in the past?” \\
& \textbf{Antacids.} \\
\hline
Associated symptoms & “Did you have any other symptoms with those prior episodes of chest pain?” \\
& \textbf{None.} \\
\hline
Past medical history & “What medical problems do you have?” \\
& \textbf{Hypertension for 5 years, treated with a diuretic. High cholesterol, managed with diet; I have not been very compliant with the diet. GERD 10 years ago, treated with antacids.} \\
\hline
Current medications & “What medications do you currently take?” \\
& \textbf{Maalox, diuretic.} \\
\hline
Past surgical history & “Have you ever undergone surgery?” \\
& \textbf{None.} \\
\hline
Family history & “Has anyone in your family been diagnosed with heart disease or suffered sudden cardiac death or stroke?” \\
& \textbf{My father died of lung cancer at age 72. My mother is alive and has a peptic ulcer. No early heart attacks.} \\
\hline
Occupation & “What do you do for a living?” \\
& \textbf{Accountant.} \\
\hline
Alcohol use & “Do you drink alcohol?” \\
& \textbf{Once in a while.} \\
\hline
Illicit drug use & “Do you ever use any recreational drugs?” \\
& \textbf{Cocaine, once a week.} \\
\hline
Duration of cocaine use & “For how long have you been using cocaine?” \\
& \textbf{Ten years.} \\
\hline
Last time of cocaine use & “When was the last time you used cocaine?” \\
& \textbf{Yesterday afternoon.} \\
\hline
Tobacco & “Do you smoke cigarettes or use tobacco?” \\
& \textbf{Stopped 3 months ago.} \\
\hline
Duration & “How long have you been smoking cigarettes?” \\
& \textbf{Twenty-five years.} \\
\hline
Amount & “How many packs of cigarettes do you smoke per day?” \\
& \textbf{One pack a day.} \\
\hline
Sexual activity & “Are you sexually active?” \\
& \textbf{Well, doctor, to be honest, I haven’t had sex with my wife for the past 3 months because I get this pain in my chest during sex.} \\
\hline
Exercise & “Do you exercise regularly?” \\
& \textbf{No.} \\
\hline
Diet & “How would you describe your diet?” \\
& \textbf{My doctor gave me a strict diet last year to lower my cholesterol, but I always cheat.} \\
\hline
Drug allergies & “Are you allergic to any medications?” \\
& \textbf{No.} \\
\hline
\end{longtable}

\subsection*{Connecting With the Patient}
\begin{itemize}
    \item Examinee recognized the SP’s emotions and responded with PEARLS.
\end{itemize}

\section*{Physical Examination}
\begin{itemize}
    \item Examinee washed his/her hands.
    \item Examinee asked permission to start the exam.
    \item Examinee used respectful draping.
    \item Examinee did not repeat painful maneuvers.
\end{itemize}

\begin{table}[h!]
\begin{tabular}{|l|p{10cm}|}
\hline
\textbf{Exam Component} & \textbf{Maneuver} \\ \hline
Neck exam               & JVD, carotid auscultation          \\ \hline
CV exam                 & Inspection, auscultation, palpation \\ \hline
Pulmonary exam          & Auscultation, palpation, percussion \\ \hline
Abdominal exam          & Auscultation, palpation, percussion \\ \hline
Extremities             & Checked peripheral pulses, checked blood pressure in both arms, looked for edema and cyanosis \\ \hline
\end{tabular}
\end{table}

\section*{Closure}
\begin{itemize}
    \item Examinee discussed initial diagnostic impressions.
    \item Examinee discussed initial management plans:
    \begin{itemize}
        \item Follow-up tests.
        \item Lifestyle modification (diet, exercise).
    \end{itemize}
    \item Examinee asked if the SP had any other questions or concerns.
\end{itemize}

\subsection*{Sample Closure}
Mr. Short, the source of your pain can be a cardiac problem such as a heart attack, or it may be due to acid reflux, lung problems, or disorders related to the large blood vessels in your chest. It is crucial that we perform some tests to identify the source of your problem. We will start with an ECG and some blood work, but more complex tests may be needed as well. In the meantime, I strongly recommend that you stop using cocaine, since use of this drug can lead to a variety of medical problems, including heart attacks. I commend you for quitting smoking and encourage you to continue not to smoke as cigarettes are known to worsen cardiovascular disease and increase your risk of developing a heart attack in the future. Do you have any questions for me?

\section*{Patient Note}

\subsection*{History}
HPI: 46 yo M complains of substernal chest pain. The pain started 40 minutes before the patient presented to the ED. The pain woke the patient from sleep at 5:00 AM with a steady 7/10 pressure sensation in the middle of his chest that radiated to the left arm, upper back, and neck. Nothing makes it worse or better. Nausea, sweating, and dyspnea are also present. Similar episodes have occurred during the past 3 months, 2-3 times/week. These episodes were precipitated by walking up the stairs, strenuous work, sexual intercourse, and heavy meals. Pain during these episodes was less severe, more burning in quality, lasted for 5-10 minutes, and disappeared spontaneously or after taking antacids.

ROS: Negative except as above.

Allergies: NKDA.

Medications: Maalox, diuretic.

PMH: HTN, HLD, GERD.

SH: 1 PPD for 25 years; stopped 3 months ago. Occasional EtOH, occasional cocaine for 10 years (last used yesterday afternoon). No regular exercise; poorly adherent to diet.

FH: Noncontributory.

\subsection*{Physical Examination}
Patient is in severe pain.

VS: BP 165/85 mm Hg (both arms), RR 22/minute.

Neck: No JVD, no bruits.

Chest: No tenderness, clear symmetric breath sounds bilaterally.

Heart: Apical impulse not displaced; RRR; S1/S2 WNL; no murmurs, rubs, or gallops.

Abdomen: Soft, nondistended, nontender, \(\oplus\) BS x 4 quadrants, no hepatosplenomegaly.

Extremities: No peripheral edema, peripheral pulses 2+ and symmetric in upper and lower extremities, no cyanosis or clubbing.

\subsection*{Differential Diagnosis}
\begin{itemize}
    \item \textbf{Diagnosis \#1: Myocardial ischemia or infarction}
    \begin{itemize}
        \item \textit{History Finding(s):} Pressure-like substernal chest pain, Pain radiates to left arm, upper back, and neck, Pain awakens patient at night
        \item \textit{Physical Exam Finding(s):} \textbf{None specific}
    \end{itemize}
    \item \textbf{Diagnosis \#2: Cocaine-induced myocardial ischemia}
    \begin{itemize}
        \item \textit{History Finding(s):} History of cocaine use, Last used yesterday afternoon, Pressure-like substernal chest pain
        \item \textit{Physical Exam Finding(s):} \textbf{None specific}
    \end{itemize}
    \item \textbf{Diagnosis \#3: GERD}
    \begin{itemize}
        \item \textit{History Finding(s):} Pain in midchest, Previous pain was relieved by antacids, Previous pain occurred after heavy meals
        \item \textit{Physical Exam Finding(s):} \textbf{None specific}
    \end{itemize}
\end{itemize}

\subsection*{Diagnostic Workup}
\begin{itemize}
    \item ECG
    \item Cardiac enzymes (CPK, CPK-MB, troponins)
    \item Transthoracic echocardiography
    \item Upper endoscopy
    \item Urine toxicology
    \item Chest x-ray
\end{itemize}

\small
\newpage
\section{Prompts}
\label{prompts}

Table 8: All prompts used in our proposed MedStuLLM (LLM-as-medical-student) USMLE Step 2 Clinical Skills test pipeline.

\lstset{
    basicstyle=\ttfamily,
    columns=fullflexible,
    breaklines=true,
    postbreak=\mbox{\textcolor{red}{$\hookrightarrow$}\space}
}
{\tiny
\centering

\begin{tabular}{|p{1cm}|p{14cm}|}
\hline
\textbf{Type}  & \textbf{Prompt} \\ \hline
InfoGatherQA & 
\begin{lstlisting}
    You are a doctor and your job is to  take a detailed focused history is essential to the formulation of a 
    differential diagnosis and workup plan later. You should begin by reading the history information and previous dialogue 
    below carefully. Then decide which question to ask in order to get more detailed information of patient's concern and give the
    reason why you choose this question. 
    Your question should include the following aspects if you think it is necessary: Chief complaint, Onset, Precipitating 
    events, Progression,Severity on a scale, Location, Radiation, Quality, Alleviating/ exacerbating factors, Shortness of breath,
    Nausea/vomiting, Sweating, Previous episodes of similar pain, Severity, Frequency, Precipitating events,
    Alleviating factors, Associated symptoms, Past medical history, Current medications, Past surgical history,
    Family history, Occupation, Alcohol use, Illicit drug use, Duration of cocaine use, Last time of cocaine use,
    Tobacco, Duration, Amount, Sexual activity, Exercise, Diet, Drug allergies

    history:
    {opening}
   
    Previous dialogue:
    {chat_history}
  
    You should only ask one question at a time!!! 
    Your output should in the json format:
    {{"symptom": "The symptom you want to ask", "reason": "the reason why you choose this question", "question": "the question you ask, you should ask one question at a time"}}
\end{lstlisting}
\\ \hline
Physical Exams & 
\begin{lstlisting}
   You are a doctor and want to figure out what the physical exam you have to do to physical test the patient's symptoms
    appear in the opening and previous dialogue with patient. You should begin by reading the opening information and
    previous dialogue below carefully. Then decide which physical exam and corresponding maneuver should take: 
    Physical exams should be choose from the following component: Neck, Chest, Heart, Abdomen, Extremities, VS, CV, Pulmonary,
    Back, Hips, Neuro, HEENT, Skin, 
    You should only choose the physical exams that mentioned in the opening and previous dialogue and also output the reason
    why you choose this physical exam
    opening:
    {opening}
   
    Previous dialogue:
    {chat_history}
    

    Your output should in the following format, you may output one or more physical exams: 
    {{"exam1": {{"physical exam": "the exam you choose", "maneuver": "the maneuver corresponding to exam", "reason": "the reason you choose this exam"}},"exam2": {{"physical exam": "the exam you choose", "maneuver": "the maneuver corresponding to exam", "reason": "the reason you choose this exam"}}, ...}}

    Output:
\end{lstlisting}
\\ \hline
Closure & 
\begin{lstlisting}
   You are a medical student who is taking the USMLE. You have already taken the first two parts patient encounter and physical exam. 
    The current section is called closure. There are two parts of this section, writing the closure and answer the question from the patient.
    For both parts, you should response based on previous information and show comfort to the patient or In the first part, your job is to write a brief closure to your patient based on the
    opening, previous dialogue and physical examinations. Your closure should includes the following:  
    1. Briefly summarize the history and physical findings. 
    2. Briefly discuss the diagnostic possibilities. 
    3. Do not give a definitive diagnosis. 
    4. Briefly explain the planned diagnostic workup. 
    5. Avoid complicated medical terms. 
    6. Make the closure as concise as possible

\end{lstlisting}
\\ \hline
\end{tabular}

\begin{tabular}{|p{1cm}|p{14cm}|}
\hline
\textbf{Type}  & \textbf{Prompt} \\ \hline
Closure & 
\begin{lstlisting}
    Remember you will not receive the question from patient when you writing the closure!!!
        
    opening:
    {opening}
   
    Previous dialogue:
    {chat_history}
    
    physical examinations:
    {pre_closure}
    
    Here is an example of the closure: 
    As an example, if a patient's chief complaint is chest pain, the question that the case 
    embodies is, what is causing the chest pain? In this instance, closure should include 
    the formulation of a differential diagnosis consisting of the most likely causes of the 
    patient's chest pain along with their associated workups. By contrast, if the patient has 
    a history of diabetes mellitus and is presenting for follow,up, the case is posing two 
    questions: First, is the patient's diabetes well controlled? And second, is the patient 
    experiencing complications such as diabetic retinopathy or nephropathy? Here, both 
    questions should he addressed, and the workup should aim to determine whether 
    the diabetes is well controlled (HhA1c) as well as to look for complications such as 
    nephropathy (urine microalbuminuria). 
  
    In the second part, answer the following question. Your answer should based on the previous information including
    opening, previous dialogue, physical examination, and the closure you write in the part 1:
    {challenge_question}
    
    Output:
\end{lstlisting}
\\ \hline
Differential Diagnosis & 
\begin{lstlisting}
    You are a doctor and want to write a differential diagnosis to your patient based on the history, the answer provided by patient during QA, and the result of physical examination. In writing the differential diagnosis, you should list your three possible diagnoses and the historical and physical exam data that support them. You should list these three diagnoses in order of probability, from the most to the least probable, so the first diagnosis is the most essential diagnosis. 
    
    Here is the patient's information:
    {opening}
   
    You output should in the following format:
    {{"diagnosis1": {{"diagnosis": "name of the diagnosis", "Historical Findings": "a list that contains at most 3 historical data that support this diagnosis, output N/A if you think there is no historical findings support the diagnosis", "Historical reasons": "a list that contains the reason corresponding to the Historical Findings", "Physical exam data":, "a list that contains at most 3 Physical exam data that support this diagnosis, output N/A if you think there is no physical exam results support the diagnosis", "Physical exam data reasons": "a list that contains the reasons why you choose this physical exam result as support to each physical exam data" }}, "diagnosis2": {{"diagnosis": "name of the diagnosis", "Historical Findings": "a list that contains at most 3 historical data that support this diagnosis, output N/A if you think there is no historical findings support the diagnosis", "Historical reasons": "a list that contains the reason corresponding to the Historical Findings", "Physical exam data":, "a list that contains at most 3 Physical exam data that support this diagnosis, output N/A if you think there is no physical exam results support the diagnosis", "Physical exam data reasons": "a list that contains the reasons why you choose this physical exam result as support to each physical exam data" }}, "diagnosis3": {{"diagnosis": "name of the diagnosis", "Historical Findings": "a list that contains at most 3 historical data that support this diagnosis, output N/A if you think there is no historical findings support the diagnosis", "Historical reasons": "a list that contains the reason corresponding to the Historical Findings", "Physical exam data":, "a list that contains at most 3 Physical exam data that support this diagnosis, output N/A if you think there is no physical exam results support the diagnosis", "Physical exam data reasons": "a list that contains the reasons why you choose this physical exam result as support to each physical exam data" }} }}    
    
    Output:
\end{lstlisting}
\\ \hline
\end{tabular}
}

\newpage

\small
Table 9: All prompts used in our proposed MedExamLLM (LLM-as-clinical-skill-examiner) evaluation pipeline.

\lstset{
    basicstyle=\ttfamily,
    columns=fullflexible,
    breaklines=true,
    postbreak=\mbox{\textcolor{red}{$\hookrightarrow$}\space}
}
{\tiny
\centering

\begin{tabular}{|p{1cm}|p{14cm}|}
\hline
\textbf{Type}  & \textbf{Prompt}
\\ \hline
InfoGatherQA\\
Evaluation & 
\begin{lstlisting}
  As an evaluator for the United States Medical Licensing Examination (USMLE), your task is to assess the quality of the conversation between a medical student and a standardized patient. 
  Your role is to compare the medical student's question against the questions in the ground truth. You should look at the previous conversation as context for the evaluation.

  Here's how you should proceed:

  1. Look at the previous conversation and opening to understand the current context
  2. Check if the medical student's question is in the ground truth list of questions 
  3. Evaluate whether the medical student's question is consistent or mostly consistent with a question asked in the ground truth. If it is consistent or mostly consistent and not a repetition of a question
  asked in the previous conversation, assign the score of 1. It cannot be a follow-up to a question in the ground truth or an unrelated question, even if it is useful for gathering new information. 
  In that case, assign the score of 0.
  4. Provide a reason for your evaluation, and any additional feedback on the student's performance


  If the medical student answers the question correctly, you should set check to TRUE, otherwise set it to FALSE. Provide a reason for your evaluation, and any additional feedback on the student's performance.

  Previous Conversation:

  {prev_conversation}

  Opening:

  {opening}

  Medical Student's Question:

  {question}

  Questions in the Ground Truth:
  {ground_truth}

  Output Format:

  Your evaluation should be provided in JSON format, as follows:

  {{"reason": "the reason why you gave the score, and any additional feedback on the student's performance", "score": "0 or 1"}}

\end{lstlisting}
\\ \hline

Physical Exams\\
Evaluation & 
\begin{lstlisting}
As a USMLE evaluator, your role involves assessing and comparing a medical student's physical exam response to the provided target answer. Begin the assessment by carefully reviewing the provided chat history and target answer. Then, based on following specific criteria, determine the score for the student's answer.

Evaluation Criteria:
  1. Exam Coverage (0-60): The score reflects the proportion of target "Exam Component Maneuver" covered in the student's answer, with full marks at 60. For instance, covering 3 out of 5 target exams scores 36.
  2. Reason Relevance and Accuracy (0/20/40): Scores are allocated based on the relevance and accuracy of the reasons for exams, with fully relevant and accurate reasons scoring 40, partially relevant or accurate scoring 20, and irrelevant or inaccurate scoring 0.
  3. Extra Exams Penalty: Unmentioned exams in the target but present in the student's answer incur a -10 penalty each if unrelated to chat\_history, indicating overdiagnosis. If some extra exams are related mentioned in chat\_history, the score will be 0.

  medical student's answer:
  {pred}

  target answer provided:
  {target}

  chat_history:
  {chat_history}
  
Your evaluation should be provided in JSON format, as follows:
Do not include ```json ``` in the output.

  {{"score1": "score for Exam Coverage", "explanation1": "Detail each exam's alignment between the student's and target answers. Note any exams without correspondence.", "score2": "score for Reason Relevance and Accuracy", "explanation2": "Evaluate each reason's relevance and accuracy against the student's answer and chat history, noting mismatches.", "score3": "score for Extra Exams Penalty", "explanation3": "Identify extra exams in the student's answer, explain their relevance to chat history or lack thereof.","overall score": "sum of score1, score2 and score3"}}
\end{lstlisting}

\\ \hline
\end{tabular}

\begin{tabular}{|p{1cm}|p{14cm}|}
\hline
\textbf{Type}  & \textbf{Prompt} \\ \hline

Closure Evaluation & 
\begin{lstlisting}
  As a USMLE evaluator, your task is to thoroughly compare a medical student's answers "pred" with the provided "sample_closure" and "sample_answer" to ensure accuracy and completeness. The evaluation focuses primarily on two critical elements: the closure of the case and the handling of challenging questions.
  
  Each student's response should be meticulously scored based on how well it aligns with the sample responses. Deductions are made for discrepancies, omitted details, unclear explanations, or insufficient empathy. It's crucial to articulate the reasons for each deduction to provide meaningful feedback.
  
  Evaluation Criteria:
  
  Closure Evaluation:
  1. Diagnostic Impressions (0-20):
  For example, if the sample closure mentions three main diagnostic impressions, each equally important, and the student answer misses one aspect, approximately 1/3 * 20 = 6 points should be deducted. Specifically, if the sample closure states, "...the source of your pain can be a cardiac problem such as a heart attack, or it may be due to acid reflux, lung problems, or disorders related to the large blood vessels in your chest," it includes four diagnostic impressions: (1) cardiac problem, (2) acid reflux, (3) lung problems, (4) disorders related to the large blood vessels in your chest. If the student answer is, "we are tentatively considering a few possibilities: one could be a heart condition like angina or a heart attack; the other could be your GERD causing severe heartburn," covering (1) cardiac problem and (2) acid reflux but missing (3) and (4), then the score should be approximately 2/4 * 20 = 10 points. Of course, you can adjust the importance weighting based on the context and provide a reasonable score and rationale.
  
  2. Management Plans (0-30):
  For example, if the sample closure mentions five main management plans, each equally important, and the student answer misses two aspects, approximately 2/5 * 30 = 12 points should be deducted. Specifically, if the sample closure states, "I would like to perform a rectal exam to assess your prostate for benign growths or cancer. I would also like to order some blood tests, an X-ray, and possibly an MRI of your back to better determine the cause of your pain," it includes two main management plans: (1) rectal exam to assess prostate cancer, (2) blood tests, an X-ray, and possibly an MRI of your back to better determine the cause of your pain. If the student answer is, "We should start with blood tests and imaging studies, like an MRI, to get a better look at your spine," covering (2) but missing (1), then the score should be approximately 1/2 * 30 = 15 points. Of course, you can adjust the importance weighting based on the context and provide a reasonable score and rationale.

  Response to Challenging Questions Evaluation:
  3. Challenging Questions (0-30):
  30 points: Award full points if the student thoroughly and compassionately addresses all posed questions. The responses do not need to exactly match the sample answers but must be appropriate.
  0 points: Applied if the questions are left unanswered or the responses are completely off-topic, showing a failure to engage with the challenging aspects of the scenario.
  
  Additional Evaluation Areas:
  4. Language Accessibility (0-10):
  Award 10 points for using clear language and adequately explaining or avoiding medical jargon.
  Award 0 points if the language is inaccessible due to unexplained medical terms.

  Compassion (0-10):
  Award 10 points for demonstrating genuine warmth and empathy in responses.
  Award 0 points if the response lacks compassion.
  

  medical student's answer:
  {pred}

  sample_closure_and_answer:
  {sample_closure_and_answer}

  Your evaluation should be provided in JSON format, as follows:
  do not include ```json ``` in the output.

  {{"score1": "score for Diagnostic Impressions", "explanation1":...", "score2": "score for Management Plans, "explanation2": "...", "score3": "score for Challenging Questions", "explanation3": "...", "score4": "score for Language Accessibility", "explanation4": "...", "score5": "score for Compassion", "explanation5": "...", "overall score": "sum of score1, score2, score3, score4, score5"}}

  Output:
\end{lstlisting}

\\ \hline
\end{tabular}

\begin{tabular}{|p{1cm}|p{14cm}|}
\hline
\textbf{Type}  & \textbf{Prompt} \\ \hline

Diagnosis Evaluation & 
\begin{lstlisting}|
You are an evaluator for the United States Medical Licensing Examination (USMLE), your task is to assess the quality of the diagnoses written by medical student. 
  Your task is to compare the medical student's diagnosis book with the target diagnosis book provided. Both diagnosis books will contain three possible diagnoses. Each possible diagnosis follow the same structure: 
  1. Diagnosis Name: Begins with the name of the diagnosis being considered.
  2. Historical Finding: Lists the relevant historical information about the patient, including symptoms and medical history.
  3. Physical Exam Finding: Describes the pertinent physical examination finding observed in the patient.
  
  There is also an additional diagnosis book which include alternative but less likely diagnosis. The additional diagnosis book has the following structure: "Additional Diagnosis name: The explanation for this diagnosis".
  
  Here are the guidelines and metrics that help you grade the diagnosis book:
  1. Carefully examine the medical student's diagnosis book and the target diagnosis book.
  2. For each diagnosis listed in the medical student's diagnosis book:
    - Award 10 points if the medical student's diagnosis name exact match with the diagnosis in the target diagnosis book.
    - If the medical student mentions a similar diagnosis, or the same diagnosis with a different name comparing to the target diagnosis book, award on a scale of 0 to 10 points based on the similarity of the diagnosis names.
    - If no similar diagnosis is found in the target diagnosis book, consult the additional diagnosis book.
    - Award 5 points if the additional diagnosis book contains an exact match for the medical student's diagnosis name.
    - In cases where the medical student's diagnosis is similar but not identical to a diagnosis in the additional diagnosis book, award a score between 0 and 5 points based on the degree of similarity between the diagnosis names.
    - Note that the diagnosis in the additional diagnosis book is less accuracy than the target diagnosis book.
  3. For each matched diagnosis, carefully compare the historical finding between the medical student and the target:
    - Award 1 point for each of the medical student's historical finding that match the target's historical finding.
    - Student can get at most 3 points for the historical finding.
    - If the diagnosis name itself get 0 point, the historical finding also get 0 point.
    - If this target diagnosis does not have any historical finding, list "N/A" in the result output.
  4. For each matched diagnosis, carefully compare the physical examination finding between the medical student and the target:
    - Award 1 point for each of the medical student's physical finding that match the target's physical finding.
    - Student can get at most 3 points for the physical finding.
    - If the diagnosis name itself get 0 point, the physical finding also get 0 point.
    - If this target diagnosis does not have any physical finding, list "N/A" in the result output.
  5. The first diagnosis is the most essential one:
    - Award 10 points only if the medical student's first diagnosis name match the first diagnosis name in the target book.
    - Otherwise 0 point.

  Here is the medical student's diagnosis book:
  {pred}

  Here is the target diagnosis book:
  {target}
  
  Here is the additional diagnosis book and explanation:
  {additional_diagnosis}

  Your evaluation should be provided in JSON format, as follows:
  {{"diagnosis 1 name": "the score medical student get for the first diagnosis name", "diagnosis 1 historical finding": "the score medical student get for the first diagnosis historical finding", "diagnosis 1 physical finding": "the score medical student get for the first diagnosis physical finding", "diagnosis 2 name": "the score medical student get for the second diagnosis name", "diagnosis 2 historical finding": "the score medical student get for the second diagnosis historical finding", "diagnosis 2 physical finding": "the score medical student get for the second diagnosis physical finding", "diagnosis 3 name": "the score medical student get for the third diagnosis name", "diagnosis 3 historical finding": "the score medical student get for the third diagnosis historical finding", "diagnosis 3 physical finding": "the score medical student get for the third diagnosis physical finding", "order": "award 10 points only if the medical student's first diagnosis name match the first diagnosis name in the target book. Otherwise 0 point", "total score": "the total sum of the score for all three diagnosis name and finding", "quality score": "a single quality score to indicate the overall performance of the medical student. The score is selected from [1, 2, 3, 4, 5] (1=incomprehensible and incorrect, failing to provide a plausible differential diagnosis, 5=clear and correct, offering a well-reasoned differential diagnosis supported by relevant findings)", "reason": "the reason why you gave the score, and any additional feedback on the student's performance"}}
  
  Output:
\end{lstlisting}
\\ \hline
\end{tabular}
}

\newpage

{
\small
Table 10: Additional prompts for InfoGatherQA (LLM-as-clinical-skill-examiner) evaluation pipeline.
In Configuration 1, no credit is given for questions displaying slight consistency, representing a hard examiner. Configuration 2 gives partial credit (0.5) for these questions, representing a more lenient examiner. Configuration 3, on the other hand, does not address slight consistent questions in its prompt.

\lstset{
    basicstyle=\ttfamily,
    columns=fullflexible,
    breaklines=true,
    postbreak=\mbox{\textcolor{red}{$\hookrightarrow$}\space}
}
{\tiny
\centering

\begin{tabular}{|p{1cm}|p{14cm}|}
\hline
\textbf{Type}  & \textbf{Prompt}
\\ \hline
Configuration\\
1 & 
\begin{lstlisting}
  As a USMLE evaluator, your role is to assess the quality of the conversation between a medical student and a standardized patient for information gathering. 
  To evaluate, you have to compare the medical student's question against the questions in the ground truth. You should look at the previous conversation and opening as context for the evaluation.

  Here's how you should proceed:

  1. Look at the previous conversation and opening to understand the current context.
  2. Check if the medical student's question is in the ground truth list of questions.
  3. Evaluate whether the medical student's question is consistent with a question asked in the ground truth. 
      - If the question is consistent and explicitly mentioned (it is ok if it is more specific) in the ground truth, set the score to 1.
      - If the question is slightly consistent but not explicitly mentioned in the ground truth, set the score to 0. 
      - If the question is a repetition of a question asked in the previous conversation, or not covered in the ground truth, set the score to 0.
  4. Provide a reason for your evaluation, citing specific questions from the ground truth for consistency, and offer any additional feedback on the student's performance.

  Previous Conversation:
  {prev_conversation}

  Opening:
  {opening}

  Medical Student's Question:
  {question}

  Questions in the Ground Truth:
  {ground_truth}

  Let's think step by step and evaluate the student's performance using the criteria mentioned above.

  Output Format:
  Your evaluation should be provided in JSON format, as follows:
  {{"reason": "the reason why you gave the score, and any additional feedback on the student's performance", "score": "0 or 1"}}

\end{lstlisting}
\\ \hline

Configuration\\
2 & 
\begin{lstlisting}
  As a USMLE evaluator, your role is to assess the quality of the conversation between a medical student and a standardized patient for information gathering. 
  To evaluate, you have to compare the medical student's question against the questions in the ground truth. You should look at the previous conversation and opening as context for the evaluation.

  Here's how you should proceed:

  1. Look at the previous conversation and opening to understand the current context.
  2. Check if the medical student's question is in the ground truth list of questions.
  3. Evaluate whether the medical student's question is consistent with a question asked in the ground truth.
      - If it is consistent and explicitly mentioned (it is ok if it is more specific or include more details) in the ground truth, set the score to 1.
      - If it is consistent but not explicitly mentioned in the ground truth, set the score to 0.5.
      - If it is a repetition of a question asked in the previous conversation, or not covered in the ground truth, set the score to 0.
  4. Provide a reason for your evaluation, citing specific questions from the ground truth for consistency, and offer any additional feedback on the student's performance.

  Previous Conversation:
  {prev_conversation}

  Opening:
  {opening}

  Medical Student's Question:
  {question}

  Questions in the Ground Truth:
  {ground_truth}

  Let's think step by step and evaluate the student's performance using the criteria mentioned above.

  Output Format:
  Your evaluation should be provided in JSON format, as follows:
  {{"reason": "the reason why you gave the score, and any additional feedback on the student's performance", "score": "0 or 0.5 or 1"}}

\end{lstlisting}

\\ \hline
\end{tabular}

\begin{tabular}{|p{1cm}|p{14cm}|}
\hline
\textbf{Type}  & \textbf{Prompt} \\ \hline

Configuration \\ 3 & 
\begin{lstlisting}
  As an evaluator for the United States Medical Licensing Examination (USMLE), your task is to assess the quality of the conversation between a medical student and a standardized patient. 
  Your role is to compare the medical student's question against the questions in the ground truth. You should look at the previous conversation as context for the evaluation.

  Here's how you should proceed:

  1. Look at the previous conversation and opening to understand the current context
  2. Check if the medical student's question is in the ground truth list of questions 
  3. Evaluate whether the medical student's question is consistent or mostly consistent with a question asked in the ground truth. If it is consistent or mostly consistent and not a repetition of a question
  asked in the previous conversation, assign the score of 1. It cannot be a follow-up to a question in the ground truth or an unrelated question, even if it is useful for gathering new information. 
  In that case, assign the score of 0.
  4. Provide a reason for your evaluation, and any additional feedback on the student's performance


  If the medical student answers the question correctly, you should set check to TRUE, otherwise set it to FALSE. Provide a reason for your evaluation, and any additional feedback on the student's performance.

  Previous Conversation:

  {prev_conversation}

  Opening:

  {opening}

  Medical Student's Question:

  {question}

  Questions in the Ground Truth:
  {ground_truth}

  Output Format:

  Your evaluation should be provided in JSON format, as follows:

  {{"reason": "the reason why you gave the score, and any additional feedback on the student's performance", "score": "0 or 1"}}

\end{lstlisting}

\\ \hline
\end{tabular}
}


\normalsize

\twocolumn

\section{Detailed Data Processing and Copyright Compliance}
\label{apx:sec:data_processing_details}
\textbf{Data Sources and Scope}:
The original USMLE Step 2 CS case materials were obtained exclusively from public, non-commercial sources. The 44 cases include structured patient information, physician instructions, and sample answers for each encounter.

\paragraph{Transformative Paraphrasing Pipeline}:
To ensure the dataset meets the standards of the U.S. Copyright Office Fair Use Index and aligns with practices in MedQA~\cite{jin2021disease}, we designed a two-step pipeline:
\begin{itemize}
\item \textbf{Step 1: Automated Paraphrasing.} We randomly sampled one-third of sentences from the InfoGatherQA and Closure sections. These were paraphrased with ChatGPT, performing:
\begin{itemize}
\item Replacement of specialized medical terminology with synonyms of equivalent meaning and appropriateness;
\item Sentence-level and dialogue-level rewriting, altering structure and wording, and reworking the flow of the interaction, while rigorously preserving medical intent and facts.
\end{itemize}
This process constitutes a substantially deeper transformation than MCQ option shuffling, as adopted in MedQA~\cite{jin2021disease}, since it operates on open-ended, multi-turn dialogue data.
\item \textbf{Step 2: Domain Expert Verification.} All cases were independently reviewed by experienced clinicians to ensure:
\begin{itemize}
\item The paraphrased content is clinically accurate and preserves intended meaning;
\item The linguistic and stylistic changes are sufficiently transformative at both the sentence and discourse levels;
\item No new factual inaccuracies or clinical ambiguities were introduced during paraphrasing.
\end{itemize}
\end{itemize}

\paragraph{Dataset Statistics and Distribution.}
Finally, the MedQA-CS dataset comprises 44 clinical cases, each spanning the full simulated patient encounter. For each case, we generate:
\begin{itemize}
    \item \textbf{Information Gathering (InfoGatherQA):} Multiple utterance-level question-answer pairs per case (totaling 1,535 data points across all cases);
    \item \textbf{Physical Exam:} One structured output per case (44 in total);
    \item \textbf{Closure:} One structured output per case (44 in total);
    \item \textbf{Differential Diagnosis:} One structured output per case (44 in total).
\end{itemize}
Thus, while the Information Gathering section accounts for the majority of dataset size due to its utterance-level granularity, each of the other sections (Physical Exam, Closure, Differential Diagnosis) is represented by one comprehensive data point per case. In total, MedQA-CS consists of 1,667 instruction–input–output triplets.

We highlight these numbers here for transparency, in response to reviewer requests for a clear data breakdown by category. Additional details and code for generating these statistics are available in our GitHub repository.

\paragraph{Licensing and Intended Use}:
The MedQA-CS dataset, all processing scripts, and related code are released under a CC BY-NC 4.0 license, strictly for non-commercial, research, and educational purposes.

\paragraph{Transparency and Reproducibility}:
All details of the data source URLs, processing scripts, sampling strategies, paraphrasing instructions, and expert validation procedures are fully documented in our GitHub repository\footnote{URL will be made available upon publication acceptance}. This ensures transparency, enables reproducibility, and allows the community to further scrutinize and extend our work.

\section{MedQA-CS details}
\label{apx:sec:method}

    \subsection{InfoGatherQA}
    \label{apx:sec:method_InfoGatherQA}
    \paragraph{MedStuLLM}
\label{apx:sec:method_InfoGatherQA_student}
The patient encounters are designed to replicate situations commonly seen in clinics, doctors' offices, and emergency departments.
In this part, medical students need to interact with the patient or Standardized Patient (SP), address questions, and discuss diagnoses and follow-up plans. 
As shown in Table~\ref{tab:InfoGatherQA_workflow}, the input for MedStuLLM in this InfoGatherQA stage should be the doorway information and prior conversation history between the student and SP. 
Consequently, the output is the subsequent question that the student will ask the SP.
It is important to note that we treat each conversation round as independent for the sake of fair comparison. 
Therefore, the question output generated by the MedStuLLM in one round does not influence the input of subsequent rounds. 
In a real exam environment, incorrect questions could impact the following steps. 
However, by simplifying this part into independent InfoGatherQA, we can maximize the reliability of the MedExamLLM evaluation, which is reflected in the high human-human and human-AI agreement demonstrated in Sections \ref{sec:method_data_quality} and \ref{sec:result_MedExamLLM}.
In InfoGatherQA, each data point has a fixed input and reference output, enabling us to convert this section into open-ended question generation and answering with reference output. 
This setting, where we have already seen some successful cases of LLM-as-Judge in clinical NLP domain~\cite{zheng2024judging}, significantly reduces the "subjective" nature of evaluations that occur in free dialogue, a task current LLMs are not yet reliably equipped to handle as Judges in the clinical domain.

\begin{table}[t!]
    \scriptsize
    \centering
    \scalebox{1}{
    \begin{tabular}{|p{7.2cm}|}
        \hline
        \textbf{Doorway Information} \\
        Opening Scenario: Joseph Shorr, a 46-year-old male, comes to the ED complaining of chest pain. \\
        Vital Signs: BP: 165/85 mm Hg Temp: 98.6°F (37°C) RR: 22/minute HR: 90/minute, regular \\
        Examinee Tasks: \\
        1. Take a focused history. \\
        2. Perform a focused physical exam (do not perform rectal, genitourinary, or female breast exam). \\
        3. Explain your clinical impression and workup plan to the patient. \\
        4. Write the patient note after leaving the room. \\
        \hline

        \rowcolor{white}
        \multicolumn{1}{|c|}{\scriptsize \textbf{MedStuLLM (LLM-as-medical-student)}} \\
        \hline

        \rowcolor{yellow!15}
        \textbf{INPUT:} doorway information; chat history from round 0 to i-1 (i is current round num, 1 <= i <= max round number) \\
        \hline
        
        \rowcolor{green!5}
        \textbf{Instruction:} You are a doctor and your job is to take a detailed focused history is essential to the formulation of a differential diagnosis and workup plan later. You should begin by reading the history information and previous dialogue below carefully. Then decide which question to ask in order to get more detailed information of patient's concern and give the reason why you choose this question. Your question should include the following aspects if you think it is necessary: ......\\
        \hline
        
        \rowcolor{cyan!15}
        \textbf{OUTPUT:} Can you describe where exactly you feel the pain and whether it spreads to any other parts of your body? \\
        \hline
        
        \rowcolor{white}
        \multicolumn{1}{|c|}{\scriptsize \textbf{MedExamLLM (LLM-as-clinical-skill-examiner)}} \\
        \hline

        \rowcolor{yellow!15}
        \textbf{INPUT:} 
        \textbf{MedStuLLM input and output} (InfoGatherQA round i); \textbf{ground truth} InfoGatherQA question bank \\
        \hline
        
        \rowcolor{green!10}
        \textbf{Instruction:} As a USMLE evaluator, your role is to assess the quality of the conversation between a medical student and a standardized patient for information gathering.
        \textbf{Evaluation Criteria}:
        \textbf{Consistency}: The question must align with the ground truth question bank. \textbf{Originality}: The question must be novel and not repeat any previously asked question within the same conversation. ...... \\
        \hline
        
        \rowcolor{cyan!15}
        \textbf{OUTPUT: Correct.} Reason: The medical student's question 'Can you describe where exactly you feel the pain and whether it spreads to any other parts of your body?' is consistent with the ground truth questions 'Please show me exactly where you feel the pain.' and 'Does the pain move anywhere?'. The student's question is relevant and helps in gathering important information about the patient's condition \\
        \hline
    \end{tabular}
    }
    \caption{InfoGatherQA workflow. More details for this section can be found in appendix~\ref{apx:sec:method_InfoGatherQA}.}
    \label{tab:InfoGatherQA_workflow}
\end{table}

\noindent\textbf{MedExamLLM}
\label{apx:sec:method_InfoGatherQA_examiner}
For Scoring Criteria, we employ a comprehensive evaluation framework to assess the quality of the questions generated by our InfoGatherQA system. Each generated question is compared against a reference bank of ground truth questions, previous conversation, and the opening. Our LLMs and human evaluators follow the same rubric, where a question can be awarded a point if it satisfies the following criteria: 1. Consistency: The generated question aligns with the information and requirements outlined in the ground truth question bank. It must be found in the ground truth bank no matter how relevant it is otherwise.
2. Originality: The question is novel and does not repeat a previously asked question within the same conversation.
A question is awarded a point only if it meets both the consistency and originality criteria. Otherwise, it does not earn a point.

Our InfoGatherQA pipeline generates questions independently without considering previously generated questions. Consequently, we evaluate each question separately rather than assessing the entire conversation case at once. It is important to note that a complete conversation case typically represents an encounter between a medical student and a patient. However, since our current approach treats each question generation as an independent event, evaluating the individual questions separately provides a more accurate and focused assessment of our InfoGatherQA pipeline's performance in this specific task. To provide the evaluator with the necessary context, we supply the \textbf{chat history (H)}, \textbf{ground truth questions bank (B)}, and \textbf{opening (O)} as input. These are used to evaluate \textbf{generated question (Q)} against the consistency and originality criteria.
The choice of using chat history and opening reflects the inputs of the InfoGatherQA pipeline. This gives the examiner proper context of the situation. The ground truth questions bank can be obtained by extracting all the questions in the ground truth, which are target questions provided by the USMLE. By using the ground truth questions bank instead of a single target question, we ensure that the evaluation is agnostic to the order in which the questions are presented. 
The output is generated in JSON format, containing two components:  (1) a reasoning section explaining the decision and providing additional feedback and (2) the score of 0 or 1. 
Once all question within a case have been evaluated, we calculate the overall performance score using the following equation:
\vspace{-1mm}
\begin{equation}\nonumber
\text{Performance Score} = \frac{\text{Number of Points Awarded}}{\text{Number of Points Available}}
\end{equation}
\vspace{-0.5mm}
This performance score represents the proportion of generated questions that met both the consistency and originality criteria, providing a quantitative measure of the system's ability to ask highly relevant questions based on USMLE provided ground truths.

    \subsection{Physical Exams}
    \label{apx:sec:method_PhysicalExams}

\textbf{MedStuLLM}
\label{apx:sec:method_PhysicalExams_student}
The physical exam is after the patient encounter section. 
The medical student needs to write down the physical exam based on the doorway information and chat history in the patient encounter section.
We used zero-shot prompt for designing the prompt template for Physical Exam section by giving the model basic instructions. 
We take both doorway information and completed ground truth conversation history between the student and SP as input. 
For the output, to better evaluate the performance, we designed a prompt template to output both physical exams and the reason why the patient needs to take those exams.

\noindent\textbf{MedExamLLM}
\label{apx:sec:method_PhysicalExams_examiner}

\begin{table}[ht]

\centering
\small
\renewcommand{\arraystretch}{1.5}
\scalebox{0.6}{
\begin{tabular}{p{5cm}|p{2cm}|p{2.8cm}}
\hline
\textbf{Scoring Criteria} & \textbf{USMLE} & \textbf{Evaluation} \\ \hline
Fail to perform a required exam & Deduct points & Deduct points \\ \hline
Conduct extra maneuvers or Examine nonrequired systems & Not get or deduct points & Deduct points for not relevant exams \\ \hline
Perform a given procedure wrong (including preparation) & Deduct points & N/A \\ \hline
Be not respectful or friendly in conducting processing & Deduct points & N/A \\ \hline
Have a wrong reason for a maneuver & N/A & Deduct points \\ \hline
\end{tabular}
}
\caption{Scoring criteria for the assessment of physical exams in USMLE and our evaluation}
\label{tab:PhysicalExamsEvaluationCriteria}
\end{table}

\begin{table}[t!]
    \scriptsize
    \centering
    \scalebox{1}{
    \begin{tabular}{|p{7.2cm}|}
        \hline

        \rowcolor{white}
        \multicolumn{1}{|c|}{\scriptsize \textbf{MedStuLLM (LLM-as-medical-student)}} \\
        \hline

        \rowcolor{yellow!15}
        \textbf{INPUT:} doorway information; ground truth chat history \\
        \hline
        
        \rowcolor{green!5}
        \textbf{Instruction:} You are a doctor and want to figure out what the physical exam you have to do to physical test the patient's symptoms appear in the opening and previous dialogue with patient. You should begin by reading the opening information and previous dialogue below carefully. Then decide which physical exam and corresponding maneuver should take: Physical exams should be choose from the following component: Neck, Chest, Heart, Abdomen, Extremities, VS, CV, Pulmonary, Back, Hips, Neuro, HEENT, Skin, You should only choose the physical exams that mentioned in the opening and previous dialogue and also output the reason why you choose this ......\\
        \hline
        
        \rowcolor{cyan!15}
        \textbf{OUTPUT:} Heart: Auscultation for possible abnormal heart sounds and rhythm such as murmurs, gallop sound, or arrhythmias reason: The patient is presenting with chest pain, shortness of breath, and sweating which can indicate a cardiac issue like angina or a heart attack. This is further supported by his hypertension, high cholesterol, history of GERD, and cocaine use. \\
        \rowcolor{cyan!15}
        Chest: ... reason: ... \\
        \rowcolor{cyan!15}
        Pulmonary: ... reason: ... \\
        \hline
        
        \rowcolor{white}
        \multicolumn{1}{|c|}{\scriptsize \textbf{MedExamLLM (LLM-as-clinical-skill-examiner)}} \\
        \hline

        \rowcolor{yellow!15}
        \textbf{INPUT:} 
        \textbf{MedStuLLM input and output} (Physical Exams); \textbf{ground truth} Physical Exams \\
        \hline
        \rowcolor{green!10}
        \textbf{Instruction:} As a USMLE evaluator, your role involves assessing and comparing a medical student's physical exam response to the provided target answer. ...... 
        \textbf{Evaluation Criteria}:
        \textbf{Exam Coverage (0-60 points)}: This component evaluates whether the response addresses the necessary examinations, aligning with the USMLE rubric. \textbf{Reason Relevance and Accuracy (0/20/40 points)}: This component assesses the professionalism and accuracy of the generated output. \textbf{Extra Exams Penalty}: Points are deducted for suggesting unnecessary examinations, discouraging over-diagnosis. ....... \\
        \hline
        
        \rowcolor{cyan!15}
        \textbf{OUTPUT: "overall score": 76} {"score1": 36, "explanation1": "The student's answer covered 3 out of the 5 target exam components: Heart (auscultation), Chest (inspection and palpation), and Pulmonary (auscultation). However, the student did not mention Neck examination (JVD, carotid auscultation), Ocular examination (auscultation, palpation), Abdominal exam (auscultation, palpation, percussion), or Extremities (checked peripheral pulses, checked blood pressure in both arms, looked for edema and cyanosis).", "score2": 40, "explanation2": ..., "score3": 0, "explanation3": ...}
\\
        \hline
    \end{tabular}
    }
    \caption{Physical Exam workflow. More details for this section can be found in appendix~\ref{apx:sec:method_PhysicalExams}.}
    \label{tab:Physical_Exam_workflow}
\end{table}

In assessing the output of USMLE-Physical Examinations, we compare the generated responses to the expected correct answers. Both LLMs and human evaluators follow the same rubric, which is adapted from the scoring guidelines of USMLE.
As shown in Table~\ref{tab:PhysicalExamsEvaluationCriteria}, USMLE focus on whether the student correctly performs the required examinations. However, since our system operates in a virtual environment, the generated output cannot conduct actual physical examinations. Consequently, our scoring rubric comprises three components:
1. Exam Coverage (0-60 points): This component aligns with the USMLE rubric and evaluates whether the response addresses the necessary examinations.
2. Reason Relevance and Accuracy (0/20/40 points): This component assesses the professionalism and accuracy of the generated output.
3. Extra Exams Penalty: This component deducts points for suggesting unnecessary examinations, discouraging over-diagnosis.
To enhance the accuracy and precision of GPT's assessments, our input includes not only the \textbf{created answers} to be evaluated and the \textbf{correct answer} but also the preceding \textbf{chat history} between the physician and the patient. 
This additional context allows the model to extract more relevant information for its evaluations.
The output is generated in JSON format, including an overall score, scores for three separate sections, and detailed explanations for each score.
To generate the prompt for GPT, we follow a four-step process: (1) Establish evaluator's role and objective. (2) Outline the procedure for evaluation to ensure that GPT will review all the required information. (3) Provide the specific evaluation criteria. (4) Specify the format for explaining scores for each module.

    \subsection{Closure}
    \label{apx:sec:method_Closure}

\begin{table}[t!]
    \scriptsize
    \centering
    \scalebox{0.95}{
    \begin{tabular}{|p{7.2cm}|}
        \hline
        \rowcolor{white}
        \multicolumn{1}{|c|}{\scriptsize \textbf{MedStuLLM (LLM-as-medical-student)}} \\
        \hline
        \rowcolor{yellow!15}
        \textbf{INPUT:} doorway information; ground truth chat history; ground truth Physical Exams; challenge question by patient (e.g., "Is it a heart attack? Am I going to die?")" \\
        \hline
        
        \rowcolor{green!5}
        \textbf{Instruction:} You are a medical student ...... The current section is called closure. There are two parts of this section, writing the closure and answer the question from the patient ...... Your closure should includes the following: 1. Briefly summarize the history and physical findings. 2. Briefly discuss the diagnostic possibilities. 3. Do not give a definitive diagnosis. 4. Briefly explain the planned diagnostic workup. 5. Avoid complicated medical terms. 6. Make the closure as concise as possible ...... \\
        \hline
        
        \rowcolor{cyan!15}
        \textbf{OUTPUT: Closure:} Mr. Shorr, based on your recent episode of severe chest pain that woke you up from sleep, and that it radiated to your neck, upper back, and left arm, alongside your reported hypertension, high cholesterol, and past GERD, coupled with your physical exam that did not show any major abnormal findings, we are tentatively considering a few possibilities: ...... To do this, we'll be performing a few tests like an EKG and blood work to ...... If these tests are inconclusive, we might also consider an esophagus evaluation. ......  \\
        \rowcolor{cyan!15}
        \textbf{Question Answer:} I understand that you're scared. Right now, we suspect a heart condition given your symptoms, but ...... \\
        \hline
        
        \rowcolor{white}
        \multicolumn{1}{|c|}{\scriptsize \textbf{MedExamLLM (LLM-as-clinical-skill-examiner)}} \\
        \hline

        \rowcolor{yellow!15}
        \textbf{INPUT:} 
        \textbf{MedStuLLM input and output} (Closure); \textbf{ground truth} Closure and Answer \\
        \hline
        \rowcolor{green!10}
        \textbf{Instruction:} As a USMLE evaluator, your role involves assessing and comparing a medical student's answer to the provided sample. In evaluating each answer, two key components must be examined: the closure and the response to challenging questions. ...... \textbf{Evaluation Criteria: Diagnostic Impressions (0-20 points)}: Discussing all initial diagnostic impressions noted in the sample closure. \textbf{Management Plans (0-30 points)}: Outlining a complete diagnostic approach while keeping diagnostic options open. \textbf{Challenging Questions (0-30 points)}: Adequately answering all questions with compassion. \textbf{Language Accessibility (0-10 points)}: Avoiding or explaining all medical jargon. \textbf{Compassion (0-10 points)}: Demonstrating warmth and empathy in the response ...... \\
        \hline
        
        \rowcolor{cyan!15}
        \textbf{OUTPUT: "overall score": 85} {"score1": 15, "explanation1": "The student's diagnostic impressions align with those in the sample, including the possibility of a heart condition or GERD. However, the student omitted other potential causes such as lung problems or disorders related to the large blood vessels in the chest.", "score2": 20, "explanation2": ..., "score3": 30, "explanation3": ..., "score4": 10, "explanation4": ..., "score5": 10, "explanation5": ...}
\\
        \hline
    \end{tabular}
    }
    \caption{Closure workflow. More details for this section can be found in appendix~\ref{apx:sec:method_Closure}.}
    \label{tab:Closure_workflow}
\end{table}

\textbf{MedStuLLM}
\label{apx:sec:method_Closure_student}
The closure is after finishing the Patient Encounter and Physical Exams, where the doctor and patient can shift their perspective to the future, finalize plans, and say goodbye. During closure, both the doctor and patient may communicate by expressing emotions, sharing information, and making parting comments. Doctors may also use closure to educate patients, clarify, summarize, and provide parting comments
In this part, the student needs to write a summary that answer the following questions. 
\textbf{(1)}Make a transition to mark the end of your encounter. For example, ``\emph{Thank you for letting me examine you, Mr. Jones. I'd like to discuss the next steps.}.'' 
\textbf{(2)} Summarize the chief complaint and the HPI if you have not already done so before the physical exam.
\textbf{(3)} Summarize your findings from the physical exam.
To implement the prompt template, we included brief instructions covering these questions and provided a simple example from the USMLE Step 2 CS textbook. We use the doorway information, chat history from patient encounters, and physical exams as inputs to generate the closure summary.

\noindent\textbf{MedExamLLM}
\label{apx:sec:method_Closure_examiner}
In assessing the output of USMLE-Closure, we compare it to the sample closure including responses to the challenging questions. Both GPT and human evaluators follow the same rubric, which is adapted from the scoring guidelines of USMLE.
The USMLE requirements for the Closure section:
\label{ClosureShouldBe}
1. \textbf{Explain} your diagnostic possibilities/workups.
2. \textbf{Avoid} complicated medical terms.
3. \textbf{Ask} if the patient has any concerns.
4. Be prepared to handle \textbf{challenging questions}.
5. Avoid giving false reassurances.
\textbf{Counsel} the patient.
6. Say goodbye, thank the patient, and leave the encounter.
The rubric covers 100\% of the USMLE scoring criteria, with the weight of each section set according to its importance. The explanation of diagnostic workup and the ability to manage difficult inquiries stand out as the most crucial aspects, each contributing 30\% to the overall score. The remaining three sections, which focus on the suitability of communication language, collectively constitute 40\% of the score.
1. Diagnostic Impressions (0-20):  discussing all initial diagnostic impressions noted in the sample closure.
2. Management Plans (0-30): outlining a complete diagnostic approach while keeping diagnostic options open.
3. Challenging Questions (0-30): adequately answering all questions with compassion.
4. Language Accessibility (0-10): avoiding or explaining all medical jargon.
5. Compassion (0-10): warmth and empathy in the response.

    \subsection{Differential Diagnosis}
    \label{apx:sec:method_Diagnosis}
    \begin{table}[t!]
    \scriptsize
    \centering
    \scalebox{0.95}{
    \begin{tabular}{|p{7.2cm}|}
        \hline

        \rowcolor{white}
        \multicolumn{1}{|c|}{\scriptsize \textbf{MedStuLLM (LLM-as-medical-student)}} \\
        \hline
        \rowcolor{yellow!15}
        \textbf{INPUT:} doorway information; ground truth chat history; ground truth Physical Exams; ground truth Closure; ground truth Clinical Note (History section and Physical Exam section) \\
        \hline
        
        \rowcolor{green!5}
        \textbf{Instruction:} You are a doctor and want to write a differential diagnosis to your patient based on the history, the answer provided by patient during QA, and the result of physical examination ...... \\
        \hline

        \rowcolor{cyan!15}
        \textbf{OUTPUT: Diagnosis1: Acute Coronary Syndrome. Historical Finding(s)}: Substernal chest pain radiating to the left arm, upper back, and neck Associated symptoms of nausea, sweating, and dyspnea History of hypertension and high cholesterol. Historical \textbf{reasons:} Radiation of pain to the left arm and neck is typical for ...... \textbf{Physical Exam Finding(s)}: Elevated blood pressure Tachypnea N/A. Physical exam data \textbf{reasons}: Hypertension can be a chronic condition contributing to ...... Diagnosis2: Myocardial Infarction Historical Finding(s): ...... Physical Exam Finding(s): ...... Diagnosis3: Cocaine-Induced Chest Pain. Historical Finding(s): ...... Physical Exam Finding(s): ...... \\
        \hline
        
        \rowcolor{white}
        \multicolumn{1}{|c|}{\scriptsize \textbf{MedExamLLM (LLM-as-clinical-skill-examiner) }} \\
        \hline

        \rowcolor{yellow!15}
        \textbf{INPUT:} 
        \textbf{MedStuLLM input and output} (Closure); \textbf{ground truth} Clinical Note (Differential Diagnosis section) \\
        \hline
        
        \rowcolor{green!10}
        \textbf{Instruction:} You are an evaluator for USMLE, your task is to assess the quality of the diagnoses written by medical student. Your task is to compare the medical student's diagnosis book with the target diagnosis book provided. ...... \textbf{Evaluation Criteria: Diagnosis Name (0-10 points)}: Exact match: 10 points; Partial match: 0-10 points based on similarity. \textbf{Historical Findings (0-3 points)}: 1 point for each matching historical finding.
        \textbf{Physical Findings (0-3 points)}: 1 point for each matching physical finding. Refer to additional diagnoses if no match is found in the correct answer. Exact match in additional diagnoses: 5 points; Partial match: 0-5 points based on similarity. ...... \\
        \hline

        \rowcolor{cyan!15}
        \textbf{OUTPUT: "total score": 27/52=51.92\%} {"diagnosis 1 name": 7, "diagnosis 1 historical finding": 2, "diagnosis 1 physical finding"": "N/A", ...... "order": 0, \textbf{"quality score": 4}, "reason": "The student did a good job in identifying relevant conditions based on symptoms and history, with a particularly strong match on Myocardial Infarction. Improvement could be seen in diagnosis order and closer matching of historical details."}
\\
        \hline
    \end{tabular}
    }
    \caption{Diagnosis workflow. More details for this section can be found in appendix~\ref{apx:sec:method_Diagnosis}.}
    \label{tab:Diagnosis_workflow}
\end{table}

\textbf{MedStuLLM}
\label{apx:sec:method_Diagnosis_student}
The patient note is the final section of the USMLE exam. It includes the following components: a summary of the information collected during the InfoGatherQA stage, a summary of the physical exam and its results, and the differential diagnoses supported by evidence from the previous two sections in patient notes.
Therefore, we will focus on designing the differential diagnosis part of the MedStuLLM pipeline. 
To implement the prompt template, we provide brief instructions and use zero-shot prompts. 
We input all previous information, including doorway information, ground truth InfoGatherQA conversation history, physical exams and closure information, and the first two parts of the patient note.
The output is the required content as described above.

\noindent\textbf{MedExamLLM}
\label{apx:sec:method_Diagnosis_examiner}
In assessing the quality of the output for Differential Diagnosis, we compare the generated responses to the expected correct answers. Both GPT models and human evaluators follow the same rubric, which is adapted from the scoring guidelines of USMLE. The evaluation primarily focuses on assessing the accuracy of the diagnosis name and the presence of relevant supporting findings that substantiate the diagnosis, with the following criteria: 
1.Diagnosis name (10 points): Award based on the correctness of the diagnosis. Give partial credit for the similar diagnosis. 
1) Exact match with the correct diagnosis name: 10 points
2) Partially correct, like a similar diagnosis or the same diagnosis with a different name: 0-10 points based on the similarity
2. Historical findings (0-3 points): 1 point for each historical finding that matches the correct answer's historical finding. 
3. Physical findings(0-3 points): 1 point for each physical finding that match the correct answer's physical finding. In cases where the correct answer has fewer or no physical findings listed, the maximum attainable score for this part will be equal to the number of physical findings present in the correct answer.
To enhance the accuracy and precision of GPT's assessments, our input includes not only the \textbf{generate answer} and the \textbf{correct answer} but also \textbf{additional diagnoses and explanation} to help GPT to evaluate the generated answer more precisely. This additional diagnosis and explanation is extracted from the Case Discussion Section of the USMLE answer, which provides alternative but less likely diagnoses. 
By incorporating this additional context, the model gains access to more diagnosis-relevant information, enabling a more comprehensive and nuanced evaluation of the generated responses.
In the scenario of incorporating additional diagnoses during the evaluation process, we extend our rubric with the following criteria. The primary objective is to consider the diagnoses from the additional diagnoses list only when the medical student's diagnosis is not present in the correct answer. Furthermore, we assign a reduced score for matches found in the additional diagnoses compared to those in the correct answer:
1. If no similar diagnosis is found in the correct answer, refer to the additional diagnoses list.
2. Award 5 points if the additional diagnoses list contains an exact match for the medical student's diagnosis name.
3. In cases where the medical student's diagnosis is similar but not identical to a diagnosis in the additional diagnoses list, award a score between 0 and 5 points based on the degree of similarity between the diagnosis names.
This approach ensures that the evaluation process prioritizes the correct answer as the primary reference while allowing for partial credit if the medical student's diagnosis aligns with the less likely or alternative diagnoses listed in the additional diagnoses section.
The output is generated in JSON format, comprising four key sections: scores for three differential diagnoses, a cumulative total score, a subjective overall quality score, and detailed explanations justifying each assigned score.
1. The scores for each diagnosis will be broken down into 3 separate detail scores: diagnosis name, historical finding, and physical finding.
2. The total score is calculated by summing the points for each diagnosis, historical finding, physical finding, and order score, then dividing by the maximum possible points to yield a final score between 0 and 1. 
3. The subjective overall quality score, ranging from 1 to 5, will be assigned by the evaluation model to indicate the overall quality of the generated answer. This score is based on the model's subjective assessment and is not tied to specific rubric criteria. A rating of 1 signifies an output that is incomprehensible and entirely incorrect, failing to provide a plausible differential diagnosis. Conversely, a rating of 5 denotes a clear and correct output, offering a well-reasoned differential diagnosis supported by relevant findings. This subjective scoring component allows the evaluation model to provide an overarching assessment of the answer's quality, complementing the objective rubric-based scoring.
4. The detailed explanation will list the reason behind the assigned scores for each component, providing insights into the evaluation model's assessment process.

\onecolumn

\section{Human Annotation Guidelines}
\label{apx:sec:human_annotation_guidelines}
The goal of human eval is to verify the reliability of the MedQA-CS design, and was evaluated through the agreement among experts who assessed the MedStuLLM (GPT-4) results across four sections. 
To do so, we worked with 3 medical practitioners with at least 2 years of hospital experience. 
Their tasks were to follow each requirement of MedExamLLM to evaluate MedStuLLM's output.

\subsection{InfoGatherQA}

Although the USMLE Step 2 CS does not provide an official guideline specifying what constitutes as a point (a check in the checklist), we have developed the following system based on expert recommendations. To award a point, the question must satisfy two criteria:
\textbf{Consistency}: The question must be present in the ground truth list of questions and convey the same idea. Highly relevant questions that are not found in the ground truth list do not meet this criterion.
\textbf{Originality}: The question must not repeat any question from the chat history and must present unique findings.

During evaluation, both the opening and the chat history will be reviewed to understand the context. The evaluation process is as follows:
1. Verify if the question appears in the ground truth list of questions and conveys the same idea.
2. Examine the chat history to ensure that the question does not repeat any previously asked questions.
Your evaluations should include a binary score (1 for meeting both criteria, 0 for not meeting them) and provide the reasoning behind the evaluation. Please see the following for an example.

\begin{table}[ht]
\vspace{-1mm}
\centering
\scalebox{0.62}{
\begin{tabular}{|p{3cm}|p{7.5cm}|c|}
\hline
\textbf{Question} & \textbf{Reason} &  \\
\hline
What were you doing when the chest pain started? & The student's question is consistent with the ground truth. They are on the right track in gathering information about the onset of the chest pain. & \checkmark \\
\hline
\end{tabular}
}
\label{tab:InfoGatherQA_human_anno_guideline_example}
\vspace{-7mm}
\end{table}

\subsection{Physical Exams}
\textbf{Exam Coverage (0-60):} The score reflects the proportion of target "Exam Component Maneuver" covered in the student's answer, with full marks at 60. For instance, covering 3 out of 5 target exams scores 36.
\textbf{Reason Relevance and Accuracy (0/20/40):} Scores are allocated based on the relevance and accuracy of the reasons for exams, with fully relevant and accurate reasons scoring 40, partially relevant or accurate scoring 20, and irrelevant or inaccurate scoring 0.
\textbf{Extra Exams Penalty:} Unmentioned exams in the target but present in the student's answer incur a -10 penalty each if unrelated to chat history, indicating overdiagnosis. If some extra exams are related mentioned in chat history, the score will be 0.
Your evaluations are expected to include scores for each part and reasons why deduct some points.
Please give some brief reason for any point deduction. Here is an example as the table \ref{tab:physical_exam_human_example} shown.

\begin{table}[ht]
\centering
\scalebox{0.5}{
\begin{tabular}{|p{3cm}|p{6.8cm}|p{2.4cm}|}
\hline
\textbf{Ground Truth} & \textbf{MedStuLLM} & \textbf{Scores and Reason} \\
\hline
\textbf{1.CV} Exam: Auscultation.

\textbf{2.Pulmonary} Exam: Auscultation.

\textbf{3.Abdominal} Exam: Inspection, auscultation, palpation, percussion.Check for CVA tenderness.

\textbf{4.Extremities} Exam: Inspection.

 & \textbf{1.Abdomen:} Palpation and percussion
 
 (reason: To check for any abdominal tenderness, masses, or kidney enlargement which can be related to bloody urine).
 
 \textbf{2.CV:} Auscultation
 
 (reason: To assess for any cardiovascular issues that might affect kidney function).
 
 \textbf{3.VS:} Monitoring vital signs
 
 (reason: To establish a baseline for the patient's general health).
 
 \textbf{4.Skin:} Visual inspection
 
 (reason: To assess for signs of jaundice or other skin changes related to kidney disease). & \textbf{overall score:} 60 / 100
 
 \textbf{score1:}  30 / 60
 
 (reason: Missed Pulmonary and Extremities Exam, 2/4 * 60 = 30).
 
 \textbf{score2}: 40 / 40
 
 \textbf{score3:} -10
 
 (reason: No skin test required) \\
\hline
\end{tabular}
}
\vspace{-3mm}
\caption{Annotation Example for Physical Exams}
\vspace{-6mm}
\label{tab:physical_exam_human_example}
\end{table}

\subsection{Closure}
In evaluating the closure aspect, you are required to examine two specific sections: "Diagnostic Impressions" and "Management Plans":

\noindent\textbf{Diagnostic Impressions (0-20):}
20 points for discussing all initial diagnostic impressions noted in the sample closure.
0 points for not addressing any initial diagnostic impressions.
\textbf{Management Plans (0-30):}
30 points for a comprehensive plan that includes all diagnostic approaches mentioned in the sample closure, encompassing both follow-up tests and lifestyle modifications (diet, exercise, smoking, alcohol).
0 points if there is no diagnostic plan or if a diagnosis is made prematurely.
In evaluating the response to the challenging questions, you are required to examine two specific sections:
\textbf{Challenging Questions (0-30):}
Comparing with the Sample Examine Response.
30 points for adequately answering all questions and with compassion.
0 points for not answering any questions.
In assessing both the closure component and the response to challenging questions, it is necessary to evaluate two key areas: "Language Accessibility" and "Compassion."
\textbf{Language Accessibility (0-10):}
10 points for clear communication, avoiding or explaining all medical jargon.
0 points for difficult comprehension due to many unexplained terms.
\textbf{Compassion (0-10):}
10 points for demonstrating warmth and empathy.
0 points for no compassion.
Please give some brief reason for any point deduction. Here is an example as the table \ref{tab:closure_human_example_1}, and table \ref{tab:closure_human_example_2} shown.

\begin{table}
\vspace{-2mm}
\centering
\scalebox{0.4}{
\begin{tabular}{|p{8cm}|p{10cm}|}
\hline
\textbf{Sample Closure} & \textbf{MedStuLLM} \\
\hline
"Mr. Short, the source of your pain can be a \textbf{cardiac problem} such as a \textbf{heart attack}, or it may be due to \textbf{acid reflux}, \textbf{lung problems}, or \textbf{disorders related to the large blood vessels} in your chest. It is crucial that we perform some tests to identify the source of your problem. We will start with an \textbf{ECG and some blood work}, but more complex tests may be needed as well. In the meantime, I strongly recommend that you \textbf{stop using cocaine}, since use of this drug can lead to a variety of medical problems, including heart attacks. I commend you for \textbf{quitting smoking} and encourage you to continue not to smoke, as cigarettes are known to worsen cardiovascular disease and increase your risk of developing a heart attack in the future. Do you have any questions for me?"
 & "Mr. Shorr, based on your recent episode of severe chest pain that woke you up from sleep, and that it radiated to your neck, upper back, and left arm, alongside your reported hypertension, high cholesterol, and past GERD, coupled with your physical exam that did not show any major abnormal findings, we are tentatively considering a few possibilities: one could be a \textbf{heart condition} like angina or a \textbf{heart attack}; the other could be your \textbf{GERD} causing severe heartburn. These are serious conditions, and we need to properly investigate. To do this, we'll be performing a few tests like an \textbf{EKG and blood work} to look for signs of heart damage or strain, and we might also need to run a stress test to see how your heart responds during exertion. If these tests are inconclusive, we might also consider an esophagus evaluation. Please remember that these are just possibilities, and we will have a better idea once we have the results of the tests." \\
\hline
\textbf{Sample Answer for Question} & \textbf{MedStuLLM} \\
\hline
\textbf{Challenging Question:}
Is it a heart attack? Am I going to die?

\textbf{Samlpe Answer:}
"Your chest pain is of significant concern. However, chest pain can be caused by a large variety of issues. We need to learn more about what's going on to know if your pain is life-threatening."
 & \textbf{"I understand that you're scared.} Right now, we suspect a heart condition given your symptoms, but we're conducting \textbf{further tests} to be sure. Know that even if it is something serious like a heart attack, we have excellent treatments available. Let's \textbf{not worry} about the worst case right now, and let's focus on getting you tested and treated. You're in good hands, \textbf{we will do everything we can for you.}" \\
\hline
\end{tabular}
}
\vspace{-3mm}
\caption{Human Annotation Example for Closure (1)}
\vspace{-5mm}
\label{tab:closure_human_example_1}
\end{table}

\subsection{Differential Diagnosis}
Compare the medical student's differential diagnosis with the correct answer's differential diagnosis provided. Both differential diagnoses will contain three possible diagnoses. Each diagnosis follows the same structure:
1. Diagnosis Name: The name of the diagnosis being considered.
2. Historical Finding: Lists the relevant historical information about the patient, including symptoms and medical history.
3. Physical Exam Finding: Describes the pertinent physical examination findings observed in the patient. 
There are also additional diagnoses listed, which are alternative but less likely possibilities. These have the format: "Additional Diagnosis Name: Explanation for this diagnosis."

Here are the guidelines and metrics that help you grade the differential diagnosis:

1. For each diagnosis listed in the medical student's differential diagnosis: 
1) Award 10 points if the medical student's diagnosis name exact match with the correct answer.
2) If the medical student mentions a similar diagnosis, or the same diagnosis with a different name comparing to the correct answer, award on a scale of 0 to 10 points based on the similarity of the diagnosis names.
3) If no similar diagnosis is found in the target diagnosis, consult the additional diagnoses.
4) Award 5 points if the additional diagnoses contain an exact match for the medical student's diagnosis name. 
5) In cases where the medical student's diagnosis is similar but not identical to a diagnosis in the additional diagnoses, award a score between 0 and 5 points based on the degree of similarity between the diagnosis names.
6) Note that the diagnosis in the additional diagnoses is less likely and accuracy than the diagnoses in the correct answer.

\begin{table}
\vspace{-2mm}
\centering
\scalebox{0.55}{
\begin{tabular}{|p{13.4cm}|}
\hline
\textbf{Scores and Reason} \\
\hline
\textbf{overall score:} 75 / 100
\textbf{Diagnostic Impressions:} 10 / 20
(reason: Missed lung problems, or disorders related to the large blood vessels in the chest)
\textbf{Management Plans:} 15 / 30
(reason: missed lifestyle modification, like stop using cocaine and quitting smoking)
\textbf{Challenging Questions:} 30 / 30
\textbf{Language Accessibility:} 10 / 10
\textbf{Compassion:} 10 / 10 \\
\hline
\end{tabular}
}
\vspace{-4mm}
\caption{Human Annotation Example for Closure (2)}
\vspace{-6mm}
\label{tab:closure_human_example_2}
\end{table}

2. For each matched diagnosis:
1) Award 1 point for each of the medical student's historical findings that match the correct answer's historical finding.
2) Student can get at most 3 points for the historical finding.
3) If the diagnosis name itself gets 0 point, the historical finding also gets 0 point.
4) If the corresponding diagnosis in the correct answer does not have any historical finding, list N/A in the scoring part.

3. For each matched diagnosis:
1) Award 1 point for each of the medical student's physical finding that match the correct answer's physical finding.
2) Student can get at most 3 points for the physical finding.
3) If the diagnosis name itself get 0 point, the physical finding also get 0 point.
4) If the corresponding diagnosis in correct answer does not have any physical finding, list N/A in the scoring part.
5) The first diagnosis is the most essential one. Award 10 points only if the medical student's first diagnosis name match the first diagnosis name in the correct answer. Otherwise 0 point.

Scoring:
1. Diagnosis name: The score medical student get for the diagnosis name
2. Diagnosis historical finding: The score medical student get for the diagnosis historical finding
3. Diagnosis physical finding: The score medical student get for the diagnosis physical finding
4. Order score: The score medical student get for the correct first diagnosis name
5. Total score: The total score is calculated by summing the points for each diagnosis, historical finding, physical finding, and order score, then dividing by the maximum possible points to yield a final score between 0 and 1. 
6. Quality score: A single quality score to indicate the overall performance of the medical student. The score is selected from [1, 2, 3, 4, 5] (1=incomprehensible and incorrect, failing to provide a plausible differential diagnosis, 5=clear and correct, offering a well-reasoned differential diagnosis supported by relevant findings)"

\section{Human Evaluation Results}
\label{apx:sec:human_annotation_results}
For the InfoGatherQA, we put the detailed human annotation results in our GitHub.

For the Physical Exams, we engaged 3 experts to review 10 cases and complete the annotation. For detailed information on Physical Exams, please refer to the table \ref{tab:human_anntation_physical_exam} and table \ref{tab:human_anntation_physical_exam_2}.

For the Closure, we engaged 3 experts to review 10 cases and completed the annotation. For detailed information in Closure, please refer to the table
\ref{tab:human_anntation_closure}, table \ref{tab:human_anntation_closure_2}, and table \ref{tab:human_anntation_closure_3}.

For the Differential Diagnosis, we engaged 3 experts to review 10 cases and completed the annotation. For detailed information in Differential Diagnosis, please refer to the table
\ref{tab:human_anntation_diagnosis}, table \ref{tab:human_anntation_diagnosis_2},
table \ref{tab:human_anntation_diagnosis_3}, and table \ref{tab:human_anntation_diagnosis_4}.

\begin{table}
\centering
\caption{Human Annotation for Physical Exams (Case 1 - Case 6). Score1 is the score of the Exam Coverage, Score2 represents Reason Relevance and Accuracy, and Score3 is for Extra Exam Penalty.}
\begin{tabular}{|p{0.7cm}|p{4.5cm}|p{4.5cm}|p{4.5cm}|}
\hline
\textbf{Case} & \textbf{Expert 1} & \textbf{Expert 2} & \textbf{Expert 3} \\
\hline
1 & \textbf{overall score: 70}
 
score1: 2.5/5 * 60 = 30

(reason: Missed Ocular examination, Extremities, and part of Abdominal exam)

score2: 40

score3: 0

(reason: Chest pain is fine)
& \textbf{overall score: 70}

score1: 2.5/5 * 60 = 30

(reason: Missed Ocular examination, Extremities, and part of Abdominal exam)

score2: 40

score3: 0
& \textbf{overall score: 70}

score1: 2.5/5 * 60 = 30

(reason: Missed Extremities, Ocular examination, and part of Abdominal exam)

score2: 40

score3: 0
\\
\hline
2 & \textbf{overall score: 60}
 
score1: 2/4 * 60 = 30

(reason: Missed Pulmonary and Extremities Exam)

score2: 40

score3: -10

(reason: Skin is not required)

& \textbf{overall score: 70}

score1: 2/4 * 60 = 30

(reason: Missed Pulmonary and Extremities Exam)

score2: 40

score3: 0
& \textbf{overall score: 60}

score1: 2/4 * 60 = 30
(reason: Missed Pulmonary and Extremities Exam)

score2: 40

score3: -10

(reason: No skin test is required)
\\
\hline
3 & \textbf{overall score: 70}
 
score1: 2/3 * 60 = 40

(reason: Missed Extremities Exam)

score2: 40

score3: -10

(reason: Abdomen is not required)
& \textbf{overall score: 60}

score1: 2/3 * 60 = 40

(reason: Missed Extremities Exam)

score2: 20

score3: 0
& \textbf{overall score: 70}

score1: 2/3 * 60 = 40

score2: 40

score3: -10

(reason: No Abdomen test is required)
\\
\hline
4 & \textbf{overall score: 56}

score1: 3/5 * 60 = 36

(reason: Missed Neurologic exam and Skin exam)

score2: 40

score3: -20

(reason: Chest and Heart are not required)

& \textbf{overall score: 76}

score1: 3/5 * 60 = 36

(reason: Missed Neurologic exam and Skin exam)

score2: 40

score3: 0

& \textbf{overall score: 66}

score1: 3/5 * 60 = 36

(reason: Missed Neurologic exam and Skin exam)

score2: 40

score3: -10

(reason: No heart test required)
\\
\hline
5 & \textbf{overall score: 65}

score1: 3/5 * 60 = 35

(reason: Missed CV Exam and Pulmonary Exam)

score2: 40

score3: -10

(reason: Neck is not required)

& \textbf{overall score: 76}

score1: 3/5 * 60 = 36

(reason: Missed CV Exam and Pulmonary Exam)

score2: 40

score3: 0
& \textbf{overall score: 76}

score1: 3/5 * 60 = 36

(reason: Missed CV Exam and Pulmonary Exam)

score2: 40

score3: 0
\\
\hline
6 & \textbf{overall score: 0}

score1: 0

(reason: The patient is not here.)

score2: 0

score3: 0
& \textbf{overall score: 0}

score1: 0

score2: 0

score3: 0
& \textbf{overall score: 0}

score1: 0

score2: 0

score3: 0
\\
\hline
\end{tabular}
\label{tab:human_anntation_physical_exam}
\end{table}

\begin{table}
\centering
\caption{Human Annotation for Physical Exams (Case 7 - Case 10). Score1 is the score of the Exam Coverage, Score2 represents Reason Relevance and Accuracy, and Score3 is for Extra Exam Penalty.}
\begin{tabular}{|p{0.7cm}|p{4.5cm}|p{4.5cm}|p{4.5cm}|}
\hline
\textbf{Case} & \textbf{Expert 1} & \textbf{Expert 2} & \textbf{Expert 3} \\
\hline
7 & \textbf{overall score: 45}

score1: 1/4 * 60 = 15

(reason: Missed Head and neck exam, Cardiovascular exam, and Pulmonary exam)

score2: 40

score3: -10

(reason: Neuro exam is not required)
& \textbf{overall score: 55}

score1: 1/4 * 60 = 15

(reason: Missed Head and neck exam, Cardiovascular exam, and Pulmonary exam)

score2: 40

score3: 0

& \textbf{overall score: 55}

score1: 1/4 * 60 = 15

(reason: Missed Head and neck exam, Cardiovascular exam, and Pulmonary exam)

score2: 40

score3: 0
\\
\hline
8 & \textbf{overall score: 70}

score1: 3/6 * 60 = 30

(reason: Missed HEENT, Pulmonary exam, and Abdominal exam)

score2: 40

score3: 0
& \textbf{overall score: 70}

score1: 3/6 * 60 = 30

(reason: Missed HEENT, Pulmonary exam, and Abdominal exam)

score2: 40

score3: 0

& \textbf{overall score: 60}

score1: 2/6 * 60 = 20

(reason: Missed HEENT, Pulmonary exam, Abdominal exam, and Musculoskeletal exam)

score2: 40

score3: 0

\\
\hline
9 & \textbf{overall score: 76}

score1:3/5 * 60 = 36

(reason: Missed Neurological exam, and Head and neck exam)

score2: 40

score3: 0

& \textbf{overall score: 76}

score1:3/5 * 60 = 36

(reason: Missed Neurological exam, and Head and neck exam)

score2: 40

score3: 0

& \textbf{overall score: 76}

score1:3/5 * 60 = 36

(reason: Missed Neurological exam, and Head and neck exam)

score2: 40

score3: 0

\\
\hline
10 & \textbf{overall score: 45}

score1: 1/4 * 60 = 15

(reason: Missed CV/pulmonary exam, Musculoskeletal exam, and Neurologic exam)

score2: 40

score3: -10

(reason: Skin is not required)
& \textbf{overall score: 70}

score1: 2/4 * 60 = 30
(reason: Missed CV/pulmonary exam and Neurologic exam)

score2: 40

score3: 0

& \textbf{overall score: 0}

score1: 1/4 * 60 = 15

(reason: Missed CV/pulmonary exam, Musculoskeletal exam, and Neurologic exam)

score2: 40

score3: -10

(reason: VS checking should not be because of suspicious infection)
\\
\hline
\end{tabular}
\label{tab:human_anntation_physical_exam_2}
\end{table}

\begin{table}
\centering
\caption{Human Annotation for Closure (Case 1 - Case 5). Score1 is the score of the Diagnostic Impressions, Score2 represents Management Plans, Score3 represents Challenging Questions, Score4 represents Language Accessibility, and Score5 is for Compassion.}
\scalebox{0.8}{
\begin{tabular}{|p{0.7cm}|p{4.5cm}|p{4.5cm}|p{4.5cm}|}
\hline
\textbf{Case} & \textbf{Expert 1} & \textbf{Expert 2} & \textbf{Expert 3} \\
\hline
1 & \textbf{overall score: 75}

score1: 10/20

(reason: Missed "lung problems, or disorders related to the large blood vessels in your chest")

score2: 15/30

(reason: Missed lifestyle modification)

score3: 30/30
score4: 10/10
score5: 10/10
& \textbf{overall score: 75}

score1: 10/20

(reason: Missed "lung problems, or disorders related to the large blood vessels in your chest")

score2: 15/30

score3: 30/30

score4: 10/10

score5: 10/10
& \textbf{overall score: 75}

score1: 10/20

(reason: Missed "lung problems, or disorders related to the large blood vessels in your chest")

score2: 15/30

(reason: Missed lifestyle modification)

score3: 30/30
score4: 10/10
score5: 10/10
\\
\hline
2 & \textbf{overall score: 100}

score1: 20/20

score2: 30/30

score3: 30/30

score4: 10/10

score5: 10/10
& \textbf{overall score: 98}

score1: 18/20

score2: 30/30

score3: 30/30

score4: 10/10

score5: 10/10
& \textbf{overall score: 100}

score1: 20/20

score2: 30/30

score3: 30/30

score4: 10/10

score5: 10/10
\\
\hline
3 & \textbf{overall score: 85}

score1: 20/20
score2: 15/30

(reason: Missed "perform a rectal exam")

score3: 30/30
score4: 10/10
score5: 10/10
& \textbf{overall score: 85}
score1: 18/20

score2: 17/30

score3: 30/30

score4: 10/10

score5: 10/10

& \textbf{overall score: 85}

score1: 20/20
score2: 15/30

(reason: Missed "perform a rectal exam")

score3: 30/30
score4: 10/10
score5: 10/10
\\
\hline
4 & \textbf{overall score: 85}

score1: 20/20

score2: 15/30

(reason: Missed "provide you with something for your pain and monitor your breathing to ensure sufficient oxygen intake")

score3: 30/30
score4: 10/10
score5: 10/10

& \textbf{overall score: 75}

score1: 20/20

score2: 10/30

score3: 15/30

score4: 10/10

score5: 10/10
& \textbf{overall score: 75}

score1: 10/20

(reason: Infection should not be included)

score2: 15/30

(reason: Missed "provide you with something for your pain and monitor your breathing to ensure sufficient oxygen intake")

score3: 30/30
score4: 10/10
score5: 10/10
\\
\hline
5 & \textbf{overall score: 75}

score1: 20/20
score2: 5/30

(reason: Missed "perform a pelvic ultrasound", "recommend that you stop drinking alcohol and avoid intense exercise and excess caffeine", and "provide you with some prenatal multivitamins")

score3: 30/30
score4: 10/10
score5: 10/10
& \textbf{overall score: 80}

score1: 20/20

score2: 10/30

score3: 30/30

score4: 10/10

score5: 10/10
& \textbf{overall score: 75}

score1: 20/20
score2: 5/30

(reason: Missed "perform a pelvic ultrasound", "recommend that you stop drinking alcohol and avoid intense exercise and excess caffeine", and "provide you with some prenatal multivitamins")

score3: 30/30
score4: 10/10
score5: 10/10

\\
\hline
\end{tabular}
}
\label{tab:human_anntation_closure}
\end{table}

\begin{table}
\centering
\caption{Human Annotation for Closure (Case 6 - Case 9). Score1 is the score of the Diagnostic Impressions, Score2 represents Management Plans, Score3 represents Challenging Questions, Score4 represents Language Accessibility, and Score5 is for Compassion.}
\begin{tabular}{|p{0.7cm}|p{4.5cm}|p{4.5cm}|p{4.5cm}|}
\hline
\textbf{Case} & \textbf{Expert 1} & \textbf{Expert 2} & \textbf{Expert 3} \\
\hline
6 & \textbf{overall score: 70}

score1: 20/20

score2: 15/30

(reason: Missed "Your daughter should always carry a snack or juice as an "emergency kit.")

score3: 15/30

(reason: Wrong answer "Your daughter may have either type 1 or type 2 diabetes...")

score4: 10/10

score5: 10/10
& \textbf{overall score: 73}

score1: 18/20

score2: 10/30

score3: 25/30

score4: 10/10

score5: 10/10
& \textbf{overall score: 85}

score1: 20/20

score2: 15/30

(reason: Missed "understand diabetes and understand it" and "know how to manage low glucose level. She should always carry a snack or juice as an "emergency kit.")

score3: 30/30

score4: 10/10

score5: 10/10
\\
\hline
7 & \textbf{overall score: 80}

score1: 20/20

score2: 15/30

(reason: No mention "social worker")

score3: 30/30

score4: 10/10

score5: 5/10

(reason: No mention financial assistance programs)

& \textbf{overall score: 79}

score1: 18/20

score2: 15/30

score3: 26/30

score4: 10/10

score5: 10/10
& \textbf{overall score: 0}

score1: 20/20

score2: 15/30

(reason: No mention "social worker" and "assisted living community / apartment complex for seniors")

score3: 30/30

score4: 10/10

score5: 5/10

(reason: No mention "financial assistance programs"
\\
\hline
8 & \textbf{overall score: 75}

score1: 10/20

(reason: Missed "resulted from a higher-than-normal dose of insulin or from skipping or delaying meals")

score2: 30/30

score3: 15/30

score4: 10/10

score5: 5/10

& \textbf{overall score: 83}

score1: 18/20

score2: 22/30

score3: 23/30

score4: 10/10

score5: 10/10

& \textbf{overall score: 75}

score1: 10/20

score2: 30/30

score3: 15/30

score4: 10/10

score5: 10/10
\\
\hline
9 & \textbf{overall score: 85}

score1: 20/20

score2: 15/30

(reason: Missed "antibiotics" and "emergency contraception")

score3: 30/30

score4: 10/10

score5: 10/10
& \textbf{overall score: 85}

score1: 20/20

score2: 15/30

score3: 30/30

score4: 10/10

score5: 10/10

& \textbf{overall score: 85}

score1: 20/20

score2: 15/30

(reason: Missed "collect evidence for charges", "potential infections - antibiotics", "emergency contraception")

score3: 30/30

score4: 10/10

score5: 10/10
\\
\hline
\end{tabular}
\label{tab:human_anntation_closure_2}
\end{table}

\begin{table}
\centering
\caption{Human Annotation for Closure (Case 10). Score1 is the score of the Diagnostic Impressions, Score2 represents Management Plans, Score3 represents Challenging Questions, Score4 represents Language Accessibility, and Score5 is for Compassion.}
\begin{tabular}{|p{0.7cm}|p{4.5cm}|p{4.5cm}|p{4.5cm}|}
\hline
\textbf{Case} & \textbf{Expert 1} & \textbf{Expert 2} & \textbf{Expert 3} \\
\hline
10 & \textbf{overall score: 85}

score1: 20/20

score2: 15/30

(reason: Missed "blood thinners to prevent further complications")

score3: 30/30

score4: 10/10

score5: 10/10

& \textbf{overall score: 85}

score1: 20/20

score2: 15/30

score3: 30/30

score4: 10/10

score5: 10/10
& \textbf{overall score: 85}

score1: 20/20

score2: 15/30

(reason: Missed "blood thinners to prevent further complications" and "avoid contraceptives")

score3: 30/30

score4: 10/10

score5: 10/10
\\
\hline
\end{tabular}
\label{tab:human_anntation_closure_3}
\end{table}

\begin{table}
\centering
\caption{Human Annotation for Differential Diagnosis (Case 1 - Case 3).}
\begin{tabular}{|p{0.7cm}|p{4.5cm}|p{4.5cm}|p{4.5cm}|}
\hline
\textbf{Case} & \textbf{Expert 1} & \textbf{Expert 2} & \textbf{Expert 3} \\
\hline
1 & 
diagnosis 1 name: 0

historical finding:0 

physical finding: 0\newline

diagnosis 2 name: 10

historical finding: 1

physical finding: N/A\newline

diagnosis 3 name: 0

historical finding:0 

physical finding: 0\newline

order points: 0\newline

\textbf{total score: 11/52=21.15\%}

\textbf{quality score: 1}
& 
diagnosis 1 name: 0

historical finding:0 

physical finding: 0\newline

diagnosis 2 name: 8

historical finding: 3

physical finding: N/A\newline

diagnosis 3 name: 0

historical finding:0 

physical finding: 0\newline

order points: 0\newline

\textbf{total score: 11/52=21.15\%}

\textbf{quality score: 1}
& 
diagnosis 1 name: 0

historical finding:0 

physical finding: N/A\newline

diagnosis 2 name: 8

historical finding: 2

physical finding: N/A\newline

diagnosis 3 name: 10

historical finding:1

physical finding: N/A\newline

order points: 0\newline

\textbf{total score: 21/52=40.38\%}

\textbf{quality score: 2}
\\
\hline
2 & 
diagnosis 1 name: 10

historical finding: 2

physical finding: N/A\newline

diagnosis 2 name: 3

historical finding: 0

physical finding: N/A\newline

diagnosis 3 name: 10

historical finding: 2 

physical finding: N/A\newline

order points: 0\newline

\textbf{total score: 27/49=55.10\%}

\textbf{quality score: 2}
& 
diagnosis 1 name: 8

historical finding: 2 

physical finding: 0\newline

diagnosis 2 name: 5

historical finding: 0

physical finding: N/A\newline

diagnosis 3 name: 9

historical finding: 3 

physical finding: 0\newline

order points: 0\newline

\textbf{total score: 27/49=55.10\%}

\textbf{quality score: 3}
& 
diagnosis 1 name: 10

historical finding: 3

physical finding: N/A\newline

diagnosis 2 name: 0

historical finding: 0

physical finding: N/A\newline

diagnosis 3 name: 10

historical finding: 2

physical finding: N/A\newline

order points: 0\newline

\textbf{total score: 25/49=51.02\%}

\textbf{quality score: 3}
\\
\hline
3 & 
diagnosis 1 name: 10

historical finding: 3

physical finding: N/A\newline

diagnosis 2 name: 10

historical finding: 2

physical finding: 1\newline

diagnosis 3 name: 5

historical finding: 0 

physical finding: 0\newline

order points: 10\newline

\textbf{total score: 41/50=82.00\%}

\textbf{quality score: 4}
& 
diagnosis 1 name: 9

historical finding: 3 

physical finding: 0\newline

diagnosis 2 name: 10

historical finding: 2

physical finding: 3\newline

diagnosis 3 name: 0

historical finding: 0

physical finding: 0\newline

order points: 10\newline

\textbf{total score: 37/50=74.00\%}

\textbf{quality score: 4}
& 
diagnosis 1 name: 10

historical finding: 3

physical finding: N/A\newline

diagnosis 2 name: 10

historical finding: 3

physical finding: 1\newline

diagnosis 3 name: 0

historical finding: 0

physical finding: N/A\newline

order points: 10\newline

\textbf{total score: 37/50=74.00\%}

\textbf{quality score: 4}
\\
\hline
\end{tabular}
\label{tab:human_anntation_diagnosis}
\end{table}

\begin{table}
\centering
\caption{Human Annotation for Differential Diagnosis (Case 4 - Case 6).}
\begin{tabular}{|p{0.7cm}|p{4.5cm}|p{4.5cm}|p{4.5cm}|}
\hline
\textbf{Case} & \textbf{Expert 1} & \textbf{Expert 2} & \textbf{Expert 3} \\
\hline
4 & 
diagnosis 1 name: 5

historical finding: 2

physical finding: N/A\newline

diagnosis 2 name: 10

historical finding: 2

physical finding: 1\newline

diagnosis 3 name: 0

historical finding: 0 

physical finding: 0\newline

order points: 0\newline

\textbf{total score: 20/55=36.36\%}

\textbf{quality score: 2}
& 
diagnosis 1 name: 5

historical finding: 0 

physical finding: 0\newline

diagnosis 2 name: 10

historical finding: 3

physical finding: 3\newline

diagnosis 3 name: 0

historical finding: 0 

physical finding: 0\newline

order points: 0\newline

\textbf{total score: 21/55=38.18\%}

\textbf{quality score: 2}
& 
diagnosis 1 name: 5

historical finding: 0

physical finding: 0\newline

diagnosis 2 name: 10

historical finding: 2

physical finding: 2\newline

diagnosis 3 name: 0

historical finding: 0

physical finding: 0\newline

order points: 0\newline

\textbf{total score: 19/55=34.55\%}

\textbf{quality score: 3}
\\
\hline
5 & 
diagnosis 1 name: 10

historical finding: 3

physical finding: N/A\newline

diagnosis 2 name: 0

historical finding: 0

physical finding: 0\newline

diagnosis 3 name: 10

historical finding: 2 

physical finding: N/A\newline

order points: 10\newline

\textbf{total score: 35/50=70.00\%}

\textbf{quality score: 3}
& 
diagnosis 1 name: 10

historical finding: 3 

physical finding: 0\newline

diagnosis 2 name: 10

historical finding: 2

physical finding: 0\newline

diagnosis 3 name: 0

historical finding: 1

physical finding: 0\newline

order points: 10\newline

\textbf{total score: 36/50=72.00\%}

\textbf{quality score: 3}
& 
diagnosis 1 name: 10

historical finding: 3

physical finding: N/A\newline

diagnosis 2 name: 0

historical finding: 0

physical finding: N/A\newline

diagnosis 3 name: 10

historical finding: 3

physical finding: N/A\newline

order points: 10\newline

\textbf{total score: 36/50=72.00\%}

\textbf{quality score: 4}
\\
\hline
6 & 
diagnosis 1 name: 10

historical finding: 2

physical finding: N/A\newline

diagnosis 2 name: 10

historical finding: 1

physical finding: N/A\newline

diagnosis 3 name: 0

historical finding: 0

physical finding: N/A\newline

order points: 10\newline

\textbf{total score: 33/49=67.35\%}

\textbf{quality score: 2}
& 
diagnosis 1 name: 10

historical finding: 3 

physical finding: N/A\newline

diagnosis 2 name: 10

historical finding: 1

physical finding: N/A\newline

diagnosis 3 name: 0

historical finding: 0

physical finding: 0\newline

order points: 10\newline

\textbf{total score: 34/49=69.39\%}

\textbf{quality score: 3}
& 
diagnosis 1 name: 10

historical finding: 1

physical finding: N/A\newline

diagnosis 2 name: 10

historical finding: 1

physical finding: N/A\newline

diagnosis 3 name: 0

historical finding: 0

physical finding: N/A\newline

order points: 10\newline

\textbf{total score: 32/49=65.31\%}

\textbf{quality score: 4}
\\
\hline
\end{tabular}
\label{tab:human_anntation_diagnosis_2}
\end{table}

\begin{table}
\centering
\caption{Human Annotation for Differential Diagnosis (Case 7 - Case 9).}
\begin{tabular}{|p{0.7cm}|p{4.5cm}|p{4.5cm}|p{4.5cm}|}
\hline
\textbf{Case} & \textbf{Expert 1} & \textbf{Expert 2} & \textbf{Expert 3} \\
\hline
7 & 
diagnosis 1 name: 5

historical finding: 1

physical finding: N/A\newline

diagnosis 2 name: 0

historical finding: 0

physical finding: N/A\newline

diagnosis 3 name: 10

historical finding: 1

physical finding: 1\newline

order points: 0\newline

\textbf{total score: 18/49=36.73\%}

\textbf{quality score: 2}
& 
diagnosis 1 name: 5

historical finding: 3

physical finding: 3\newline

diagnosis 2 name: 0

historical finding: 0

physical finding: 0\newline

diagnosis 3 name: 10

historical finding: 3 

physical finding: 3\newline

order points: 0\newline

\textbf{total score: 27/49=55.10\%}

\textbf{quality score: 2}
& 
diagnosis 1 name: 5

historical finding: 0

physical finding: N/A\newline

diagnosis 2 name: 10

historical finding: 1

physical finding: 3\newline

diagnosis 3 name: 10

historical finding: 1

physical finding: N/A\newline

order points: 0\newline

\textbf{total score: 30/49=61.22\%}

\textbf{quality score: 3}
\\
\hline
8 & 
diagnosis 1 name: 10

historical finding: 2

physical finding: 1\newline

diagnosis 2 name: 8

historical finding: 1

physical finding: N/A\newline

diagnosis 3 name: 10

historical finding: 1 

physical finding: N/A\newline

order points: 0\newline

\textbf{total score: 33/50=66.00\%}

\textbf{quality score: 3}
& 
diagnosis 1 name: 10

historical finding: 3 

physical finding: 3\newline

diagnosis 2 name: 10

historical finding: 3

physical finding: N/A\newline

diagnosis 3 name: 10

historical finding: 1

physical finding: N/A\newline

order points: 0\newline

\textbf{total score: 40/50=80.00\%}

\textbf{quality score: 3}
& 
diagnosis 1 name: 10

historical finding: 2

physical finding: 1\newline

diagnosis 2 name: 5

historical finding: 1

physical finding: N/A\newline

diagnosis 3 name: 10

historical finding: 1

physical finding: N/A\newline

order points: 0\newline

\textbf{total score: 30/50=60.00\%}

\textbf{quality score: 3}
\\
\hline
9 & 
diagnosis 1 name: 0

historical finding: 0

physical finding: N/A\newline

diagnosis 2 name: 0

historical finding: 0

physical finding: N/A\newline

diagnosis 3 name: 0

historical finding: 0

physical finding: N/A\newline

order points: 0\newline

\textbf{total score: 0/52=00.00\%}

\textbf{quality score: 1}
& 
diagnosis 1 name: 3

historical finding: 3 

physical finding: 3\newline

diagnosis 2 name: 0

historical finding: 0

physical finding: 0\newline

diagnosis 3 name: 0

historical finding: 0

physical finding: 0\newline

order points: 0\newline

\textbf{total score: 9/52=17.31\%}

\textbf{quality score: 2}
& 
diagnosis 1 name: 0

historical finding: 0

physical finding: N/A\newline

diagnosis 2 name: 0

historical finding: 0

physical finding: N/A\newline

diagnosis 3 name: 0

historical finding: 0

physical finding: N/A\newline

order points: 0\newline

\textbf{total score: 0/52=00.00\%}

\textbf{quality score: 1}
\\
\hline
\end{tabular}
\label{tab:human_anntation_diagnosis_3}
\end{table}

\begin{table}
\centering
\caption{Human Annotation for Differential Diagnosis (Case 10).}
\begin{tabular}{|p{0.7cm}|p{4.5cm}|p{4.5cm}|p{4.5cm}|}
\hline
\textbf{Case} & \textbf{Expert 1} & \textbf{Expert 2} & \textbf{Expert 3} \\
\hline
10 & 
diagnosis 1 name: 10

historical finding: 3

physical finding: 1\newline

diagnosis 2 name: 10

historical finding: 2

physical finding: 1\newline

diagnosis 3 name: 3

historical finding: 10

physical finding: 0\newline

order points: 10\newline

\textbf{total score: 40/55=72.73\%}

\textbf{quality score: 3}
& 
diagnosis 1 name: 10

historical finding: 3

physical finding: 3\newline

diagnosis 2 name: 10

historical finding: 3

physical finding: 2\newline

diagnosis 3 name: 0

historical finding: 0

physical finding: 0\newline

order points: 10\newline

\textbf{total score: 41/55=74.55\%}

\textbf{quality score: 4}
& 
diagnosis 1 name: 10

historical finding: 3

physical finding: 2\newline

diagnosis 2 name: 10

historical finding: 2

physical finding: 1\newline

diagnosis 3 name: 5

historical finding: 0

physical finding: 0\newline

order points: 10\newline

\textbf{total score: 43/55=78.18\%}

\textbf{quality score: 4}
\\
\hline
\end{tabular}
\label{tab:human_anntation_diagnosis_4}
\end{table}

\twocolumn
\section{Details for Experimental Settings}
\label{apx:sec:experimental_settings}

\paragraph{Factuality metrics: UMLS-F1}
The assessment of factual accuracy in LLMs output leverages the UMLS concept overlap metric. The Unified Medical Language System (UMLS), established by~\cite{bodenreider2004unified}, significantly contributes to the biomedical domain's interoperability. It achieves this by amalgamating and disseminating a comprehensive collection of biomedical terminologies, classification systems, and coding standards from many sources. By doing so, UMLS aids in reconciling semantic variances and representational disparities found across different biomedical concept repositories.

For the identification and alignment of medical named entities within texts to their corresponding biomedical concepts in UMLS, we employed the Scispacy library~\footnote{We used the Scispacy \textit{en\_core\_sci\_lg} model.}.
Scispacy excels in identifying and clarifying entities, thus facilitating the accurate association of named entities found in LLMs output with the relevant UMLS concepts. 
This capability is critical for evaluating the LLMs output's factual accuracy.

The analytical process for LLMs output utilizes metrics of precision and recall. Precision represents the ratio of concepts present in both the LLM output and ground truth content, serving as a measure of the LLM output's factual correctness. In contrast, recall evaluates how well the information in the LLM output matches the intended content, reflecting the relevance of the presented information.

To calculate these metrics, we consider the concept sets from both the ground truth ($C_{ref}$) and the LLM output ($C_{gen}$). The formulas for recall and precision are as follows:

$$
\text{Recall} = \frac{|C_{ref} \cap C_{gen}|}{|C_{ref}|}
$$

$$
\text{Precision} = \frac{|C_{ref} \cap C_{gen}|}{|C_{gen}|}.
$$

The F1 score, derived from the above precision and recall values, is reported to provide a balanced measure of LLMs output's accuracy and relevance.

\section{More details for Experiment results}
\label{apx:sec:results}
\subsection{MedExamLLM}

\begin{table*}[H]
\centering
\caption{Pearson correlation (p-value) between expert evaluation (average) and 1. different LLMs' MedExamLLM output (LLM-as-Judge) 2. some traditional metrics used in clinical generation tasks.}
\scalebox{1}{ 
\begin{tabular}{l|cccc}
\hline
\text{Pearson (p-value)} & \text{InfoGatherQA} & \text{Physical Exam} & \text{Closure} & \text{Diagnosis} \\
\hline
\text{GPT-4o} & 0.82 (<0.005) & 0.80(<0.01) & 0.76(<0.05) & 0.71(<0.05) \\
\text{GPT-4} & 0.9 (<0.0005) & 0.92(<0.005) & 0.47(<0.1) & 0.78(<0.01) \\
\text{GPT-3.5} & -0.25 (<0.5) & -0.14(<1) & 0.25(<0.5) & -0.05(<1) \\
\text{Claude3-Opus} & 0.78 (<0.01) & 0.82(<0.005) & 0.75(<0.05) & 0.64(<0.05)  \\
\text{Claude3-Sonnet} & 0.52 (<0.5) & 0.75(<0.05) & -0.09(<1) & 0.41(<0.5)  \\
\text{Claude3-haiku} & 0.048 (<1) & 0.36(<0.5) & -0.02(<1) & 0.43(<0.5)  \\
\hline
\text{ROUGE-1~\cite{lin2004rouge}} & 0.67 (<0.05) & 0.52(<0.5) & 0.16(<1) & 0.02(<1) \\
\text{ROUGE-2~\cite{lin2004rouge}} & 0.70 (<0.05) & 0.33(<0.5) & 0.04(<1) & 0.17(<1) \\
\text{ROUGE-L~\cite{lin2004rouge}} & 0.65 (<0.05) & 0.45(<0.5) & 0.28(<0.5) & -0.02(<1) \\
\text{METEOR~\cite{banerjee2005meteor}} & 0.62 (<0.1) & 0.72(<0.05) & -0.07(<1) & 0.05(<1) \\
\text{BERTScore~\cite{zhang2019bertscore}} & 0.86 (<0.005) & 0.28(<0.5) & 0.23(<1) & 0.03(<1) \\
\text{Exact Match (appendix~\ref{apx:sec:experimental_settings})} & - & 0.35(<0.5) & - & 0.19(<1) \\
\text{UMLS-F (appendix~\ref{apx:sec:experimental_settings})} & 0.65 (<0.05) & 0.63(<0.1) & 0.35(<0.5) & 0.25(<0.5) \\
\hline
\end{tabular}
}
\label{tab:human_ai_corr_pearson}
\end{table*}

\begin{table*}[H]
\centering
\caption{Kendall's $\tau$ (p-value) between expert evaluation (average) and 1. different LLMs' MedExamLLM output (LLM-as-Judge) 2. some traditional metrics used in clinical generation tasks.}
\scalebox{1}{ 
\begin{tabular}{l|cccc}
\hline
\text{Kendall's $\tau$ (p-value)} & \text{InfoGatherQA} & \text{Physical Exam} & \text{Closure} & \text{Diagnosis} \\
\hline
\text{GPT-4o} & 0.64 (<0.01) & 0.38(0.5) & 0.37(<0.5) & 0.56(<0.05) \\
\text{GPT-4} & 0.78 (<0.001) & 0.53(<0.05) & 0.47(<0.05) & 0.69(<0.005) \\
\text{GPT-3.5} & -0.07 (<1) & -0.56(<0.1) & 0.13(<1) & 0.11(<1) \\
\text{Claude3-Opus} & 0.63 (<0.05) & 0.35(<0.5) & 0.25(<0.5) & 0.56(<0.05)  \\
\text{Claude3-Sonnet} & 0.33 (<0.5) & 0.40(<0.5) & -0.12(<1) & 0.29(<0.5)  \\
\text{Claude3-haiku} & 0.045 (<1) & 0.12(<1) & 0.23(<0.5) & 0.29(<0.5)  \\
\hline
\text{ROUGE-1~\cite{lin2004rouge}} & 0.56 (<0.05) & 0.18(<0.5) & 0.14(<0.5) & -0.07(<1) \\
\text{ROUGE-2~\cite{lin2004rouge}} & 0.6 (<0.05) & 0.38(<0.5) & 0.07(<1) & 0.24(<0.5) \\
\text{ROUGE-L~\cite{lin2004rouge}} & 0.6 (<0.05) & 0.07(<1) & 0.35(<0.5) & -0.02(<1) \\
\text{METEOR~\cite{banerjee2005meteor}} & 0.47 (<0.1) & 0.46(<0.1) & -0.35(<0.5) & 0.07(<1) \\
\text{BERTScore~\cite{zhang2019bertscore}} & 0.56 (<0.05) & 0.44(<0.1) & 0.05(<1) & 0.02(<1) \\
\text{Exact Match (appendix~\ref{apx:sec:experimental_settings})} & - & 0.25(<0.5) & - & 0.20(<0.5) \\
\text{UMLS-F (appendix~\ref{apx:sec:experimental_settings})} & 0.47 (<0.1) & 0.28(<0.5) & 0.54(<0.05) & 0.11(<1) \\
\hline
\end{tabular}
}
\label{tab:human_ai_corr_kendall}
\end{table*}

\subsection{MedStuLLM}
For InfoGatherQA, we only run all LLMs once since the data number for this section is much larger than other threes and we cannot afford to run it multiple times.
Detailed scores for GPT and Claude3 models, including fine-grained scores for each criterion, can be found in Table \ref{tab:gpt4_3.5_physical_exam}, \ref{tab:gpt4o_opus_physical_exam} and \ref{tab:haiku_sonnet_physical_exam}.
Detailed scores for GPT and Claude3 models, including fine-grained scores for each criterion, can be found in Table \ref{tab:gpt3.5_gpt4_closure}, \ref{tab:gpt4o_opus_closure} and \ref{tab:haiku_sonnet_closure}.
Detailed scores for GPT and Claude3 models, including fine-grained scores for each criterion, can be found in our GitHub.

\begin{table*}[H]
\centering
\caption{CS benchmarking 95\% CI results.  ’-’ means that LLM cannot follow that section’s instruction to generate valid output.}
\scalebox{0.92}{
\begin{tabular}{lccc}
\hline
MedStuLLM & Physical exam & Closure & Diagnosis \\
\hline
GPT 4 & (43.99, 59.34) & (73.08, 85.58) & (37.56, 58.37) \\
GPT 4-o	& (61.27, 66.27) & (84.23, 87.10) & (47.98, 59.25) \\
GPT 3.5 & (39.03, 45.64) & (70.54, 78.13) & (28.97, 40.94) \\
Claude Haiku & (53.06, 56.54) & (81.80, 87.54) & (45.39, 52.98) \\
Claude Sonnet & (51.48, 55.72) & (81.52, 86.48) & (44.16, 53.08) \\
Claude Opus & (51.70, 60.03) & (80.90, 83.77) & (46.36, 57.06) \\
LLAMA2-7b & (25.85, 35.55) & - & (-1.11, 4.32) \\
LLAMA3-8b & - & - & (34.91, 40.60) \\
Mistral-7b & (48.92, 51.08) & (75.91, 78.67) & (20.03, 54.73) \\
BioLLAMA3-8b & (-1.197, 9.93) & - & (26.18, 49.53) \\
BioMistral-7b & (4.55, 14.18) & - & (41.41, 47.93) \\
GLM4-9b & (59.41, 66.32) & (76.73, 86.36) & (31.11, 40.09) \\
LLAMA2-70b & - & - & (29.81, 35.42) \\
LLAMA3-70b & - & - & (29.85, 53.35) \\
BioLLAMA3-70b & (39.14, 42.99) & - & (25.54, 45.16) \\
Mixtral -8x7b & - & - & (32.34, 55.58) \\
LLAMA3-8b-dpo & (36.88, 54.92) & (60.21, 78.06) & - \\
LLAMA3-8b-ipo & (27.84, 54.82) & (69.10, 79.52) & (40.59, 47.91) \\
LLAMA3-8b-kto & (15.45, 52.55) & (70.99, 75.46) & (44.05, 51.53) \\
LLAMA3-8b-simpo & (31.79, 43.08) & (66.71, 73.51) & (37.09, 41.54) \\
LLAMA3-8b-rdpo & (39.01, 67.39) & (67.55, 71.29) & - \\
LLAMA3-8b-orpo & (24.39, 34.94) & (3.54, 23.79) & (35.53, 48.01) \\
Mistral-7b-dpo & (42.45, 54.62) & (77.07, 81.79) & (39.64, 52.01) \\
Mistral-7b-ipo & (36.34, 40.46) & (72.20, 76.94) & (33.67, 49.38) \\
Mistral-7b-kto & (34.87, 41.93) & (78.36, 82.67) & (32.94, 43.03) \\
Mistral-7b-simpo & (42.08, 43.78) & (74.80, 80.03) & (34.20, 45.33) \\
Mistral-7b-rdpo & (42.02, 55.11) & (72.20, 79.91) & (36.75, 43.18) \\
Mistral-7b-orpo & (27.37, 46.03) & (61.90, 74.19) & (38.68, 52.16) \\
Qwen2-0.5b & (-13.44, -5.83) & - & (10.80, 11.84) \\
Qwen2-1.5b & (-27.56, -13.11) & (7.44, 10.11) & (18.54, 32.88) \\
Qwen2-7b & (44.39, 54.08) & (71.66, 72.40) & (29.24, 38.64) \\
Qwen2-moe-57b & (40.25, 48.22) & (81.58, 82.47) & (40.44, 51.66) \\
Qwen2-72b & (42.27, 52.13) & (84.86, 86.25) & (38.35, 55.69) \\

\hline
\end{tabular}
}
\label{tab:95CI_results}
\end{table*}

\section{Case study by section}
\label{apx:sec:case_study}
\subsection{InfoGatherQA}
\subsubsection{Human Annotation Case Study}
InfoGatherQA evaluates two crucial criteria: consistency and originality. To ensure reliable assessment, we engaged three experts to annotate 10 different cases. Our analysis revealed a strong correlation among the experts' average performance scores, demonstrating the robustness of our evaluation framework, as can seen in table \ref{tab:human_ai_corr}.

Notably, experts 1 and 2 exhibited the highest agreement as seen in table \ref{tab:human_corr}, with a Pearson correlation of 0.89, while experts 2 and 3 exhibited the least agreement, with a Pearson correlation of 0.77. To understand the reasons behind these differences, we discussed the annotations with the experts.
It became evident that while assessing originality is relatively straightforward to assess, evaluating a question's consistency can be subjective, dependent on the examiner's leniency. For instance, in case 3, conversation iteration 4, experts 1 and 2 assigned a score of 1, while expert 3 gave a score of 0. The question stated, "Has the back pain been getting worse, staying the same, or improving since it started?". There was no exact match with the ground truth questions. Instead, the related questions in the ground truth were "Does anything make your pain better?" and "Does anything make it worse?". This case highlights the nuanced nature of consistency evaluation and the potential for varying interpretations.

By engaging multiple experts and analyzing their assessments, we have gained valuable insights into the challenges and nuances associated with evaluating consistency and originality in InfoGatherQA. With a large evaluation data sample comprising 343 distinct questions, the evaluation process is susceptible to human errors, such as overlooking relevant information or inconsistent application of the criteria. These errors could contribute to the observed differences in expert assessments. By utilizing multiple human evaluators, we effectively mitigate the impact of individual biases and ensure a more comprehensive and reliable evaluation.

These findings underscore the importance of clear evaluation criteria and the need for continuous refinement to ensure fair and reliable assessments.

\begin{enumerate}
    \item \textbf{Example Case History:}
    \begin{description}[style=nextline]
        \item[Opening Scenario:] Rick Meyer, a 51-year-old male construction worker, comes to the office complaining of back pain.
        \item Vital Signs:
        \begin{itemize}
            \item BP: 120/85 mmHg
            \item Temp: 98.2°F (36.8°C)
            \item RR: 20/minute
            \item HR: 50/minute, regular
        \end{itemize}
        \item Examinee Tasks:
        \begin{enumerate}[label=(\alph*)]
            \item Take a focused history.
            \item Perform a focused physical exam (do not perform rectal, genitourinary, or felted breast exam).
            \item Explain your clinical impression and workup plan to the patient.
            \item Write the patient note after leaving the room.
        \end{enumerate}
    \end{description}
    
    \item \textbf{Chat History:}
    \begin{description}[style=nextline]
        \item[Topic: Chief Complaint]
        \begin{description}[style=nextline]
            \item[Doctor:] "What brings you to the clinic today?"
            \item[Patient:] "Pain in my back."
        \end{description}
        \vspace{0.5em}
        \item[Topic: Onset]
        \begin{description}[style=nextline]
            \item[Doctor:] "Since when have you been experiencing back pain?"
            \item[Patient:] "One week ago."
        \end{description}
        \vspace{0.5em}
        \item[Topic: Updated/Precipitating Events]
        \begin{description}[style=nextline]
            \item[Doctor:] "Please tell me more about it, Mr. Meyer."
            \item[Patient:] "I was lifting some heavy boxes; then my back started hurting right away."
        \end{description}
    \end{description}

    \item \textbf{InfoGatherQA Ground Truth (closest matches):}
    \begin{itemize}
        \item "Does anything make your pain better?"
        \item "Does anything make it worse?"
    \end{itemize}

    \item \textbf{MedStuLLM for InfoGatherQA:}
    \begin{description}[style=nextline]
        \item[Question:] "Has the back pain been getting worse, staying the same, or improving since it started?"
        \item[Reason:] I would like to know whether Rick's back pain is getting worse, staying the same, or improving since it began.
    \end{description}

    \item \textbf{Human Annotation:}
    \begin{itemize}
        \item \textbf{Expert 1:} Score: 1
        \item \textbf{Expert 2:} Score: 1
        \item \textbf{Expert 3:} Score: 0
    \end{itemize}
\end{enumerate}

\subsubsection{LLM-as-cs-examiner error analysis}
From the results in Table~\ref{tab:LLM-as-medical student for InfoGatherQA}, it appears that GPT-4 demonstrates the best performance with a performance score of 0.6279 for the LLM-as-medical student task on InfoGatherQA, with GPT-4o closely following in second place with a performance score of 0.6212. 

Although GPT-4o is a more advanced language model than GPT-4, it does not demonstrate superior performance in this specific task. The models are ranked as follows based on their performance: Claude 3 Opus, Claude 3 Sonnet, GPT-3.5, and finally, Claude 3 Haiku. This ranking indicates that larger models generally follow instructions better and generate questions more closely aligned with USMLE-provided ground-truth questions. While model size influences performance on the InfoGatherQA task, larger models like GPT-4 and GPT-4o exhibit diminishing returns. It's noteworthy that the performance gap between these top-performing models is relatively small, indicating comparable capabilities in this specific task. However, smaller LLMs like GPT 3.5 and Claude 3 Haiku often ask questions already covered in the chat history, failing the originality criteria. This error could stem from their limited context length and lack of understanding to follow detailed instructions.

Another limitation is the repetition in question generation. Each question generation is independent and only considers the opening and chat history, leading to numerous instances of repeated questions. For example, in case 1, iterations 2 and 3 involve the LLM as a medical student asking the same question, "When did you first notice the blood in your urine?" This lack of memory regarding previously generated questions reduces diversity in the generations. Moreover, if the repeated question is incorrect, it can result in multiple incorrect responses within the same case.


\begin{table*}[H]
\caption{Repetition in Generated Questions}
\centering
\begin{tabular}{|l|p{2cm}|l|p{10cm}|}
\hline
\textbf{Turn} & \textbf{Question} & \textbf{Score} & \textbf{Chat history} \\
\hline
2 & When did you first notice the blood in your urine? & 1 & \textbf{Previous conversation:} \newline  \newline Patient Response: \newline \textbf{Chief complaint:} \newline Doctor: "What brings you in today?" \newline Patient response: I have blood in my urine, doctor. \\
\hline
3 & When did you first notice blood in your urine? & 1 & \textbf{Previous conversation:} \newline \newline Patient Response: \newline \textbf{Chief complaint:} \newline Doctor: "What brings you in today?" \newline Patient response: I have blood in my urine, doctor. \newline \textbf{Description:} \newline Doctor: "Please describe it to me, Mr. Fisher?" \newline Patient response: It was bright red and later had some clots. \\
\hline
\end{tabular}
\end{table*}

\subsubsection{LLM-as-medical-student}

The performance comparison can be summarized as:
\text{GPT-4} $\approx$ \text{GPT-4o} > \text{Claude 3 Opus} >\text{Claude 3 Sonnet} > \text{GPT-3.5} > \text{Claude 3 Haiku}

\begin{table*}[!ht]
\centering
\caption{LLM-as-medical-student for InfoGatherQA}
\label{tab:LLM-as-medical student for InfoGatherQA}
\scalebox{1}{ 
\begin{tabular}{lrrrrrr}
\toprule
Cases & gpt-4o & gpt4 & gpt3.5 & claude-3-opus & claude-3-sonnet & claude-3-haiku \\
\midrule
1 & 0.9730 & 0.8919 & 0.7027 & 0.8378 & 0.8649 & 0.6486 \\
2 & 0.7097 & 0.4516 & 0.1935 & 0.6452 & 0.6452 & 0.2581 \\
3 & 0.5769 & 0.8462 & 0.3077 & 0.8077 & 0.7692 & 0.3846 \\
4 & 0.1515 & 0.3939 & 0.0606 & 0.6364 & 0.3636 & 0.1212 \\
5 & 0.7027 & 0.6216 & 0.4595 & 0.7568 & 0.6486 & 0.3784 \\
6 & 0.9189 & 0.6486 & 0.3514 & 0.7297 & 0.7838 & 0.2162 \\
7 & 0.2222 & 0.3704 & 0.3333 & 0.2593 & 0.1852 & 0.1111 \\
8 & 0.5581 & 0.5349 & 0.5581 & 0.5581 & 0.7442 & 0.6512 \\
9 & 0.8718 & 0.7436 & 0.7179 & 0.5385 & 0.7436 & 0.3590 \\
10 & 0.4545 & 0.6061 & 0.4242 & 0.7879 & 0.6061 & 0.3636 \\
11 & 0.8750 & 0.7500 & 0.4688 & 0.9062 & 0.8750 & 0.3438 \\
12 & 0.5769 & 0.6923 & 0.3077 & 0.6154 & 0.6923 & 0.3077 \\
13 & 0.7097 & 0.9355 & 0.5806 & 0.6452 & 0.7097 & 0.4516 \\
14 & 0.8718 & 0.8205 & 0.5385 & 0.4872 & 0.4872 & 0.7436 \\
15 & 0.5897 & 0.4872 & 0.2564 & 0.6667 & 0.4872 & 0.2051 \\
16 & 0.2821 & 0.1795 & 0.1538 & 0.1795 & 0.2051 & 0.2051 \\
17 & 0.4375 & 0.4688 & 0.3438 & 0.6875 & 0.6250 & 0.0938 \\
18 & 0.6061 & 0.4545 & 0.2121 & 0.5152 & 0.3333 & 0.2424 \\
19 & 0.4444 & 0.4815 & 0.1481 & 0.3704 & 0.5185 & 0.1852 \\
20 & 0.6774 & 0.5806 & 0.7097 & 0.8710 & 0.6129 & 0.3226 \\
21 & 0.9302 & 0.7907 & 0.8140 & 0.8837 & 0.8837 & 0.5814 \\
22 & 0.7188 & 0.6250 & 0.5625 & 0.6250 & 0.3750 & 0.2812 \\
23 & 0.1739 & 0.6087 & 0.1087 & 0.4565 & 0.3043 & 0.3043 \\
24 & 0.5862 & 0.8276 & 0.5172 & 0.6207 & 0.4828 & 0.7586 \\
25 & 0.3750 & 0.4375 & 0.0938 & 0.2500 & 0.0938 & 0.3125 \\
26 & 0.8824 & 0.8235 & 0.3529 & 0.8529 & 0.5588 & 0.3529 \\
27 & 0.8636 & 0.7955 & 0.6591 & 0.6591 & 0.4773 & 0.3636 \\
28 & 0.5278 & 0.5000 & 0.1111 & 0.4444 & 0.3611 & 0.1944 \\
29 & 0.8000 & 0.6750 & 0.4500 & 0.7250 & 0.3000 & 0.3500 \\
30 & 0.2188 & 0.6250 & 0.3125 & 0.3438 & 0.2812 & 0.1562 \\
31 & 0.7500 & 0.7045 & 0.4545 & 0.5227 & 0.4545 & 0.6591 \\
32 & 0.4103 & 0.4872 & 0.4872 & 0.5897 & 0.2308 & 0.1795 \\
33 & 0.6571 & 0.4857 & 0.5714 & 0.7429 & 0.3429 & 0.3714 \\
34 & 0.4750 & 0.3750 & 0.4750 & 0.5250 & 0.2000 & 0.2000 \\
35 & 0.7750 & 0.7750 & 0.4250 & 0.7500 & 0.3750 & 0.2000 \\
36 & 0.9268 & 0.6341 & 0.2195 & 0.7805 & 0.5610 & 0.3902 \\
37 & 0.8857 & 0.7143 & 0.1143 & 0.7714 & 0.3429 & 0.2857 \\
38 & 0.3500 & 0.8000 & 0.3500 & 0.4500 & 0.1500 & 0.1000 \\
39 & 0.5588 & 0.7941 & 0.3824 & 0.7353 & 0.4706 & 0.3824 \\
40 & 0.6176 & 0.6176 & 0.3824 & 0.5294 & 0.4118 & 0.2794 \\
41 & 0.7333 & 0.5000 & 0.4000 & 0.5000 & 0.2000 & 0.2000 \\
42 & 0.6875 & 0.5938 & 0.2188 & 0.5000 & 0.2500 & 0.2500 \\
43 & 0.3684 & 0.6316 & 0.1579 & 0.2632 & 0.3421 & 0.5263 \\
44 & 0.8485 & 0.8485 & 0.7576 & 0.9394 & 0.1818 & 0.4545 \\ \hline
\textbf{Average} & 0.6212 & 0.6279 & 0.3911 & 0.6128 & 0.4666 & 0.3347 \\
\bottomrule
\end{tabular}
}
\end{table*}

\subsection{Physical Exams}
\subsubsection{Human Annotation Case Study}
There are three key dimensions for Physical Exams Examination: Exam Coverage, Reason Relevance and Accuracy, and Extra Exams Penalty. Three experts were engaged to annotate 10 different cases, with detailed results provided in the table \ref{tab:human_anntation_physical_exam} and table \ref{tab:human_anntation_physical_exam_2}.

Our analysis revealed a high correlation among the experts in the dimensions of Exam Coverage, and Reason Relevance and Accuracy. However, the Extra Exams Penalty dimension exhibited significant discrepancies in the experts' judgments. Specifically, Expert 2 did not penalize any case under this dimension. Expert 1 assigned a total of 70 penalty points across the 10 cases. Expert 3 assigned 40 penalty points in total.

To understand the reasons behind these differences, we discussed the annotations with the experts. It became clear that the judgments on Extra Exams Penalty were highly subjective and dependent on the individual expert's clinical experience and judgment.

For instance, as the following example case shown, Expert 2 believed that "the patient reported left chest pain, it makes sense to check the chest and heart.", Expert 3 found that "a chest exam is make sense, but a heart exam is not necessary." But Expert 1 argued that after examining the Pulmonary and Abdomen, there was no need to further examine the chest and heart.

Thus, given the subjective nature of the Extra Exams Penalty dimension and the significant differences in expert judgments, we decided to exclude this dimension from the final agreement calculations. This approach ensures a more objective and consistent assessment of the annotations.

\begin{enumerate}
    \item Example Case History:
    \begin{itemize}
        \item HPI: A 25-year-old male complains of left chest pain and LUQ pain following an MVA. The patient struck a tree with his car at a slow speed. The chest pain is 8/10. It is exacerbated with movement or when he takes a deep breath, and nothing relieves it. He reports dyspnea and a productive cough with a low-grade fever but denies LOC, headache, change in mental status, or change in vision. No cardiovascular or neurologic symptoms. No nausea, vomiting, neck stiffness, or unusual fluid from the mouth or nose. No dysuria. His last meal was 5 hours ago. He denies being under the influence of alcohol or drugs.
        \item ROS: As per HPI.
        \item Allergies: NKDA.
        \item Medications: None.
        \item PMH: Infectious mononucleosis 2 months ago.
        \item PSH: None. 
        \item SH: No smoking, occasional EtOH, no illicit drugs. 
        \item FH: Noncontributory. 
    \end{itemize}

    \item Physical Exams Ground Truth:
    \begin{itemize}
        \item Head and neck exam: Inspection, palpation, auscultation
        \item Pulmonary exam: Inspection, auscultation, palpation, percussion
        \item Abdominal exam: Inspection, auscultation, palpation (examined specifically for organomegaly such as splenomegaly)
        \item H8 exam: Mental status, cranial nerves, gross motor function, sensation
        \item Skin exam: Inspection for abrasions, lacerations, bruising
    \end{itemize}

    \item MedStuLLM for Physical Exams:
    \begin{itemize}
        \item Chest: Palpation and Auscultation.
        
        reason: The patient is experiencing sharp pain in the chest, particularly on left side, and difficulty breathing, which needs to be physically examined for signs of injury or illness.
        
        \item Pulmonary: Auscultation
        
        reason: The patient has been coughing and bringing up yellow mucus, indicating possible respiratory issues.
        \item Abdomen: Palpation
        
        reason: The patient is experiencing sharp pain in their left upper abdominal quadrant, indicating possible internal injury or illness.
        \item Heart: Auscultation
        
        reason: The patient is experiencing breathing problems, which can be associated with cardiovascular health.
        \item Neck: Inspection and palpation
        
        reason: The patient had a motor vehicle accident, so it's important to rule out any neck injuries (including in the carotid area) which could be causing the reported difficulties in breathing.
    \end{itemize}

    \item Human Annotation:
    \begin{itemize}
        \item \textbf{Expert 1: 56}
        
        score1(Exam Coverage): 3/5 * 60 = 36 (reason: Missed Neurologic exam and Skin exam)

        score2(Reason Relevance and Accuracy): 40

        score3(Extra Exams Penalty): -20 (reason: Chest and Heart are not required)
        \item \textbf{Expert 2: 76}
        
        score1(Exam Coverage): 3/5 * 60 = 36 (reason: Missed Neurologic exam and Skin exam)
        
        score2(Reason Relevance and Accuracy): 40
        
        score3(Extra Exams Penalty): 0
        \item \textbf{Expert 3: 66}
        
        score1(Exam Coverage): 3/5 * 60 = 36 (reason: Missed Neurologic exam and Skin exam)
        
        score2(Reason Relevance and Accuracy): 40
        
        score3(Extra Exams Penalty): -10 (reason: No heart test required)
    \end{itemize}
 \end{enumerate}

\subsubsection{LLM-as-cs-examiner error analysis}
The correlation of GPT-4 with the experts' scores is the highest among all models, indicating its strong alignment with expert judgment. When looking at the average scores, GPT-4o and Claude Opus show the closest match to the Experts' average scores, suggesting their overall performance is also consistent with human experts. Detailed scores are provided in the Table \ref{tab:performance scores of different LLMs as examiners on Physical Exams}.

On the other hand, GPT-3.5 and Claude Haiku exhibit unusually high scores, primarily due to inaccurate assessments of Score1 (exam coverage), leading to inflated scores across all cases.

\subsubsection{LLM-as-medical-student}
Analyzing the average scores from 44 cases as presented in Table\ref{tab:performance scores of different LLMs as medical student on Physical Exams}, and further detailed in Tables\ref{tab:gpt4_3.5_physical_exam}, \ref{tab:gpt4o_opus_physical_exam}, and \ref{tab:haiku_sonnet_physical_exam}, we observe a clear hierarchy among the models: GPT-4o outperforms all, followed by Claude Sonnet, Claude Haiku, Claude Opus, GPT-4, and finally GPT-3.5. 

We can examine the detailed scores and the content generated by each model (see Table \ref{tab:LLM-as-Medical-Student Output Examples for Physical Exams} and \ref{tab:LLM-as-Medical-Student Output Examples for Physical Exams 2}).
\begin{enumerate}
    \item Score1 (Exam Coverage) and Score3 (Extra Exam Penalty) reflect the models' abilities to summarize information and balance thoroughness with relevance. Some models achieve high scores in Score1 by increasing the breadth of exam coverage to improve accuracy. For instance, GPT-4o, Haiku, and Sonnet generate a higher number of exams per case compared to other models, often exceeding the ground truth. 
    \item Score2 (Reason Relevance and Accuracy): This score assesses the models' ability to provide coherent and comprehensive explanations. Opus scores highest at 86, demonstrating superior reasoning capabilities. This is followed closely by GPT-4, GPT-4o, and Sonnet, each scoring around 80. Haiku scores the lowest at 75. GPT-3.5’s anomalously high score in this category indicates a potential flaw in the scoring mechanism. GPT-3.5 typically generates only 1-2 direct exams, resulting in a limited number of reasons. These few reasons are evaluated more thoroughly, inflating the score disproportionately.
 \end{enumerate}

\begin{table*}
\centering
\begin{tabular}{lcccccccc}
\toprule
\textbf{Case} & \textbf{Experts} & \textbf{GPT 4} & \textbf{GPT 3.5} & \textbf{GPT 4-o} & \textbf{Claude 3 Opus} & \textbf{Sonnet} & \textbf{Haiku} \\
\midrule
1 & 70.0 & 58 & 88 & 64 & 64 & 88 & 84 \\
2 & 70.0 & 76 & 80 & 56 & 50 & 60 & 88 \\
3 & 73.3 & 76 & 80 & 76 & 88 & 88 & 88 \\
4 & 76.0 & 76 & 76 & 76 & 56 & 88 & 90 \\
5 & 75.7 & 76 & 76 & 56 & 68 & 44 & 82 \\
6 & 0.0 & 0 & 80 & 20 & 0 & 20 & 70 \\
7 & 55.0 & 75 & 80 & 76 & 60 & 76 & 100 \\
8 & 66.7 & 64 & 80 & 48 & 90 & 68 & 88 \\
9 & 75.7 & 76 & 76 & 64 & 76 & 88 & 68 \\
10 & 60.0 & 76 & 80 & 64 & 64 & 64 & 80 \\
\midrule
\textbf{Average} & 62.23 & 65.3 & 79.6 & 60 & 61.6 & 68.4 & 83.8 \\
\bottomrule
\end{tabular}
\caption{Performance scores of different LLMs as examiners on Physical Exams. The total score is calculated as follows: Score1(Exam Coverage) * 0.6 + Score2(Reason Relevance and Accuracy) * 0.4.}
\label{tab:performance scores of different LLMs as examiners on Physical Exams}
\end{table*}

\begin{table*}
\centering
\caption{Performance scores of different LLMs as medical student on Physical Exams. Score1 is the score of the Exam Coverage, Score2 represents Reason Relevance and Accuracy, and Score3 is for Extra Exam Penalty. Each of Score1 and Score2 has been adjusted to a 100-point scale, while Score3 is calculated by deducting 5 points for each extra exam. The total score is calculated as follows: Score1 * 0.6 + Score2 * 0.4 + Score3 (All Score3 are negative or zero)}
\label{tab:performance scores of different LLMs as medical student on Physical Exams}
\begin{tabular}{p{3cm}p{2cm}p{2cm}p{2cm}p{2cm}}
\hline
\textbf{Model} & \textbf{Score1} & \textbf{Score2} & \textbf{Score3} & \textbf{Overall Score} \\ 
\hline
GPT3.5 & 16.55 & 87.5 & -3.18 & 43.34 \\ 
GPT4 & 37.05 & 79.55 & -10.91 & 48.59 \\ 
GPT4o & 51.02 & 80.68 & -20 & 52.89 \\ 
Claude Opus & 38.07 & 86.36 & -14.09 & 50.34 \\ 
Claude Haiku & 51.82 & 75 & -20.45 & 50.86 \\ 
Claude Sonnet & 46.36 & 80.68 & -14.55 & 52.82 \\ 
\hline
\end{tabular}
\end{table*}

\begin{table*}
\centering
\caption{LLM-as-medical-student for Physical Exams: GPT3.5 (left) and GPT4 (right). Score1 is the score of the Exam Coverage, Score2 represents Reason Relevance and Accuracy, and Score3 is for Extra Exam Penalty. Each of Score1 and Score2 has been adjusted to a 100-point scale, while Score3 is calculated by deducting 10 points for each extra exam and then applying a 0.5 coefficient, effectively deducting 5 points per extra exam. The total score is calculated as follows: Score1 * 0.6 + Score2 * 0.4 + Score3 (All Score3 are negative or zero).}
\begin{tabular}{|p{0.5cm}|p{1cm}|p{1cm}|p{1cm}|p{1cm}||p{0.5cm}|p{1cm}|p{1cm}|p{1cm}|p{1cm}|}
\hline
\textbf{case} & \textbf{score1} & \textbf{score2} & \textbf{score3} & \textbf{overall} & \textbf{case} & \textbf{score1} & \textbf{score2} & \textbf{score3} & \textbf{overall} \\ 
\hline
1 & 0 & 100 & -10 & 35 & 1 & 30 & 100 & -10 & 53 \\ 
2 & 20 & 100 & 0 & 52 & 2 & 60 & 100 & 0 & 76 \\ 
3 & 33.33 & 100 & 0 & 60 & 3 & 60 & 100 & -10 & 71 \\ 
4 & 20 & 100 & 0 & 52 & 4 & 60 & 100 & -10 & 71 \\ 
5 & 20 & 50 & -10 & 27 & 5 & 60 & 100 & -10 & 71 \\ 
6 & 0 & 0 & -20 & -10 & 6 & 0 & 0 & 0 & 0 \\ 
7 & 20 & 100 & 0 & 52 & 7 & 33.33 & 100 & -10 & 55 \\ 
8 & 20 & 100 & 0 & 52 & 8 & 40 & 50 & -10 & 39 \\ 
9 & 20 & 100 & 0 & 52 & 9 & 60 & 100 & -10 & 71 \\ 
10 & 20 & 100 & 0 & 52 & 10 & 60 & 100 & 0 & 76 \\ 
11 & 40 & 100 & 0 & 64 & 11 & 60 & 100 & 0 & 76 \\ 
12 & 33.33 & 100 & 0 & 60 & 12 & 50 & 100 & 0 & 70 \\ 
13 & 20 & 100 & 0 & 52 & 13 & 50 & 100 & 0 & 70 \\ 
14 & 20 & 100 & 0 & 52 & 14 & 40 & 100 & 0 & 64 \\ 
15 & 0 & 100 & -20 & 30 & 15 & 20 & 100 & -20 & 42 \\ 
16 & 0 & 100 & -10 & 35 & 16 & 33.33 & 100 & -10 & 55 \\ 
17 & 20 & 100 & 0 & 52 & 17 & 20 & 100 & 0 & 52 \\ 
18 & 0 & 100 & -10 & 35 & 18 & 0 & 0 & -30 & -15 \\ 
19 & 0 & 0 & -10 & -5 & 19 & 0 & 0 & -40 & -20 \\ 
20 & 20 & 100 & 0 & 52 & 20 & 50 & 100 & -10 & 65 \\ 
21 & 16.67 & 100 & 0 & 50 & 21 & 33.33 & 100 & 0 & 60 \\ 
22 & 10 & 100 & 0 & 46 & 22 & 33.33 & 100 & 0 & 60 \\ 
23 & 16.67 & 100 & 0 & 50 & 23 & 20 & 50 & -20 & 22 \\ 
24 & 10 & 100 & 0 & 46 & 24 & 20 & 100 & -10 & 47 \\ 
25 & 25 & 100 & 0 & 55 & 25 & 50 & 100 & 0 & 70 \\ 
26 & 20 & 100 & 0 & 52 & 26 & 40 & 100 & -10 & 59 \\ 
27 & 16.67 & 100 & 0 & 50 & 27 & 20 & 100 & -10 & 47 \\ 
28 & 16.67 & 100 & 0 & 50 & 28 & 66.67 & 100 & -10 & 75 \\ 
29 & 16.67 & 100 & 0 & 50 & 29 & 66.67 & 100 & -10 & 75 \\ 
30 & 0 & 50 & -10 & 15 & 30 & 0 & 0 & -40 & -20 \\ 
31 & 20 & 100 & 0 & 52 & 31 & 33.33 & 100 & -10 & 55 \\ 
32 & 20 & 100 & 0 & 52 & 32 & 20 & 100 & -30 & 37 \\ 
33 & 40 & 50 & -10 & 39 & 33 & 60 & 100 & -10 & 71 \\ 
34 & 20 & 100 & 0 & 52 & 34 & 66.67 & 100 & 0 & 80 \\ 
35 & 20 & 100 & 0 & 52 & 35 & 40 & 100 & -10 & 59 \\ 
36 & 20 & 100 & 0 & 52 & 36 & 20 & 50 & -20 & 22 \\ 
37 & 33.33 & 100 & 0 & 60 & 37 & 66.67 & 100 & -10 & 75 \\ 
38 & 20 & 100 & 0 & 52 & 38 & 40 & 50 & -10 & 39 \\ 
39 & 20 & 100 & 0 & 52 & 39 & 40 & 100 & 0 & 64 \\ 
40 & 0 & 0 & -10 & -5 & 40 & 0 & 0 & -40 & -20 \\ 
41 & 0 & 50 & -10 & 15 & 41 & 0 & 0 & -20 & -10 \\ 
42 & 0 & 50 & -10 & 15 & 42 & 0 & 0 & -30 & -15 \\ 
43 & 20 & 100 & 0 & 52 & 43 & 40 & 100 & 0 & 64 \\ 
44 & 20 & 100 & 0 & 52 & 44 & 66.67 & 100 & 0 & 80 \\  
\hline
Ave & 16.55 & 87.5 & -3.18 & 43.34 & Ave & 37.05 & 79.55 & -10.91 & 48.59 \\ 		
\hline
\end{tabular}
\label{tab:gpt4_3.5_physical_exam}
\end{table*}

\begin{table*}
\centering
\caption{LLM-as-medical-student for Physical Exams: GPT4o (left) and Claude Opus (right). Score1 is the score of the Exam Coverage, Score2 represents Reason Relevance and Accuracy, and Score3 is for Extra Exam Penalty. Each of Score1 and Score2 has been adjusted to a 100-point scale, while Score3 is calculated by deducting 10 points for each extra exam and then applying a 0.5 coefficient, effectively deducting 5 points per extra exam. The total score is calculated as follows: Score1 * 0.6 + Score2 * 0.4 + Score3 (All Score3 are negative or zero).}
\begin{tabular}{|p{0.5cm}|p{1cm}|p{1cm}|p{1cm}|p{1cm}||p{0.5cm}|p{1cm}|p{1cm}|p{1cm}|p{1cm}|}
\hline
\textbf{case} & \textbf{score1} & \textbf{score2} & \textbf{score3} & \textbf{overall} & \textbf{case} & \textbf{score1} & \textbf{score2} & \textbf{score3} & \textbf{overall} \\ 
\hline
1 & 66.67 & 100 & -20 & 70 & 1 & 40 & 100 & -10 & 59 \\ 
2 & 100 & 100 & -10 & 95 & 2 & 75 & 100 & 0 & 85 \\ 
3 & 75 & 100 & -10 & 80 & 3 & 75 & 100 & 0 & 85 \\ 
4 & 60 & 100 & -10 & 71 & 4 & 60 & 100 & -10 & 71 \\ 
5 & 80 & 100 & -20 & 78 & 5 & 20 & 50 & -20 & 22 \\ 
6 & 0 & 0 & -50 & -25 & 6 & 0 & 100 & -40 & 20 \\ 
7 & 20 & 50 & -30 & 17 & 7 & 33.33 & 100 & -20 & 50 \\ 
8 & 66.67 & 100 & -10 & 75 & 8 & 50 & 100 & 0 & 70 \\ 
9 & 80 & 100 & -30 & 73 & 9 & 40 & 100 & -20 & 54 \\ 
10 & 60 & 100 & -10 & 71 & 10 & 50 & 100 & -10 & 65 \\ 
11 & 80 & 100 & -10 & 83 & 11 & 60 & 100 & 0 & 76 \\ 
12 & 50 & 100 & -10 & 65 & 12 & 66.67 & 100 & -10 & 75 \\ 
13 & 33.33 & 100 & -10 & 55 & 13 & 33.33 & 100 & -10 & 55 \\ 
14 & 60 & 100 & -10 & 71 & 14 & 66.67 & 100 & -10 & 75 \\ 
15 & 20 & 100 & -30 & 37 & 15 & 0 & 100 & -40 & 20 \\ 
16 & 33.33 & 50 & -20 & 30 & 16 & 33.33 & 100 & -10 & 55 \\ 
17 & 50 & 100 & 0 & 70 & 17 & 40 & 100 & 0 & 64 \\ 
18 & 0 & 0 & -60 & -30 & 18 & 0 & 0 & -20 & -10 \\ 
19 & 0 & 0 & -70 & -35 & 19 & 0 & 0 & -30 & -15 \\ 
20 & 80 & 100 & -10 & 83 & 20 & 33.33 & 100 & -10 & 55 \\ 
21 & 83.33 & 100 & -10 & 85 & 21 & 50 & 100 & 0 & 70 \\ 
22 & 100 & 100 & -20 & 90 & 22 & 75 & 100 & 0 & 85 \\ 
23 & 50 & 100 & -10 & 65 & 23 & 50 & 100 & 0 & 70 \\ 
24 & 40 & 50 & -10 & 39 & 24 & 20 & 100 & -10 & 47 \\ 
25 & 80 & 100 & -10 & 83 & 25 & 0 & 100 & -40 & 20 \\ 
26 & 83.33 & 100 & -10 & 85 & 26 & 50 & 100 & -10 & 65 \\ 
27 & 83.33 & 100 & -10 & 85 & 27 & 20 & 100 & -20 & 42 \\ 
28 & 66.67 & 100 & -20 & 70 & 28 & 40 & 100 & -10 & 59 \\ 
29 & 50 & 100 & -10 & 65 & 29 & 50 & 100 & 0 & 70 \\ 
30 & 0 & 0 & -40 & -20 & 30 & 0 & 0 & -30 & -15 \\ 
31 & 66.67 & 100 & -20 & 70 & 31 & 50 & 100 & -20 & 60 \\ 
32 & 60 & 100 & -20 & 66 & 32 & 40 & 100 & -10 & 59 \\ 
33 & 66.67 & 100 & -10 & 75 & 33 & 40 & 50 & -20 & 34 \\ 
34 & 66.67 & 100 & -10 & 75 & 34 & 50 & 100 & -10 & 65 \\ 
35 & 40 & 100 & -10 & 59 & 35 & 60 & 100 & -10 & 71 \\ 
36 & 40 & 100 & -10 & 59 & 36 & 40 & 100 & -10 & 59 \\ 
37 & 33.33 & 100 & -20 & 50 & 37 & 33.33 & 100 & -20 & 50 \\ 
38 & 60 & 100 & -20 & 66 & 38 & 40 & 100 & 0 & 64 \\ 
39 & 20 & 100 & -30 & 37 & 39 & 50 & 100 & -10 & 65 \\ 
40 & 0 & 0 & -40 & -20 & 40 & 0 & 100 & -40 & 20 \\ 
41 & 0 & 0 & -30 & -15 & 41 & 0 & 0 & -40 & -20 \\ 
42 & 0 & 0 & -60 & -30 & 42 & 0 & 0 & -40 & -20 \\ 
43 & 60 & 100 & -10 & 71 & 43 & 60 & 100 & 0 & 76 \\ 
44 & 80 & 100 & -10 & 83 & 44 & 80 & 100 & 0 & 88 \\ 
\hline
Ave & 51.02 & 80.68 & -20 & 52.89 & Ave & 38.07 & 86.36 & -14.09 & 50.34 \\ 	
\hline
\end{tabular}
\label{tab:gpt4o_opus_physical_exam}
\end{table*}

\begin{table*}
\centering
\caption{LLM-as-medical-student for Physical Exams: Claude Haiku (left) and Claude Sonnet (right). Score1 is the score of the Exam Coverage, Score2 represents Reason Relevance and Accuracy, and Score3 is for Extra Exam Penalty. Each of Score1 and Score2 has been adjusted to a 100-point scale, while Score3 is calculated by deducting 10 points for each extra exam and then applying a 0.5 coefficient, effectively deducting 5 points per extra exam. The total score is calculated as follows: Score1 * 0.6 + Score2 * 0.4 + Score3 (All Score3 are negative or zero).}
\begin{tabular}{|p{0.5cm}|p{1cm}|p{1cm}|p{1cm}|p{1cm}||p{0.5cm}|p{1cm}|p{1cm}|p{1cm}|p{1cm}|}
\hline
\textbf{case} & \textbf{score1} & \textbf{score2} & \textbf{score3} & \textbf{overall} & \textbf{case} & \textbf{score1} & \textbf{score2} & \textbf{score3} & \textbf{overall} \\ 
\hline
1 & 66.67 & 100 & -10 & 75 & 1 & 60 & 100 & -10 & 71 \\ 
2 & 80 & 100 & -40 & 68 & 2 & 83.33 & 100 & -10 & 85 \\ 
3 & 75 & 100 & -10 & 80 & 3 & 66.67 & 100 & -10 & 75 \\ 
4 & 40 & 100 & -20 & 54 & 4 & 60 & 100 & -10 & 71 \\ 
5 & 80 & 100 & -10 & 83 & 5 & 60 & 100 & -10 & 71 \\ 
6 & 0 & 0 & -90 & -45 & 6 & 0 & 0 & -40 & -20 \\ 
7 & 50 & 100 & -10 & 65 & 7 & 50 & 100 & 0 & 70 \\ 
8 & 50 & 100 & -10 & 65 & 8 & 40 & 100 & -10 & 59 \\ 
9 & 100 & 100 & 0 & 100 & 9 & 80 & 100 & -10 & 83 \\ 
10 & 50 & 100 & -10 & 65 & 10 & 60 & 100 & -10 & 71 \\ 
11 & 60 & 100 & -10 & 71 & 11 & 60 & 100 & 0 & 76 \\ 
12 & 75 & 100 & 0 & 85 & 12 & 50 & 100 & -10 & 65 \\ 
13 & 75 & 100 & -10 & 80 & 13 & 75 & 100 & -10 & 80 \\ 
14 & 100 & 100 & -20 & 90 & 14 & 60 & 100 & 0 & 76 \\ 
15 & 0 & 0 & -30 & -15 & 15 & 40 & 50 & -20 & 34 \\ 
16 & 33.33 & 100 & -20 & 50 & 16 & 33.33 & 50 & -30 & 25 \\ 
17 & 50 & 100 & -10 & 65 & 17 & 50 & 100 & 0 & 70 \\ 
18 & 0 & 0 & -50 & -25 & 18 & 0 & 0 & -30 & -15 \\ 
19 & 0 & 0 & -80 & -40 & 19 & 0 & 0 & -50 & -25 \\ 
20 & 60 & 100 & -10 & 71 & 20 & 66.67 & 100 & 0 & 80 \\ 
21 & 83.33 & 100 & -10 & 85 & 21 & 50 & 100 & 0 & 70 \\ 
22 & 100 & 100 & -10 & 95 & 22 & 75 & 100 & 0 & 85 \\ 
23 & 50 & 100 & -20 & 60 & 23 & 50 & 100 & -10 & 65 \\ 
24 & 40 & 100 & -10 & 59 & 24 & 33.33 & 100 & -10 & 55 \\ 
25 & 100 & 100 & 0 & 100 & 25 & 66.67 & 100 & -10 & 75 \\ 
26 & 83.33 & 100 & -10 & 85 & 26 & 80 & 100 & 0 & 88 \\ 
27 & 40 & 50 & -20 & 34 & 27 & 40 & 100 & -10 & 59 \\ 
28 & 33.33 & 50 & -10 & 35 & 28 & 50 & 100 & -10 & 65 \\ 
29 & 60 & 50 & -20 & 46 & 29 & 60 & 100 & -10 & 71 \\ 
30 & 0 & 0 & -50 & -25 & 30 & 0 & 0 & -30 & -15 \\ 
31 & 25 & 50 & -20 & 25 & 31 & 66.67 & 100 & -20 & 70 \\ 
32 & 50 & 100 & -10 & 65 & 32 & 40 & 100 & -30 & 49 \\ 
33 & 66.67 & 100 & -20 & 70 & 33 & 40 & 50 & -20 & 34 \\ 
34 & 83.33 & 100 & -10 & 85 & 34 & 50 & 100 & -10 & 65 \\ 
35 & 60 & 100 & -10 & 71 & 35 & 40 & 100 & -10 & 59 \\ 
36 & 40 & 50 & -20 & 34 & 36 & 40 & 100 & -10 & 59 \\ 
37 & 50 & 100 & -20 & 60 & 37 & 66.67 & 100 & -30 & 65 \\ 
38 & 60 & 50 & -10 & 51 & 38 & 40 & 100 & -10 & 59 \\ 
39 & 83.33 & 100 & -10 & 85 & 39 & 50 & 100 & -10 & 65 \\ 
40 & 0 & 0 & -40 & -20 & 40 & 0 & 0 & -40 & -20 \\ 
41 & 0 & 0 & -40 & -20 & 41 & 0 & 0 & -40 & -20 \\ 
42 & 0 & 0 & -70 & -35 & 42 & 0 & 0 & -40 & -20 \\ 
43 & 60 & 100 & -10 & 71 & 43 & 40 & 100 & 0 & 64 \\ 
44 & 66.67 & 100 & 0 & 80 & 44 & 66.67 & 100 & -10 & 75 \\ 
\hline
Ave & 51.82 & 75 & -20.45 & 50.86 & Ave & 46.36 & 80.68 & -14.55 & 52.82 \\ 
\hline
\end{tabular}
\label{tab:haiku_sonnet_physical_exam}
\end{table*}

\begin{table*}
\caption{LLM-as-Medical-Student Output Examples for Physical Exams (Part 1)}
\label{tab:LLM-as-Medical-Student Output Examples for Physical Exams}
    \begin{scriptsize}
    \setlength{\tabcolsep}{5pt}
    \renewcommand{\arraystretch}{1.05}
    \centering
    \begin{tabularx}{\textwidth}{lp{14.1cm}}
    \toprule
    \multirow{2}{*}{Ground Truth} & 
    \textbf{Neck examination: JVD, carotid auscultation}\\
    & \textbf{Ocular examination: auscultation, palpation}\\
    & \textbf{Pulmonary examination: auscultation, palpation, percussion}\\
    & \textbf{Abdominal exam: auscultation, palpation, percussion}\\
    & \textbf{Extremities: Checked peripheral pulses, checked blood pressure in both arms, looked for edema and cyanosis}\\
    \midrule
    \multirow{2}{*}{GPT3.5} & 
    \textbf{Heart: Cardiac auscultation}
    \newline
    ``\emph{reason: The patient's chief complaint of chest pain, along with associated symptoms such as shortness of breath, nausea, sweating, and radiation of pain to the neck and left arm, warrant a focused examination of the heart to assess for any possible cardiac issues.}'' \\
    \midrule
    \multirow{2}{*}{GPT4} & 
    \textbf{Heart: Auscultation for possible abnormal heart sounds and rhythm such as murmurs, gallop sound, or arrhythmias}
    \newline
    ``\emph{reason: The patient is presenting with chest pain, shortness of breath, and sweating which can indicate a cardiac issue like angina or a heart attack. This is further supported by his hypertension, high cholesterol, history of GERD, and cocaine use.}'' \\
    & \textbf{Chest: Inspection and palpation of the chest to check for any visible or palpable abnormalities}
    \newline
    ``\emph{reason: The patient's chief complaint is chest pain.}'' \\
    & \textbf{Pulmonary: Auscultation of the lungs for any abnormal breath sounds such as wheezing, crackles or decreased breath sounds}
    \newline
    ``\emph{reason: The patient is experiencing difficulty breathing and chest pain which could indicate a pulmonary issue.}'' \\
    \midrule
    \multirow{2}{*}{GPT4o} & 
    \textbf{Chest}
    \newline
    ``\emph{reason: The patient presents with chest pain, thus examining the chest is essential for assessing potential cardiac or respiratory issues}'' \\
    & \textbf{Heart}
    \newline
    ``\emph{reason: Given the chest pain with radiation to the neck, upper back, and left arm accompanied by nausea and diaphoresis, it's important to listen for murmurs, rubs, or gallops that might suggest cardiac pathology.}'' \\
    & \textbf{Neck: Inspection and palpation of the carotid arteries.}
    \newline
    ``\emph{reason: Since the pain radiates to the neck, examining the neck including carotid arteries for bruits or other abnormalities can provide information about vascular causes of chest pain.}'' \\
    & \textbf{Abdomen}
    \newline
    ``\emph{reason: To rule out gastrointestinal causes of chest pain such as GERD, especially given the patient's history of GERD and the past relief of pain with antacids.}'' \\
    & \textbf{Pulmonary}
    \newline
    ``\emph{reason: To assess for any respiratory component to the chest pain such as pleurisy or pneumothorax, especially since the patient reports associated shortness of breath.}'' \\
    & \textbf{Vital Signs (VS)}
    \newline
    ``\emph{reason: To obtain current vital signs to assess for fever, tachycardia, tachypnea, or hypertension that may be related to the patient's chest pain and overall condition.}'' \\
    & \textbf{Extremities: Inspection for edema, palpation for temperature, and assessment of pulses}
    \newline
    ``\emph{reason: To check for signs of peripheral vascular disease or deep vein thrombosis, as these may have cardiorespiratory implications and relate to the potential for pulmonary embolism given the complaint of chest pain and shortness of breath.}'' \\
    & \textbf{Neuro: Assessment of cranial nerves, motor and sensory examination, coordination, and reflexes.}
    \newline
    ``\emph{reason: To rule out neurological deficits that could be associated with stroke or other central nervous system pathologies, especially considering the patient’s use of cocaine which is associated with an increased risk of stroke.}'' \\
    & \textbf{HEENT (Head, Eyes, Ears, Nose, Throat)}
    \newline
    ``\emph{reason: To assess for any abnormalities or signs that could contribute to the patient's chief complaint, such as jugular vein distention that may indicate heart failure.}'' \\
    & \textbf{Skin: Inspection for pallor, cyanosis, or sweating.}
    \newline
    ``\emph{reason: The patient complains of increased sweating, and skin examination can provide clues to systemic conditions, such as cyanosis in hypoxemia or pallor in anemia.}'' \\
    \midrule
    \multirow{2}{*}{Claude Opus} & 
    \textbf{Chest: Palpation and auscultation.}
    \newline
    ``\emph{reason: The patient's chief complaint is chest pain, which has been present for 40 minutes, is constant, and rates 7/10 in severity. The pain is described as pressure-like and is located in the middle of the chest.}'' \\
    & \textbf{Heart: Auscultation.}
    \newline
    ``\emph{reason: Given the patient's risk factors (hypertension, high cholesterol, cocaine use, and smoking history) and the nature of the chest pain (pressure-like, radiating to neck, upper back, and left arm), a cardiac examination is warranted to assess for potential cardiac causes of the pain, such as acute coronary syndrome.}'' \\
    & \textbf{Pulmonary: Auscultation and percussion.}
    \newline
    ``\emph{reason: The patient reports shortness of breath, which could be related to a pulmonary cause or secondary to cardiac issues. Auscultation and percussion of the lungs can help assess for abnormalities such as wheezing, crackles, or reduced breath sounds.}'' \\
    & \textbf{Neck: Inspection and palpation.}
    \newline
    ``\emph{reason: The patient's chest pain radiates to the neck, warranting an examination of the neck for any visible swelling, tenderness, or lymphadenopathy that could provide clues to the etiology of the pain.}'' \\
    & \textbf{Back: Palpation and percussion.}
    \newline
    ``\emph{reason: As the chest pain also radiates to the upper back, an examination of the back, including palpation and percussion, can help rule out potential musculoskeletal or spinal causes of the pain.}'' \\
    \bottomrule
    \end{tabularx}
    \end{scriptsize}
\end{table*}

\begin{table*}
\caption{LLM-as-Medical-Student Output Examples for Physical Exams (Part 2)}
\label{tab:LLM-as-Medical-Student Output Examples for Physical Exams 2}
    \begin{scriptsize}
    \setlength{\tabcolsep}{5pt}
    \renewcommand{\arraystretch}{1.05}
    \centering
    \begin{tabularx}{\textwidth}{lp{14.1cm}}
    \toprule
    \multirow{2}{*}{Sonnet} & 
    \textbf{Chest: Inspection, palpation, auscultation}
    \newline
    ``\emph{reason: The patient is presenting with chest pain, which could be indicative of a cardiac or pulmonary condition, necessitating a thorough examination of the chest.}'' \\
    & \textbf{Heart: Auscultation, palpation}
    \newline
    ``\emph{reason: The patient's chest pain radiates to the neck, upper back, and left arm, which could be a sign of cardiac involvement. Examining the heart is crucial to assess for any abnormalities.}'' \\
    & \textbf{Extremities: Inspection, palpation, peripheral pulses}
    \newline
    ``\emph{reason: The patient's chest pain radiates to the left arm, which could be a sign of vascular involvement or peripheral vascular disease. Examining the extremities can provide valuable information.}'' \\
    & \textbf{VS: Measure blood pressure, heart rate, respiratory rate, temperature}
    \newline
    ``\emph{reason: The patient's vital signs, including elevated blood pressure, are provided, and monitoring them can help assess the patient's overall condition and identify any potential abnormalities.}'' \\
    \midrule
    \multirow{2}{*}{Claude Haiku} & 
    \textbf{Abdomen: Palpation and percussion}
    \newline
    ``\emph{reason: The patient has reported bloody urine, which could be a symptom of an underlying abdominal or urological condition. Palpating and percussing the abdomen can help assess for any masses, tenderness, or other abnormalities that may be contributing to the patient's symptoms.}'' \\
    & \textbf{Genitourinary: Visual inspection and palpation of the external genitalia}
    \newline
    ``\emph{reason: The patient's chief complaint is bloody urine, which is a genitourinary symptom. Examining the external genitalia can help identify any visible abnormalities that may be causing the bleeding.}'' \\
    & \textbf{Back: Palpation and percussion}
    \newline
    ``\emph{reason: The patient reported no abdominal or flank pain, but a thorough examination of the back, including the flanks, is important to rule out any kidney or urinary tract issues that may be causing the hematuria.}'' \\
    & \textbf{Pulmonary: Auscultation of the lungs}
    \newline
    ``\emph{reason: The patient reported no respiratory symptoms, but a pulmonary examination can help rule out any underlying lung conditions that may be contributing to the patient's overall health status.}'' \\
    & \textbf{Cardiovascular: Palpation of the pulse, auscultation of the heart}
    \newline
    ``\emph{reason: The patient's vital signs, including heart rate, were within normal limits, but a focused cardiovascular exam can help identify any underlying cardiac conditions that may be related to the patient's presentation.}'' \\
    & \textbf{Extremities: Inspection and palpation}
    \newline
    ``\emph{reason: The patient's occupation as a painter may increase the risk of certain health conditions, and a thorough examination of the extremities can help identify any signs of peripheral vascular disease or other issues.}'' \\
    & \textbf{Neuro: Cranial nerve examination, assessment of motor and sensory function}
    \newline
    ``\emph{reason: While the patient did not report any neurological symptoms, a focused neurological exam can help rule out any underlying neurological conditions that may be contributing to the patient's overall health status.}'' \\
    & \textbf{HEENT: Examination of the head, eyes, ears, nose, and throat}
    \newline
    ``\emph{reason: The patient's bloody urine could be related to an underlying condition affecting the head, eyes, ears, nose, or throat, and a thorough HEENT exam can help identify any abnormalities in these areas.}'' \\
    & \textbf{Skin}
    \newline
    ``\emph{reason: The patient's occupation as a painter may increase the risk of certain skin conditions, and a focused skin exam can help identify any signs of dermatological issues that may be contributing to the patient's overall health status.}'' \\
    \bottomrule
    \end{tabularx}
    \end{scriptsize}
\end{table*}

\subsection{Closure}
\subsubsection{Human Annotation Case Study}
There are five key dimensions for Closure Examination: Diagnostic Impressions, Management Plans, Challenging Questions, Language Accessibility, and Compassion. Three experts were engaged to annotate 10 different cases, with detailed results provided in the table \ref{tab:human_anntation_closure} and table \ref{tab:human_anntation_closure_2}.

Unlike other sections, the Closure part cannot be easily quantified into discrete items for direct comparison, resulting in significant variability in subjective judgments among physicians and consequently lower inter-rater agreement compared to other modules.

Analyzing the score details, particularly Score3 (Challenge Question), reveals that the correlation between experts ranges from 0.6 to 0.7. Scores like Score1 (Diagnostic Impressions) and Score2 (Management Plans), which focus more on summarizing information points, show higher correlations between 0.7 and 0.8. However, substantial differences still exist in the degree of deductions within the same case. For instance, in Score2 (Management Plans), although all three experts cited the same reasons for deduction, Expert 2 considered the missing element very critical, leading to a 67\% deduction, whereas Experts 1 and 3 viewed its absence as less significant, resulting in a 50\% deduction.

Similarly, in Score1 (Diagnostic Impressions), Experts 1 and 2 agreed that the MedStuLLM's answer covered all the essential points, while Expert 3 believed that mentioning the possibility of infection was unnecessary and inappropriate, as it could cause patient distress, and thus deducted points.

During the final score calculation, we adjusted the weights to reflect the greater importance of conveying the plan to the patient in the closure part and the more subjective nature of the challenge question. Therefore, the weights were adjusted from the initial 0.2, 0.3, 0.3, 0.1, 0.1 to 0.2, 0.4, 0.2, 0.1, 0.1.

Although Scores4 (Language Accessibility) and Score5 (Compassion) were consistently given full marks by experts across 10 cases, the LLM-as-CS-examiner provides more granular scoring. This granularity helps reflect the model's ability to communicate when scoring the outputs of various  MedStuLLMs.

\begin{enumerate}
    \item Closure Ground Truth:
    \begin{itemize}
        \item Sample Closure: Mr. Matthews, it's a good thing that you decided to come to the emergency department. For future reference, it's important to seek medical treatment after an accident like this. We will need to run some tests and take imaging studies of your chest to determine the cause of your pain and breathing difficulties. We will also provide you with something for your pain and monitor your breathing to ensure sufficient oxygen intake. Do you have any questions for me?
        \item Challenging Question: Do you think I am going to die?
        \item Samlpe Answer: Your condition raises concern and is obviously urgent. We will start by taking some images of your chest. Then, once we have a better idea of what is wrong, we can give you some medication to help you with your pain. If there is air or blood around your lungs, there is a procedure we can perform to release the pressure. We will be monitoring you very closely from this point on, and if you have any significant problems, we will be available to help.
    \end{itemize}

    \item MedStuLLM for Closure:
    \begin{itemize}
        \item Closure: Mr. Matthew, based on your history and physical exam, you had a chest trauma from the car accident, you’re experiencing severe pain in your chest, having difficulty with breathing, and you have a cough producing yellow sputum. These symptoms could suggest a possible lung injury or infection, or possibly a rib fracture from the accident. At this point, we cannot be certain of your diagnosis, so we plan to do further workup. We’ll perform some imaging studies like X-Rays or a CT scan to check for any fractures or lung injuries, and we might carry out some blood tests to check for infection or inflammation. Please do not worry, our team is here to help you.
        \item Challenging Question: Do you think I am going to die?
        \item Answer: Mr. Matthew, I understand your concern and it’s completely normal to be worried in such situations. However, based on your symptoms and the information you’ve provided, I would like to assure you that we are going to do everything in our capacity to help you. Our team will continue testing to confirm the cause of your pain and will provide appropriate treatment. Therefore, although you are in pain, I don’t think you are in any immediate life-threatening situation.
    \end{itemize}

    \item Human Annotation:
    \begin{itemize}
        \item \textbf{Expert 1: 85}
        
        score1(Diagnostic Impressions): 20/20

        score2(Management Plans): 15/30 (reason: Missed "provide you with something for your pain and monitor your breathing to ensure sufficient oxygen intake")

        score3(Challenging Questions): 30/30

        score4(Language Accessibility): 10/10

        score5(Compassion): 10/10
        \item \textbf{Expert 1: 75}
        
        score1(Diagnostic Impressions): 20/20

        score2(Management Plans): 10/30 (reason: Missed "provide you with something for your pain and monitor your breathing to ensure sufficient oxygen intake")

        score3(Challenging Questions): 25/30

        score4(Language Accessibility): 10/10

        score5(Compassion): 10/10
        \item \textbf{Expert 1: 75}
        
        score1(Diagnostic Impressions): 10/20 (reason: infection should not be mentioned)

        score2(Management Plans): 15/30 (reason: Missed "provide you with something for your pain and monitor your breathing to ensure sufficient oxygen intake")

        score3(Challenging Questions): 30/30

        score4(Language Accessibility): 10/10

        score5(Compassion): 10/10
    \end{itemize}
 \end{enumerate}

\subsubsection{LLM-as-cs-examiner error analysis}
The correlation of GPT4o and Claude Opus with the experts' scores are both higher than all other models, indicating its strong alignment with expert judgment. Detailed scores are provided in the Table \ref{tab:performance scores of different LLMs as examiners on Closure}.

\begin{table*}
\centering
\caption{Performance scores of different LLMs as examiners on Closure. The total score is calculated by weights 0.2, 0.4, 0.2, 0.1, and 0.1 for Score1 to Score5 respectively. Score1 is the score of the Diagnostic Impressions, Score2 represents Management Plans, Score3 represents Challenging Questions, Score4 represents Language Accessibility, and Score5 is for Compassion. Each score has been adjusted to a 100-point scale.}
\label{tab:performance scores of different LLMs as examiners on Closure}
\begin{tabular}{lcccccccc}
\toprule
\textbf{Case} & \textbf{Experts} & \textbf{GPT 4} & \textbf{GPT 3.5} & \textbf{GPT 4-o} & \textbf{Claude3 Opus} & \textbf{Sonnet} & \textbf{Haiku} \\
\midrule
1 & 70 & 70 & 82 & 73.3 & 78.3 & 70 & 64.67 \\ 
2 & 99 & 82 & 90 & 88.3 & 90.3 & 76.3 & 64.67 \\ 
3 & 80 & 85 & 68 & 78.3 & 71 & 69.67 & 70 \\ 
4 & 73 & 72 & 82 & 78.3 & 71 & 70 & 62.67 \\ 
5 & 69 & 82 & 82 & 81.7 & 68.3 & 88.3 & 70 \\ 
6 & 73 & 76 & 72 & 76.3 & 78.3 & 78.3 & 70 \\ 
7 & 75 & 70 & 89 & 78.3 & 76 & 88.3 & 70 \\ 
8 & 81 & 88 & 82 & 78.3 & 75 & 73.3 & 78.3 \\ 
9 & 80 & 82 & 83 & 78.3 & 78.3 & 88 & 70 \\ 
10 & 80 & 85 & 70 & 78.3 & 75 & 85 & 70 \\ 
\midrule
\textbf{Average} & 78.0 & 79.1 & 79.9 & 78.9 & 76.2 & 78.7 & 69.0 \\		
\bottomrule
\end{tabular}
\end{table*}

\subsubsection{LLM-as-medical-student Case Study}
Analyzing the average scores from 44 cases as presented in Table \ref{tab:performance scores of different LLMs as medical student on Closure}, with detailed scores available in Tables \ref{tab:gpt3.5_gpt4_closure}, \ref{tab:gpt4o_opus_closure}, and \ref{tab:haiku_sonnet_closure}. Each score, ranging from Score1 to Score5, is on a 100-point scale, and the overall score is calculated using the weights of 0.2, 0.4, 0.2, 0.1, and 0.1 respectively. Overall performance ranks as follows: Claude Opus > GPT-4o > Sonnet > Haiku > GPT-4 > GPT-3.5. 

Examining the detailed performance of each score, we can see the content generated by each model in Tables \ref{tab:LLM-as-Medical-Student Output Examples for Closure} and \ref{tab:LLM-as-Medical-Student Output Examples for Closure 2}. Haiku and Sonnet tend to generate longer responses to comprehensively cover the correct answers. GPT-4o and Opus maintain a normal length, providing precise and comfortable language. GPT-4's responses are also of normal length but demonstrate weaker communication abilities compared to GPT-4o and Opus.

\begin{enumerate}
    \item Score1 (Diagnosis Impressions) and Score2 (Management Plans): These scores reflect the summarization capabilities of the models and show some correlation in their performance. Opus scored the highest, both exceeding 70 points. GPT-4, GPT-4o, Sonnet, and Haiku scored between 65 and 69 points, while GPT-3.5 scored the lowest at 51 points.
    \item Score3 (Challenge Question Response): This score highlights the communication abilities of the models. GPT-4o and Opus scored the highest, with 96 and 97 points, respectively. GPT-4, Sonnet, and Haiku scored around 90 points, while GPT-3.5 scored the lowest at 80 points. Table \ref{tab:LLM-as-Medical-Student Output Examples for Closure} shows that GPT-3.5's responses were often irrelevant, whereas the other models provided more appropriate communication.
    \item Score4 (Language Accessibility) and Score5 (Compassion): These scores also reflect the communication abilities of the models. Apart from GPT-3.5's Score5 of 93, all other models scored close to 100, showcasing strong language expression skills.
 \end{enumerate}

\begin{table*}
\centering
\caption{Performance scores of different LLMs as medical student on Closure. Score1 is the score of the Diagnostic Impressions, Score2 represents Management Plans, Score3 represents Challenging Questions, Score4 represents Language Accessibility (omitted here as all cases are full scores), and Score5 is for Compassion. Each score has been adjusted to a 100-point scale. The weights for Score1 to Score5 are 0.2, 0.4, 0.2, 0.1, and 0.1 respectively, and the total score is calculated accordingly.}
\label{tab:performance scores of different LLMs as medical student on Closure}
\begin{tabular}{p{3cm}p{1.2cm}p{1.2cm}p{1.2cm}p{1.2cm}p{1.2cm}p{2cm}}
\hline
\textbf{Model} & \textbf{Score1} & \textbf{Score2} & \textbf{Score3} & \textbf{Score4} & \textbf{Score5} & \textbf{Overall Score} \\ 
\hline
GPT3.5 & 51.82 & 51.89 & 80.3 & 100 & 93.41 & 66.52 \\
GPT4 & 69.32 & 64.58 & 88.26 & 100 & 98.64 & 77.21 \\
GPT4o & 67.5 & 64.77 & 96.21 & 100 & 97.95 & 78.45 \\
Claude Opus & 75.68 & 71.74 & 97.35 & 100 & 99.55 & 83.26 \\
Claude Haiku & 68.3 & 64.39 & 90.15 & 100 & 99.55 & 77.4 \\
Claude Sonnet & 69.77 & 64.02 & 92.27 & 100 & 98.64 & 77.88 \\
\hline
\end{tabular}
\end{table*}

\begin{table*}
\centering
\caption{LLM-as-medical-student for Closure: GPT-3.5 (left) and GPT-4 (right). Score1 is the score of the Diagnostic Impressions, Score2 represents Management Plans, Score3 represents Challenging Questions, Score4 represents Language Accessibility (omitted here as all cases are full scores), and Score5 is for Compassion. Each score has been adjusted to a 100-point scale. The weights for Score1 to Score5 are 0.2, 0.4, 0.2, 0.1, and 0.1 respectively, and the total score is calculated accordingly.}
\begin{tabular}{|p{0.5cm}|p{0.7cm}|p{0.9cm}|p{0.9cm}|p{0.9cm}|p{0.9cm}||p{0.7cm}|p{0.8cm}|p{0.9cm}|p{0.9cm}|p{1cm}|}
\hline
\textbf{case} & \textbf{score1} & \textbf{score2} & \textbf{score3} &\textbf{score5} & \textbf{overall} & \textbf{score1} & \textbf{score2} & \textbf{score3} & \textbf{score5} & \textbf{overall} \\ 
\hline
1 & 50 & 50 & 100 & 100 & 70 & 75 & 66.67 & 100 & 100 & 81.67 \\ 
2 & 75 & 50 & 83.33 & 100 & 71.67 & 75 & 66.67 & 100 & 100 & 81.67 \\ 
3 & 0 & 33.33 & 100 & 100 & 53.33 & 75 & 75 & 100 & 100 & 85 \\ 
4 & 75 & 66.67 & 100 & 100 & 81.67 & 75 & 66.67 & 83.33 & 100 & 78.33 \\ 
5 & 50 & 50 & 100 & 100 & 70 & 75 & 66.67 & 100 & 100 & 81.67 \\ 
6 & 50 & 50 & 83.33 & 70 & 63.67 & 75 & 66.67 & 83.33 & 80 & 76.33 \\ 
7 & 50 & 50 & 83.33 & 100 & 66.67 & 75 & 83.33 & 83.33 & 80 & 83 \\ 
8 & 0 & 0 & 66.67 & 100 & 33.33 & 75 & 83.33 & 100 & 100 & 88.33 \\ 
9 & 25 & 16.67 & 83.33 & 80 & 46.33 & 75 & 66.67 & 100 & 100 & 81.67 \\ 
10 & 75 & 66.67 & 83.33 & 100 & 78.33 & 75 & 83.33 & 83.33 & 100 & 85 \\ 
11 & 75 & 66.67 & 83.33 & 80 & 76.33 & 50 & 50 & 100 & 100 & 70 \\ 
12 & 50 & 50 & 100 & 100 & 70 & 50 & 50 & 66.67 & 100 & 63.33 \\ 
13 & 0 & 0 & 50 & 100 & 30 & 75 & 66.67 & 100 & 100 & 81.67 \\ 
14 & 50 & 50 & 100 & 100 & 70 & 50 & 50 & 83.33 & 100 & 66.67 \\ 
15 & 50 & 50 & 83.33 & 100 & 66.67 & 50 & 50 & 83.33 & 100 & 66.67 \\ 
16 & 75 & 66.67 & 83.33 & 100 & 78.33 & 75 & 66.67 & 83.33 & 100 & 78.33 \\ 
17 & 50 & 50 & 100 & 100 & 70 & 50 & 50 & 83.33 & 100 & 66.67 \\ 
18 & 0 & 0 & 66.67 & 100 & 33.33 & 100 & 100 & 100 & 100 & 100 \\ 
19 & 50 & 50 & 66.67 & 80 & 61.33 & 75 & 66.67 & 83.33 & 100 & 78.33 \\ 
20 & 50 & 50 & 66.67 & 100 & 63.33 & 50 & 50 & 83.33 & 100 & 66.67 \\ 
21 & 75 & 83.33 & 66.67 & 80 & 79.67 & 75 & 83.33 & 83.33 & 100 & 85 \\ 
22 & 75 & 66.67 & 100 & 100 & 81.67 & 75 & 66.67 & 83.33 & 100 & 78.33 \\ 
23 & 25 & 33.33 & 83.33 & 100 & 55 & 50 & 50 & 83.33 & 100 & 66.67 \\ 
24 & 90 & 83.33 & 66.67 & 80 & 82.67 & 75 & 66.67 & 83.33 & 100 & 78.33 \\ 
25 & 75 & 66.67 & 83.33 & 100 & 78.33 & 75 & 66.67 & 83.33 & 100 & 78.33 \\ 
26 & 75 & 66.67 & 100 & 100 & 81.67 & 75 & 66.67 & 83.33 & 100 & 78.33 \\ 
27 & 50 & 66.67 & 66.67 & 80 & 68 & 50 & 50 & 66.67 & 100 & 63.33 \\ 
28 & 25 & 16.67 & 66.67 & 100 & 45 & 50 & 50 & 83.33 & 100 & 66.67 \\ 
29 & 50 & 66.67 & 83.33 & 100 & 73.33 & 75 & 66.67 & 83.33 & 100 & 78.33 \\ 
30 & 0 & 33.33 & 0 & 50 & 28.33 & 75 & 66.67 & 83.33 & 100 & 78.33 \\ 
31 & 50 & 50 & 66.67 & 100 & 63.33 & 50 & 50 & 83.33 & 100 & 66.67 \\ 
32 & 75 & 66.67 & 100 & 100 & 81.67 & 75 & 33.33 & 100 & 100 & 68.33 \\ 
33 & 75 & 66.67 & 66.67 & 100 & 75 & 75 & 66.67 & 83.33 & 100 & 78.33 \\ 
34 & 0 & 33.33 & 66.67 & 50 & 41.67 & 75 & 66.67 & 100 & 100 & 81.67 \\ 
35 & 50 & 50 & 83.33 & 100 & 66.67 & 75 & 66.67 & 83.33 & 100 & 78.33 \\ 
36 & 50 & 50 & 50 & 80 & 58 & 50 & 50 & 100 & 100 & 70 \\ 
37 & 50 & 66.67 & 100 & 100 & 76.67 & 75 & 66.67 & 100 & 100 & 81.67 \\ 
38 & 75 & 66.67 & 83.33 & 100 & 78.33 & 75 & 66.67 & 83.33 & 100 & 78.33 \\ 
39 & 90 & 83.33 & 100 & 100 & 91.33 & 75 & 66.67 & 83.33 & 100 & 78.33 \\ 
40 & 75 & 66.67 & 100 & 100 & 81.67 & 75 & 66.67 & 83.33 & 80 & 76.33 \\ 
41 & 50 & 50 & 100 & 100 & 70 & 75 & 50 & 83.33 & 100 & 71.67 \\ 
42 & 75 & 66.67 & 66.67 & 80 & 73 & 100 & 100 & 100 & 100 & 100 \\ 
43 & 50 & 50 & 66.67 & 100 & 63.33 & 50 & 66.67 & 83.33 & 100 & 73.33 \\ 
44 & 75 & 66.67 & 83.33 & 100 & 78.33 & 75 & 66.67 & 100 & 100 & 81.67 \\ 
\hline
Ave & 51.82 & 51.89 & 80.3 & 93.41 & 66.52 & 69.32 & 64.58 & 88.26 & 98.64 & 77.21 \\ 
\hline
\end{tabular}
\label{tab:gpt3.5_gpt4_closure}
\end{table*}

\begin{table*}
\centering
\caption{LLM-as-medical-student for Closure: GPT4o (left) and Claude Opus (right). Score1 is the score of the Diagnostic Impressions, Score2 represents Management Plans, Score3 represents Challenging Questions, Score4 represents Language Accessibility (omitted here as all cases are full scores), and Score5 is for Compassion. Each score has been adjusted to a 100-point scale. The weights for Score1 to Score5 are 0.2, 0.4, 0.2, 0.1, and 0.1 respectively, and the total score is calculated accordingly.}
\begin{tabular}{|p{0.5cm}|p{0.7cm}|p{0.9cm}|p{0.9cm}|p{0.9cm}|p{0.9cm}||p{0.7cm}|p{0.8cm}|p{0.9cm}|p{0.9cm}|p{1cm}|}
\hline
\textbf{case} & \textbf{score1} & \textbf{score2} & \textbf{score3} &\textbf{score5} & \textbf{overall} & \textbf{score1} & \textbf{score2} & \textbf{score3} & \textbf{score5} & \textbf{overall} \\ 
\hline
1 & 75 & 50 & 100 & 100 & 75 & 75 & 66.67 & 100 & 100 & 81.67 \\ 
2 & 75 & 83.33 & 100 & 100 & 88.33 & 90 & 83.33 & 100 & 100 & 91.33 \\ 
3 & 75 & 66.67 & 100 & 100 & 81.67 & 50 & 50 & 100 & 100 & 70 \\ 
4 & 75 & 66.67 & 100 & 100 & 81.67 & 75 & 66.67 & 100 & 100 & 81.67 \\ 
5 & 75 & 66.67 & 100 & 100 & 81.67 & 75 & 66.67 & 100 & 100 & 81.67 \\ 
6 & 75 & 66.67 & 100 & 80 & 79.67 & 75 & 66.67 & 100 & 100 & 81.67 \\ 
7 & 75 & 66.67 & 83.33 & 100 & 78.33 & 75 & 66.67 & 83.33 & 100 & 78.33 \\ 
8 & 50 & 66.67 & 100 & 100 & 76.67 & 50 & 50 & 100 & 100 & 70 \\ 
9 & 75 & 83.33 & 100 & 100 & 88.33 & 75 & 66.67 & 100 & 100 & 81.67 \\ 
10 & 75 & 66.67 & 83.33 & 100 & 78.33 & 75 & 66.67 & 100 & 100 & 81.67 \\ 
11 & 50 & 50 & 100 & 100 & 70 & 75 & 66.67 & 100 & 100 & 81.67 \\ 
12 & 65 & 50 & 100 & 100 & 73 & 75 & 83.33 & 100 & 100 & 88.33 \\ 
13 & 50 & 50 & 83.33 & 100 & 66.67 & 100 & 100 & 100 & 100 & 100 \\ 
14 & 75 & 66.67 & 83.33 & 100 & 78.33 & 75 & 66.67 & 100 & 100 & 81.67 \\ 
15 & 50 & 50 & 100 & 100 & 70 & 75 & 66.67 & 83.33 & 100 & 78.33 \\ 
16 & 75 & 66.67 & 100 & 100 & 81.67 & 75 & 83.33 & 100 & 100 & 88.33 \\ 
17 & 75 & 66.67 & 100 & 100 & 81.67 & 75 & 66.67 & 100 & 100 & 81.67 \\ 
18 & 100 & 66.67 & 100 & 100 & 86.67 & 100 & 83.33 & 100 & 100 & 93.33 \\ 
19 & 50 & 66.67 & 100 & 100 & 76.67 & 90 & 90 & 100 & 100 & 94 \\ 
20 & 50 & 66.67 & 100 & 100 & 76.67 & 50 & 66.67 & 100 & 100 & 76.67 \\ 
21 & 90 & 83.33 & 100 & 100 & 91.33 & 90 & 83.33 & 100 & 100 & 91.33 \\ 
22 & 75 & 66.67 & 100 & 100 & 81.67 & 75 & 66.67 & 83.33 & 100 & 78.33 \\ 
23 & 50 & 50 & 100 & 100 & 70 & 75 & 66.67 & 83.33 & 100 & 78.33 \\ 
24 & 90 & 83.33 & 100 & 100 & 91.33 & 90 & 83.33 & 100 & 100 & 91.33 \\ 
25 & 75 & 66.67 & 100 & 100 & 81.67 & 75 & 66.67 & 100 & 100 & 81.67 \\ 
26 & 75 & 83.33 & 100 & 100 & 88.33 & 75 & 83.33 & 100 & 100 & 88.33 \\ 
27 & 50 & 50 & 100 & 100 & 70 & 50 & 66.67 & 100 & 100 & 76.67 \\ 
28 & 75 & 66.67 & 83.33 & 100 & 78.33 & 50 & 50 & 100 & 100 & 70 \\ 
29 & 50 & 66.67 & 100 & 100 & 76.67 & 75 & 66.67 & 100 & 100 & 81.67 \\ 
30 & 75 & 66.67 & 83.33 & 100 & 76.33 & 75 & 66.67 & 83.33 & 100 & 78.33 \\ 
31 & 50 & 50 & 83.33 & 100 & 66.67 & 75 & 66.67 & 100 & 100 & 81.67 \\ 
32 & 75 & 66.67 & 83.33 & 100 & 78.33 & 75 & 66.67 & 100 & 100 & 81.67 \\ 
33 & 50 & 66.67 & 83.33 & 100 & 73.33 & 75 & 66.67 & 83.33 & 100 & 78.33 \\ 
34 & 50 & 66.67 & 100 & 50 & 71.67 & 75 & 83.33 & 100 & 80 & 86.33 \\ 
35 & 75 & 66.67 & 100 & 100 & 81.67 & 75 & 66.67 & 100 & 100 & 81.67 \\ 
36 & 50 & 50 & 100 & 100 & 70 & 75 & 83.33 & 83.33 & 100 & 85 \\ 
37 & 75 & 66.67 & 100 & 100 & 81.67 & 75 & 66.67 & 100 & 100 & 81.67 \\ 
38 & 50 & 66.67 & 83.33 & 100 & 73.33 & 75 & 66.67 & 100 & 100 & 81.67 \\ 
39 & 75 & 66.67 & 100 & 100 & 81.67 & 90 & 83.33 & 100 & 100 & 91.33 \\ 
40 & 75 & 66.67 & 100 & 100 & 81.67 & 90 & 83.33 & 100 & 100 & 91.33 \\ 
41 & 75 & 66.67 & 100 & 100 & 81.67 & 75 & 83.33 & 100 & 100 & 88.33 \\ 
42 & 75 & 66.67 & 100 & 100 & 81.67 & 90 & 83.33 & 100 & 100 & 91.33 \\ 
43 & 50 & 50 & 100 & 100 & 70 & 75 & 66.67 & 100 & 100 & 81.67 \\ 
44 & 75 & 66.67 & 100 & 100 & 81.67 & 75 & 66.67 & 100 & 100 & 81.67 \\
\hline
Ave & 67.5 & 64.77 & 96.21 & 97.95 & 78.45 & 75.68 & 71.74 & 97.35 & 99.55 & 83.26 \\ 
\hline
\end{tabular}
\label{tab:gpt4o_opus_closure}
\end{table*}

\begin{table*}
\centering
\caption{LLM-as-medical-student for Closure: Claude Haiku (left) and Claude Sonnet (right). Score1 is the score of the Diagnostic Impressions, Score2 represents Management Plans, Score3 represents Challenging Questions, Score4 represents Language Accessibility (omitted here as all cases are full scores), and Score5 is for Compassion. Each score has been adjusted to a 100-point scale. The weights for Score1 to Score5 are 0.2, 0.4, 0.2, 0.1, and 0.1 respectively, and the total score is calculated accordingly.}
\begin{tabular}{|p{0.5cm}|p{0.7cm}|p{0.9cm}|p{0.9cm}|p{0.9cm}|p{0.9cm}||p{0.7cm}|p{0.8cm}|p{0.9cm}|p{0.9cm}|p{1cm}|}
\hline
\textbf{case} & \textbf{score1} & \textbf{score2} & \textbf{score3} &\textbf{score5} & \textbf{overall} & \textbf{score1} & \textbf{score2} & \textbf{score3} & \textbf{score5} & \textbf{overall} \\ 
\hline
1 & 75 & 66.67 & 100 & 100 & 81.67 & 75 & 66.67 & 83.33 & 100 & 78.33 \\ 
2 & 75 & 66.67 & 100 & 100 & 81.67 & 75 & 66.67 & 100 & 100 & 81.67 \\ 
3 & 50 & 66.67 & 100 & 100 & 76.67 & 75 & 66.67 & 100 & 100 & 81.67 \\ 
4 & 75 & 66.67 & 100 & 100 & 81.67 & 90 & 83.33 & 100 & 100 & 91.33 \\ 
5 & 75 & 66.67 & 83.33 & 100 & 78.33 & 75 & 66.67 & 100 & 100 & 81.67 \\ 
6 & 75 & 66.67 & 83.33 & 100 & 78.33 & 75 & 66.67 & 100 & 80 & 79.67 \\ 
7 & 75 & 83.33 & 83.33 & 100 & 85 & 75 & 66.67 & 83.33 & 100 & 78.33 \\ 
8 & 50 & 50 & 100 & 100 & 70 & 75 & 83.33 & 100 & 100 & 88.33 \\ 
9 & 75 & 66.67 & 100 & 100 & 81.67 & 75 & 50 & 100 & 100 & 75 \\ 
10 & 75 & 83.33 & 100 & 100 & 88.33 & 75 & 66.67 & 83.33 & 100 & 78.33 \\ 
11 & 75 & 83.33 & 100 & 100 & 88.33 & 50 & 66.67 & 100 & 100 & 76.67 \\ 
12 & 50 & 50 & 100 & 100 & 70 & 50 & 66.67 & 83.33 & 100 & 73.33 \\ 
13 & 0 & 0 & 0 & 100 & 20 & 0 & 0 & 0 & 100 & 20 \\ 
14 & 75 & 66.67 & 83.33 & 100 & 78.33 & 75 & 66.67 & 83.33 & 100 & 78.33 \\ 
15 & 50 & 50 & 83.33 & 100 & 66.67 & 75 & 66.67 & 100 & 100 & 81.67 \\ 
16 & 75 & 66.67 & 83.33 & 100 & 78.33 & 75 & 66.67 & 83.33 & 80 & 76.33 \\ 
17 & 75 & 66.67 & 100 & 100 & 81.67 & 75 & 66.67 & 100 & 100 & 81.67 \\ 
18 & 100 & 83.33 & 100 & 100 & 93.33 & 0 & 0 & 100 & 100 & 40 \\ 
19 & 75 & 83.33 & 83.33 & 100 & 85 & 75 & 66.67 & 83.33 & 100 & 78.33 \\ 
20 & 50 & 50 & 100 & 100 & 70 & 75 & 66.67 & 100 & 100 & 81.67 \\ 
21 & 90 & 83.33 & 100 & 100 & 91.33 & 90 & 83.33 & 93.33 & 100 & 90 \\ 
22 & 75 & 66.67 & 100 & 100 & 81.67 & 75 & 66.67 & 100 & 100 & 81.67 \\ 
23 & 50 & 50 & 83.33 & 100 & 66.67 & 50 & 50 & 100 & 100 & 70 \\ 
24 & 90 & 66.67 & 83.33 & 100 & 81.33 & 90 & 83.33 & 100 & 100 & 91.33 \\ 
25 & 75 & 66.67 & 100 & 100 & 81.67 & 75 & 66.67 & 100 & 100 & 81.67 \\ 
26 & 50 & 66.67 & 100 & 100 & 76.67 & 75 & 50 & 100 & 100 & 75 \\ 
27 & 50 & 50 & 100 & 100 & 70 & 50 & 66.67 & 100 & 100 & 76.67 \\ 
28 & 50 & 50 & 100 & 100 & 70 & 75 & 66.67 & 100 & 100 & 81.67 \\ 
29 & 75 & 66.67 & 100 & 100 & 81.67 & 75 & 66.67 & 100 & 100 & 81.67 \\ 
30 & 75 & 66.67 & 83.33 & 100 & 78.33 & 75 & 66.67 & 83.33 & 100 & 78.33 \\ 
31 & 75 & 66.67 & 100 & 100 & 81.67 & 75 & 66.67 & 83.33 & 100 & 78.33 \\ 
32 & 75 & 66.67 & 83.33 & 100 & 78.33 & 75 & 66.67 & 83.33 & 100 & 78.33 \\ 
33 & 75 & 83.33 & 100 & 100 & 88.33 & 75 & 66.67 & 83.33 & 100 & 78.33 \\ 
34 & 50 & 50 & 100 & 100 & 70 & 75 & 66.67 & 100 & 80 & 79.67 \\ 
35 & 75 & 66.67 & 100 & 100 & 81.67 & 75 & 66.67 & 100 & 100 & 81.67 \\ 
36 & 75 & 83.33 & 83.33 & 100 & 83 & 50 & 66.67 & 83.33 & 100 & 73.33 \\ 
37 & 75 & 50 & 100 & 100 & 75 & 75 & 66.67 & 100 & 100 & 81.67 \\ 
38 & 75 & 66.67 & 83.33 & 80 & 78.33 & 75 & 66.67 & 83.33 & 100 & 78.33 \\ 
39 & 75 & 66.67 & 100 & 100 & 81.67 & 75 & 66.67 & 83.33 & 100 & 78.33 \\ 
40 & 75 & 66.67 & 83.33 & 100 & 78.33 & 75 & 66.67 & 100 & 100 & 81.67 \\ 
41 & 75 & 66.67 & 100 & 100 & 81.67 & 75 & 66.67 & 100 & 100 & 81.67 \\ 
42 & 75 & 66.67 & 100 & 100 & 81.67 & 75 & 66.67 & 100 & 100 & 81.67 \\ 
43 & 50 & 50 & 0 & 100 & 50 & 75 & 66.67 & 100 & 100 & 81.67 \\ 
44 & 75 & 66.67 & 100 & 100 & 81.67 & 75 & 66.67 & 100 & 100 & 81.67 \\ 
\hline
\textbf{Ave} & 68.3 & 64.39 & 90.15 & 99.55 & 77.4 & 69.77 & 64.02 & 92.27 & 98.64 & 77.88 \\ 
\hline
\end{tabular}
\label{tab:haiku_sonnet_closure}
\end{table*}

\begin{table*}
\caption{LLM-as-Medical-Student Output Examples for Closure (Part 1)}
\label{tab:LLM-as-Medical-Student Output Examples for Closure}
    \begin{scriptsize}
    \setlength{\tabcolsep}{4.5pt}
    \renewcommand{\arraystretch}{1.05}
    \centering
    \begin{tabularx}{\textwidth}{lp{13.6cm}}
    \toprule
    \multirow{2}{*}{Ground Truth} & 
    \textbf{Closure:}
    \newline
    ``\emph{Mr. Stern, the palpitations and sweating you have experienced are most likely due to episodes of low blood sugar, which may have resulted from a higher-than-normal dose of insulin or from skipping or delaying meals. The numbness you describe in your feet is probably related to the effect of diabetes on your nervous system. Better control of your blood sugar may help improve this problem. Many factors, including diabetes, can cause the erection difficulties you describe. I will need to perform an examination of your genital area and run some blood tests. At a later point, we may also need to conduct some more complex tests to identify the cause of your problems. Do you have any questions for me?}'' \\
    & \textbf{Question:"Will I lose my feet, doctor?"}
    \newline
    ``\emph{Amputation is a last resort in patients with diabetes who develop an infection in their feet, but fortunately, we are not at that point. The nerve damage to your feet is uncomfortable, but it will not lead to amputation as long as you take the proper measures to protect your feet from injury. If we continue to keep your blood sugar and cholesterol well-controlled, we should be able to manage this condition through medications. We'll discuss more about your diabetes management later in the visit.}'' \\
    \midrule
    \multirow{2}{*}{GPT3.5} & 
    \textbf{Closure:}
    \newline
    ``\emph{Mr. Stern, based on our discussion and physical exam today, we have reviewed your history of diabetes and the medications you are currently taking, including insulin. We have also discussed your blood sugar monitoring and results. Moving forward, we will plan to further investigate your diabetic management with additional blood work and potentially adjusting your insulin regimen. It's important for us to ensure your diabetes is well controlled to prevent any complications. We will also consider a follow-up eye exam to monitor for any signs of diabetic eye disease. Thank you for coming in today, and we will continue to work together to manage your diabetes effectively.}''\\
    & \textbf{Question:"Will I lose my feet, doctor?"}
    \newline
    ``\emph{Mr. Stern, it's important to maintain good control of your diabetes to prevent complications such as diabetic foot problems. Regular monitoring and follow-up appointments will help us address any concerns early on to prevent any issues with your feet. It's great that you are proactive about your health, and we will work together to keep you healthy and prevent any complications. If you have any concerns about your feet, please don't hesitate to reach out to us.}'' \\
    \midrule
    \multirow{2}{*}{GPT4} & 
    \textbf{Closure:}
    \newline
    ``\emph{Mr. Stern, based on our conversation and my examination, you have done a commendable job managing your diabetes over the last 25 years. I understand that you've been dealing with occasional symptoms of low blood sugar and some numbness and tingling in your feet, which can sometimes be complications of long-term diabetes. As it's been a while since your last foot exam, it might be helpful to consult with a podiatrist to check the status of your foot health as it can be impacted by diabetes. Your sexual difficulty may also be related to diabetes and it might be beneficial to consult with a urologist who specializes in sexual health. It's important to monitor your condition closely, so I would like to order blood tests including hemoglobin A1c to assess your blood sugar control over the past few months, lipid profile for your cholesterol levels, and a urine test to check for any kidney issues. I also recommend continuing your regular blood sugar checks at home.}'' \\
    & \textbf{Question:"Will I lose my feet, doctor?"}
    \newline
    ``\emph{No, Mr. Stern. With proper care and management of your diabetes, it's not inevitable that you'd lose your feet. The numbness you're experiencing is a common issue in long-term diabetes patients, and it's important to have regular foot exams to prevent any further complications. We'll work together to ensure you have the best care possible.}'' \\
    \midrule
    \multirow{2}{*}{GPT4o} & 
    \textbf{Closure:}
    \newline
    ``\emph{Mr. Stern, based on our conversation and the physical exam today, it looks like you have been managing your diabetes fairly well. Your blood sugar levels and your last HbA1c are within an acceptable range, although there is always room for improvement. You mentioned experiencing some tingling and numbness in your feet, which can be a sign of diabetic neuropathy, a common complication of diabetes. Additionally, I noticed that you haven't seen a podiatrist recently, which is important for people with diabetes to prevent foot problems. 
    To further assess your condition, we will perform some blood tests, including another HbA1c to get a more current measure of your overall blood sugar control. We will also check your kidney function and cholesterol levels given your history of high cholesterol and heart attack. I recommend scheduling an appointment with a podiatrist for a thorough foot exam and possibly an ophthalmologist to check your eyes if it has been over a year.
    Remember, managing diabetes is a team effort, and with regular monitoring and lifestyle adjustments, we can work together to prevent complications.}'' \\
    & \textbf{Question:"Will I lose my feet, doctor?"}
    \newline
    ``\emph{Mr. Stern, it's understandable to be concerned about the health of your feet. The good news is that by managing your blood sugar levels, having regular foot exams, and taking care of any issues early on, we can significantly reduce the risk of severe complications like losing a foot. It's important to follow the recommendations we discussed, including seeing a podiatrist regularly and keeping an eye on your blood sugar. If we stay on top of these things, we can work together to keep you healthy.}'' \\
    \midrule
    \multirow{2}{*}{Claude Opus} & 
    \textbf{Closure:}
    \newline
    ``\emph{Mr. Stern, based on your history and physical exam today, your diabetes appears to be fairly well-controlled on your current insulin regimen. Your blood sugars at home have been in a good range and your last HbA1c was 7.2\%, which is close to goal. However, you are experiencing some symptoms that may be related to diabetic complications, such as the numbness and tingling in your feet which has worsened recently. This could be due to diabetic neuropathy. I'd like to check your HbA1c and microalbumin levels to assess your current diabetes control and check for any kidney damage. I also recommend seeing a podiatrist and ophthalmologist regularly to monitor for foot problems and retinopathy. We should discuss optimizing your blood sugar control, and I'd like to talk more about your diet, exercise, and heart health as well.}'' \\
    & \textbf{Question:"Will I lose my feet, doctor?"}
    \newline
    ``\emph{Mr. Stern, I understand your concern about losing your feet. Diabetes can lead to serious foot problems if not properly managed. However, with good blood sugar control, regular foot exams, and prompt treatment of any foot issues, most people with diabetes do not end up losing their feet. The numbness and tingling you've noticed is likely due to some nerve damage from the diabetes. I'd like you to see a podiatrist who can examine your feet closely, and it's important you check your feet daily for any cuts, sores or changes in sensation. Let me know right away if you notice any problems. We'll work together to keep your feet healthy. Try not to worry - by taking good care of your diabetes and your feet, you can prevent serious complications.}'' \\
    \bottomrule
    \end{tabularx}
    \end{scriptsize}
\end{table*}

\begin{table*}
\caption{LLM-as-Medical-Student Output Examples for Closure (Part 2)}
\label{tab:LLM-as-Medical-Student Output Examples for Closure 2}
    \begin{scriptsize}
    \setlength{\tabcolsep}{4.5pt}
    \renewcommand{\arraystretch}{1.05}
    \centering
    \begin{tabularx}{\textwidth}{lp{13.6cm}}
    \toprule
    \multirow{2}{*}{Claude Haiku} & 
    \textbf{Closure:}
    \newline
    ``\emph{Based on the history and physical examination, the most likely diagnostic possibilities for Mr. Stern include:
    1. Poorly controlled diabetes mellitus: The patient has had diabetes for 25 years, with a recent HbA1c of 7.2\%, suggestive of suboptimal glycemic control. This could contribute to his reported numbness and tingling in the feet, which may be early signs of diabetic neuropathy.
    2. Cardiovascular disease: The patient has a history of a recent heart attack, as well as high cholesterol, which increases his risk for further cardiovascular complications. This may also be contributing to his reported palpitations.
    To further evaluate Mr. Stern's condition, the planned diagnostic workup includes:
    - Comprehensive metabolic panel and lipid panel to assess overall diabetes and cardiovascular disease control.
    - Nerve conduction studies to evaluate for the presence and severity of diabetic neuropathy.
    - Referral to a cardiologist for further evaluation of his cardiovascular history and risk.
    I will discuss these next steps with Mr. Stern and work collaboratively to optimize his diabetes management and prevent or delay any potential complications.}'' \\
    & \textbf{Question:"Will I lose my feet, doctor?"}
    \newline
    ``\emph{Mr. Stern, I understand your concern about potentially losing your feet due to your diabetes. The good news is that with proper management and monitoring, we can work together to prevent or delay the development of diabetic complications, including neuropathy and foot problems.
    Based on your history and examination, you do seem to be experiencing some early signs of diabetic neuropathy, with the numbness and tingling in your feet. However, this is quite common in patients with long-standing diabetes, and it does not necessarily mean that you will lose your feet.
    The key is to be proactive in managing your diabetes. This includes:
    1. Maintaining tight glycemic control by adhering to your insulin regimen and monitoring your blood sugars regularly. Keeping your HbA1c in the target range can help prevent or slow the progression of neuropathy.
    2. Performing daily foot inspections and practicing good foot care, such as keeping your feet clean, moisturized, and wearing proper footwear. Regular visits with a podiatrist can also help identify and address any foot issues early.
    3. Reporting any changes in sensation, pain, or wound healing in your feet to me promptly so we can address them quickly.}'' \\
    \midrule
    \multirow{2}{*}{Claude Sonnet} & 
    \textbf{Closure:}
    \newline
    ``\emph{Mr. Stern, based on our discussion today, you have had type 2 diabetes for the past 25 years, and you have been managing it with insulin therapy. Your blood sugar levels have been reasonably controlled, with recent readings between 120-145 mg/dL, though your last HbA1c was slightly elevated at 7.2\%. You also have a history of high cholesterol and a heart attack last year, for which you are taking medications.
    During our examination today, I did not find any concerning signs of diabetic eye disease or active foot problems. However, you mentioned experiencing numbness and tingling in your feet, which has worsened over the past 2 months. This could be a sign of diabetic neuropathy, a complication of long-standing diabetes.
    The diagnostic possibilities we need to explore include:
    1. Optimizing your blood sugar control to prevent further progression of neuropathy and other complications.
    2. Assessing for other potential causes of neuropathy, such as vitamin deficiencies or spinal cord compression.
    3. Evaluating for any signs of peripheral arterial disease or foot ulcers that could lead to more severe complications.
    To address these concerns, we will need to order some additional tests, such as:
    - Repeat HbA1c to assess your long-term blood sugar control
    - Blood tests to check for vitamin deficiencies and other potential causes of neuropathy
    - Imaging studies (e.g., MRI or CT scan) of your spine, if indicated
    - Vascular studies (e.g., Doppler ultrasound) to evaluate your blood flow to the feet
    - A referral to a podiatrist or foot specialist for a thorough examination and care recommendations
    We will also discuss strategies to improve your blood sugar control, such as adjusting your insulin doses or adding additional medications. It is essential to maintain good control to prevent further progression of neuropathy and other diabetes-related complications.}'' \\
    & \textbf{Question:"Will I lose my feet, doctor?"}
    \newline
    ``\emph{Your answer: Mr. Stern, I understand your concern about the numbness and tingling in your feet, as these can be early signs of diabetic neuropathy. However, with proper management and care, it is very unlikely that you will lose your feet.
    The numbness and tingling you are experiencing can be a warning sign that we need to work on improving your blood sugar control and preventing further progression of the neuropathy. By following the recommended workup and treatment plan, we can address the underlying causes and potentially reverse or stabilize the neuropathy.
    While severe, untreated diabetic neuropathy can potentially lead to foot ulcers, infections, and in some cases, amputations, this is not an inevitable outcome. By taking proactive steps now, such as:
    1. Optimizing your blood sugar control through adjustments in your insulin regimen or adding additional medications if needed.
    2. Regular follow-up with a podiatrist or foot specialist to monitor for any foot problems and receive appropriate care.
    3. Practicing good foot care, such as inspecting your feet daily, wearing appropriate footwear, and avoiding injuries.
    4. Addressing any underlying vascular issues or poor circulation that could contribute to the problem.
    We can significantly reduce the risk of complications and preserve the health of your feet. With your commitment to managing your diabetes and following the recommended treatment plan, we can work together to prevent further progression and protect your feet.}'' \\
    \bottomrule
    \end{tabularx}
    \end{scriptsize}
\end{table*}

\subsection{Differential Diagnosis}
\subsubsection{Expert Case Study}
The process of expert annotation for the Differential Diagnosis section of the MedQA-CS benchmark presents several challenges that can impact the consistency and reliability of the evaluation. 

One significant issue arises when the MedStuLLM proposes two similar diagnoses with different names. In such cases, determining the appropriate partial credit can be subjective, as experts may have varying perspectives on how closely related the diagnoses are and how much credit should be awarded.

Another challenge emerges when the MedStuLLM provides an additional diagnosis with only a few explanatory sentences, without explicitly listing the historical or physical findings. In these instances, experts may have different interpretations of how to assign points for the findings part, leading to potential inconsistencies in the scoring.

Furthermore, the subjective nature of the overall quality score can introduce variability in the evaluation process. Different experts may have their own distinct criteria for assessing the MedStuLLM's performance, with some experts tending to be more lenient and awarding higher scores, while others may adopt a stricter grading approach, resulting in lower overall quality scores.

\textbf{Example Case:}
\begin{enumerate}
    \item \textbf{Differential Diagnosis Ground Truth}
\begin{itemize}      
\item \textbf{Diagnosis \#1: Humeral fracture}

\textbf{History Finding(s):} 
- Pain following recent fall on outstretched hand.

\textbf{Exam Finding(s): }

- Tenderness over upper and middle right arm, pain increases with hand movement, restricted range of motion.\newline

\item \textbf{Diagnosis \#2: Shoulder dislocation}

\textbf{History Finding(s): }

- Pain following recent fall on outstretched arm

\textbf{Exam Finding(s): }

- Right arm externally rotated and slightly abducted

- Pain and restricted range of motion on shoulder exam\newline

\item \textbf{Diagnosis \#3: Osteoporosis }

\textbf{History Finding(s): }

- Advanced age

- Pain following recent fall on outstretched arm
\end{itemize}

\textbf{Exam Finding(s):}

N/A
    \item \textbf{Additional Diagnoses}
\begin{itemize}
    \item \textbf{Elder abuse:} The history contains red flags (anxious behavior, delayed care-seeking) that may point to elder abuse. The American Medical Association has defined elder abuse as "an act or omission which results in harm or threatened harm to the health or welfare of an elderly person." The diagnosis of elder abuse is not readily made because often both the abuser and the victim deny abuse. Thus, the diagnosis is often inferential, and supporting evidence must be sought.

\item \textbf{Rotator cuff tear:} Patients usually present with nonspecific pain localized to the shoulder, but pain is often referred down the proximal lateral arm due to shared innervation. There may be an inability to abduct or flex the shoulder. Patients may also demonstrate significant weakness in internal or external rotation.
\end{itemize}
        
    \item \textbf{MedStuLLM for Differential Diagnosis}
\begin{itemize}
\item \textbf{Diagnosis \#1: Rotator cuff tear}

\textbf{Historical Finding(s):}

History of trauma

Pain started after a fall

Pain increases with movement

\textbf{Historical reasons: }

Rotator cuff tears can occur due to traumatic events such as falls.

The mechanism of injury (falling on an outstretched arm) is consistent with rotator cuff injury.

Rotator cuff tears typically cause pain with movement of the shoulder.

\textbf{Physical Exam Finding(s): }

Pain on shoulder movement

Restricted range of motion

Tenderness over the shoulder

\textbf{Physical exam data reasons: }

Pain with movement is a common symptom of rotator cuff tears.

Limited range of motion is indicative of a rotator cuff tear or other shoulder pathology.

Localized tenderness can be associated with rotator cuff pathology.\newline

\item \textbf{Diagnosis \#2: Shoulder fracture}

\textbf{Historical Finding(s): }

History of trauma

Pain started after a fall

Use of arm to break the fall

\textbf{Historical reasons: }

Fractures are often caused by trauma, such as a fall.

The onset of pain after a fall is consistent with a fracture.

Using the arm to break a fall can result in fractures of the shoulder, arm, or wrist.

\textbf{Physical Exam Finding(s): }

Non-localized tenderness over the arm and shoulder

Pain and restricted range of motion

Patient holding arm against body

\textbf{Physical exam data reasons:}

Tenderness over a broad area can be indicative of a fracture.

Pain and restricted range of motion are common in fractures.

Protective positioning of the arm is a common response to fracture pain.\newline

\item \textbf{Diagnosis \#3: Shoulder dislocation}

\textbf{Historical Finding(s): }

History of trauma

Pain started after a fall

Immediate pain after the fall

\textbf{Historical reasons: }

Dislocations can occur due to significant trauma or falls.

The onset of pain after a fall is consistent with a dislocation.

Immediate pain is often reported with dislocations.

\textbf{Physical Exam Finding(s): }

Pain on shoulder movement

Restricted range of motion

Patient holding arm in protective posture

\textbf{Physical exam data reasons: }

Pain with movement is typical of a dislocation.

Limited range of motion is common in shoulder dislocations.

Protective posture is often adopted to minimize pain from a dislocation.
\end{itemize}

\item \textbf{Human Annotation}
    \begin{itemize}
    \item \textbf{Expert 1:}

diagnosis 1 name: 5,
historical finding: 1,
physical finding: N/A

diagnosis 2 name: 0,
historical finding: N/A,
physical finding: N/A

diagnosis 3 name: 10,
historical finding: 1,
physical finding: 1

order: 0

total score: 18/49=36.73\%

quality score: 2

\item \textbf{Expert 2:}

diagnosis 1 name: 5,
historical finding: 3,
physical finding: 3

diagnosis 2 name: 0,
historical finding: 0,
physical finding: 0

diagnosis 3 name: 10,
historical finding: 3,
physical finding: 3

order: 0

total score: 27/49=55.10\%

quality score: 2

\item \textbf{Expert 3:}

diagnosis 1 name: 5,
historical finding: 0,
physical finding: N/A

diagnosis 2 name: 10,
historical finding: 1,
physical finding: 3

diagnosis 3 name: 10,
historical finding: 1,
physical finding: N/A

order score:0

total score: 30/49=61.22\%

quality score: 3

    \end{itemize}
\end{enumerate}


\subsubsection{LLM-as-medical-student Case Study}

The case study of evaluating different language models as medical students in the Differential Diagnosis section of the MedQA-CS benchmark reveals both advantages and disadvantages, as shown in Table\ref{tab:LLM-as-Medical-Student Output Examples for Diagnosis}, Table\ref{tab:LLM-as-Medical-Student Output Examples for Diagnosis 2}, Table\ref{tab:LLM-as-Medical-Student Output Examples for Diagnosis 3}, Table\ref{tab:LLM-as-Medical-Student Output Examples for Diagnosis 4}, Table\ref{tab:LLM-as-Medical-Student Output Examples for Diagnosis 5}, and Table\ref{tab:LLM-as-Medical-Student Output Examples for Diagnosis 6}.

One advantage observed with claude3-opus is its ability to generate findings with rich background details, such as "Substernal chest pain that started 40 minutes ago and woke patient from sleep." This level of specificity in capturing the context and timing of symptoms adds credibility and reliability to the findings, an aspect that may be lacking in other models.

On the other hand, gpt-4o demonstrates strength in providing detailed physical exam findings, often including numerical values and comprehensive descriptions without any instances of "N/A" or missing information. This level of thoroughness in reporting examination results can be valuable in the diagnostic process.

However, gpt3.5's performance in this task appears to be less impressive, as it generates relatively short finding explanations that lack a strong focus on medical terminology and domain-specific language. This could potentially lead to ambiguity or misinterpretation in the diagnostic process.

Similarly, gpt4 exhibits a shortcoming in generating an insufficient number of findings, despite being explicitly prompted to do so. Additionally, the model's descriptions of findings tend to be overly broad and lacking in detailed descriptions, which could hinder accurate diagnosis.

\begin{table*}
\caption{LLM-as-Medical-Student Output Examples for Diagnosis (Part 1)}
\label{tab:LLM-as-Medical-Student Output Examples for Diagnosis}
    \begin{scriptsize}
    \setlength{\tabcolsep}{4.5pt}
    \renewcommand{\arraystretch}{1.05}
    \centering
    \begin{tabularx}{\textwidth}{lp{14.1cm}}
    \toprule
    \multirow{2}{*}{Ground Truth} & 
    \textbf{Diagnosis \#1: Myocardial ischemia or infarction}
    
\textbf{History Finding(s):}

-Pressure-like substernal chest pain 

-Pain radiates to left arm, upper back, and neck 

-Pain awakens the patient at night

\textbf{Physical Exam Finding(s):}

-N/A\newline

\textbf{Diagnosis \#2: Cocaine-induced myocardial ischemia}

\textbf{History Finding(s):}

-History of cocaine use 

-Last used yesterday afternoon 

-Pressure-like substernal chest pain 

\textbf{Physical Exam Finding(s):}

-N/A\newline

\textbf{Diagnosis \#3: GERD}

\textbf{History Finding(s):}

-Pain in midchest 

-Previous pain was relieved by antacids 

-Previous pain occurred after heavy meals 

\textbf{Physical Exam Finding(s):}

-N/A\\
    \midrule
    \multirow{2}{*}{GPT4o} & 
    \textbf{Diagnosis \#1: Acute Myocardial Infarction (AMI)}
    
\textbf{Historical Finding(s): }

Substernal chest pain radiating to the left arm, upper back, and neck

Pain woke the patient from sleep and is described as pressure

Associated symptoms of nausea, sweating, and dyspnea

\textbf{Historical reasons: }

The location and radiation of the pain are classic for myocardial infarction.

Pain that wakes a patient from sleep and is described as pressure is concerning for cardiac origin.

Nausea, sweating, and dyspnea are common associated symptoms of myocardial infarction.

\textbf{Physical Exam Finding(s):}

BP 165/85 mm Hg

RR 22/minute

HR 90/minute, regular

\textbf{Physical exam data reasons:}

Elevated blood pressure can be a response to pain or stress, often seen in myocardial infarction.

Increased respiratory rate can indicate distress or pain, which is common in myocardial infarction.

Heart rate within normal limits but regular, which does not rule out myocardial infarction.\newline

\textbf{Diagnosis \#2: Unstable Angina}

\textbf{Historical Finding(s): }

Chest pain precipitated by exertion and heavy meals

Episodes of chest pain over the past 3 months

Pain relieved by antacids in the past

\textbf{Historical reasons:} 

Chest pain triggered by exertion and heavy meals is typical of angina.

Recurrent episodes of chest pain suggest a pattern consistent with unstable angina.

Relief with antacids could indicate a mixed presentation, but the current episode's severity and nature are more concerning for cardiac origin.

\textbf{Physical Exam Finding(s): }

BP 165/85 mm Hg

RR 22/minute

HR 90/minute, regular

\textbf{Physical exam data reasons: }

Elevated blood pressure can be a response to pain or stress, often seen in angina.

Increased respiratory rate can indicate distress or pain, which is common in angina.

Heart rate within normal limits but regular, which does not rule out angina.\newline

\textbf{Diagnosis \#3: Cocaine-Induced Chest Pain}

\textbf{Historical Finding(s): }

History of cocaine use, last used yesterday afternoon

Chest pain radiating to the left arm, upper back, and neck

Associated symptoms of nausea, sweating, and dyspnea

\textbf{Historical reasons: }

Cocaine use is a significant risk factor for chest pain and can cause coronary artery spasm or myocardial infarction.

The location and radiation of the pain are consistent with cardiac-related chest pain.

Nausea, sweating, and dyspnea are common symptoms associated with cocaine-induced chest pain.

\textbf{Physical Exam Finding(s):}

BP 165/85 mm Hg

RR 22/minute

HR 90/minute, regular

\textbf{Physical exam data reasons:} 

Elevated blood pressure can be a response to cocaine use, which can cause vasoconstriction and increased sympathetic activity.

Increased respiratory rate can indicate distress or pain, which is common in cocaine-induced chest pain.

Heart rate within normal limits but regular, which does not rule out cocaine-induced chest pain.\\
    \bottomrule
    \end{tabularx}
    \end{scriptsize}
\end{table*}

\begin{table*}
\caption{LLM-as-Medical-Student Output Examples for Diagnosis (Part 2)}
\label{tab:LLM-as-Medical-Student Output Examples for Diagnosis 2}
    \begin{scriptsize}
    \setlength{\tabcolsep}{4.5pt}
    \renewcommand{\arraystretch}{1.05}
    \centering
    \begin{tabularx}{\textwidth}{lp{14.1cm}}
    \toprule
    \multirow{2}{*}{GPT4} & 
    \textbf{Diagnosis \#1: Acute Coronary Syndrome}
    
\textbf{Historical Finding(s): }

Substernal chest pain radiating to the left arm, upper back, and neck

Associated symptoms of nausea, sweating, and dyspnea

History of hypertension and high cholesterol

\textbf{Historical reasons: }

Radiation of pain to the left arm and neck is typical for cardiac ischemia

Sympathetic activation due to myocardial ischemia can cause nausea, sweating, and shortness of breath

Hypertension and high cholesterol are risk factors for coronary artery disease

\textbf{Physical Exam Finding(s): }

Elevated blood pressure

Tachypnea

N/A

\textbf{Physical exam data reasons: }

Hypertension can be a chronic condition contributing to the development of coronary artery disease

Increased respiratory rate may be a sign of sympathetic activation due to pain or cardiac compromise

N/A\newline

\textbf{Diagnosis \#2: Myocardial Infarction}

\textbf{Historical Finding(s): }

Chest pain of constant severity that woke the patient from sleep

Chest pain episodes in the past associated with exertion

Cocaine use

\textbf{Historical reasons: }

Pain at rest can indicate unstable angina or myocardial infarction

Exertional chest pain suggests stable angina, which can progress to myocardial infarction

Cocaine use can cause coronary artery spasm leading to myocardial infarction

\textbf{Physical Exam Finding(s): }

Elevated blood pressure

Tachypnea

N/A

\textbf{Physical exam data reasons: }

Hypertension can be a chronic condition contributing to the development of coronary artery disease

Increased respiratory rate may be a sign of sympathetic activation due to pain or cardiac compromise

N/A\newline

\textbf{Diagnosis \#3: Cocaine-Induced Chest Pain}

\textbf{Historical Finding(s): }

Recent cocaine use

Chest pain during sexual intercourse

History of similar episodes of chest pain

\textbf{Historical reasons: }

Cocaine can cause coronary vasospasm and mimic symptoms of acute coronary syndrome

Chest pain during physical exertion such as sexual intercourse can be due to cocaine-induced cardiovascular effects

Recurrent chest pain episodes could be related to repeated cocaine use

\textbf{Physical Exam Finding(s): }

Elevated blood pressure

Tachypnea

N/A

\textbf{Physical exam data reasons: }

Cocaine use can cause acute hypertension

Increased respiratory rate may be a sign of sympathetic activation due to cocaine use

N/A \\
        \bottomrule
    \end{tabularx}
    \end{scriptsize}
\end{table*}

\begin{table*}
\caption{LLM-as-Medical-Student Output Examples for Diagnosis (Part 3)}
\label{tab:LLM-as-Medical-Student Output Examples for Diagnosis 3}
    \begin{scriptsize}
    \setlength{\tabcolsep}{4.5pt}
    \renewcommand{\arraystretch}{1.05}
    \centering
    \begin{tabularx}{\textwidth}{lp{14.1cm}}
    \toprule
    \multirow{2}{*}{GPT3.5} & 
    \textbf{Diagnosis \#1: Acute Coronary Syndrome (ACS)}
    
\textbf{Historical Finding(s): }

Substernal chest pain

Radiation of pain to left arm and neck

Nausea, sweating, and dyspnea

\textbf{Historical reasons:}

These symptoms are classic for ACS and indicate a lack of oxygen to the heart muscle

They suggest the possibility of ischemia affecting multiple areas of the heart

\textbf{Physical Exam Finding(s):}

Elevated blood pressure

Severe pain with no tenderness on chest palpation

\textbf{Physical exam data reasons:}

Elevated blood pressure can indicate stress on the cardiovascular system

Severe pain without tenderness suggests deep tissue involvement such as the heart muscle\newline

\textbf{Diagnosis \#2: Gastroesophageal Reflux Disease (GERD)}

\textbf{Historical Finding(s):}

Episodes of chest pain after heavy meals and during sleep

Pain relieved by antacids
History of GERD

\textbf{Historical reasons:}

The timing and relief of pain suggest a gastrointestinal origin

Previous diagnosis of GERD indicates pre-existing esophageal issues

\textbf{Physical Exam Finding(s): }

Clear symmetric breath sounds

Soft, non-tender abdomen with normal bowel sounds

\textbf{Physical exam data reasons: }

Absence of abnormal breath sounds suggests no immediate pulmonary involvement

Normal abdominal exam indicates no acute abdominal pathology\newline

\textbf{Diagnosis \#3: Aortic Dissection}

\textbf{Historical Finding(s): }

Episodes of chest pain with exertional activities

Hypertension for 5 years

Use of cocaine for 10 years

\textbf{Historical reasons: }

Exertional pain can be a sign of aortic pathology

Hypertension and cocaine use are risk factors for aortic dissection

\textbf{Physical Exam Finding(s): }

BP 165/85 mm Hg in both arms

Palpation of peripheral pulses and absence of edema or cyanosis

\textbf{Physical exam data reasons:} 

Discrepancy in blood pressure between arms can indicate possible pathology in the aorta

Normal pulses and absence of edema or cyanosis do not rule out aortic dissection but are consistent with findings in some patients with this condition \\
    \bottomrule
    \end{tabularx}
    \end{scriptsize}
\end{table*}

\begin{table*}
\caption{LLM-as-Medical-Student Output Examples for Diagnosis (Part 4)}
\label{tab:LLM-as-Medical-Student Output Examples for Diagnosis 4}
    \begin{scriptsize}
    \setlength{\tabcolsep}{4.5pt}
    \renewcommand{\arraystretch}{1.05}
    \centering
    \begin{tabularx}{\textwidth}{lp{12.1cm}}
    \toprule
    \multirow{2}{*}{Opus} & 
\textbf{Diagnosis \#1: Acute coronary syndrome (ACS)}

\textbf{Historical Finding(s): }

Substernal chest pain that started 40 minutes ago and woke patient from sleep

Pain radiates to left arm, upper back and neck

Associated nausea, sweating and shortness of breath

\textbf{Historical reasons: }

Sudden onset of severe chest pain at rest is concerning for ACS

Radiation of pain to left arm, upper back and neck is classic for cardiac ischemia

Nausea, diaphoresis and dyspnea often accompany ACS

\textbf{Physical Exam Finding(s): }

Normal heart sounds without murmurs, rubs or gallops

Symmetric pulses and lack of peripheral edema argues against aortic dissection or severe heart failure

\textbf{Physical exam data reasons: }

A normal cardiac exam in the setting of chest pain does not rule out ACS

Equal pulses and lack of edema make other life-threatening causes like aortic dissection or acute heart failure less likely
\newline

\textbf{Diagnosis \#2: Cocaine-induced chest pain}

\textbf{Historical Finding(s): }

Patient admits to cocaine use, most recently yesterday afternoon

10 year history of cocaine abuse

Pain started at rest and is constant in nature

\textbf{Historical reasons: }

Cocaine is a well known cause of chest pain and myocardial ischemia/infarction

Chronic cocaine use increases risk

Cocaine-induced chest pain and ischemia can occur at rest and tends to be prolonged

\textbf{Physical Exam Finding(s): }

Tachycardia and hypertension

Normal cardiac exam otherwise

\textbf{Physical exam data reasons: }

Cocaine's sympathomimetic effects cause increased HR and BP

Lack of overt signs of heart failure argues against severe cardiac dysfunction
\newline

\textbf{Diagnosis \#3: Gastroesophageal reflux disease (GERD)}

\textbf{Historical Finding(s): }

Past medical history of GERD

Previous episodes of chest pain relieved by antacids

Chest pain in the past precipitated by large meals

\textbf{Historical reasons: }

Known history of GERD makes this a more likely cause of chest pain

Response to antacids suggests an esophageal source of pain

Large meals can precipitate reflux and associated chest pain

\textbf{Physical Exam Finding(s): }

N/A 

\textbf{Physical exam data reasons: }

N/A \\
    \bottomrule
    \end{tabularx}
    \end{scriptsize}
\end{table*}

\begin{table*}
\caption{LLM-as-Medical-Student Output Examples for Diagnosis (Part 5)}
\label{tab:LLM-as-Medical-Student Output Examples for Diagnosis 5}
    \begin{scriptsize}
    \setlength{\tabcolsep}{4.5pt}
    \renewcommand{\arraystretch}{1.05}
    \centering
    \begin{tabularx}{\textwidth}{lp{14.1cm}}
    \toprule
    \multirow{2}{*}{Sonnet} & 
\textbf{Diagnosis \#1: Acute Coronary Syndrome (ACS)/Myocardial Infarction (MI)}

\textbf{Historical Finding(s): }

Chest pain with radiation to left arm, neck, and upper back

Chest pain associated with dyspnea, nausea, and diaphoresis

History of hypertension, hyperlipidemia, and cocaine use

\textbf{Historical reasons: }

These symptoms are classic for cardiac ischemia/myocardial infarction

Risk factors for coronary artery disease increase the likelihood of ACS/MI

\textbf{Physical Exam Finding(s): }

Elevated blood pressure

N/A

N/A

\textbf{Physical exam data reasons: }

Elevated blood pressure is common in ACS/MI due to sympathetic activation
\newline

\textbf{Diagnosis \#2: Gastroesophageal Reflux Disease (GERD)/Esophageal Spasm}

\textbf{Historical Finding(s): }

Chest pain with radiation to upper back and neck

History of GERD and relief with antacids in the past

Chest pain associated with heavy meals

\textbf{Historical reasons: }

Radiation of pain to upper back and neck can be seen in GERD/esophageal spasm

History of GERD and relief with antacids suggests a possible recurrence

Heavy meals can exacerbate GERD and trigger esophageal spasms

\textbf{Physical Exam Finding(s): }

N/A

N/A

N/A

\textbf{Physical exam data reasons: }
\newline

\textbf{Diagnosis \#3: Aortic Dissection}

\textbf{Historical Finding(s): }

Chest pain with radiation to upper back
History of hypertension

Sudden onset of severe chest pain

\textbf{Historical reasons: }

Aortic dissection can present with chest pain radiating to the back

Hypertension is a risk factor for aortic dissection

Aortic dissection often presents with sudden, severe chest pain

\textbf{Physical Exam Finding(s): }

Elevated blood pressure

N/A

N/A

\textbf{Physical exam data reasons: }

Elevated blood pressure can be seen in aortic dissection due to hypertensive emergency\\
    \bottomrule
    \end{tabularx}
    \end{scriptsize}
\end{table*}

\begin{table*}
\caption{LLM-as-Medical-Student Output Examples for Diagnosis (Part 6)}
\label{tab:LLM-as-Medical-Student Output Examples for Diagnosis 6}
    \begin{scriptsize}
    \setlength{\tabcolsep}{4.5pt}
    \renewcommand{\arraystretch}{1.05}
    \centering
    \begin{tabularx}{\textwidth}{lp{14.1cm}}
    \toprule
    \multirow{2}{*}{Haiku} & 
\textbf{"Diagnosis \#1: Acute Coronary Syndrome (Myocardial Infarction)}

\textbf{Historical Finding(s): }

Chest pain with radiation to left arm, neck, and upper back

Pain onset at rest (woke up from sleep)

Diaphoresis and nausea

\textbf{Historical reasons: }

Chest pain with radiation is a classic symptom of myocardial infarction

Onset of symptoms at rest is suggestive of acute coronary syndrome

Diaphoresis and nausea are common associated symptoms of myocardial infarction

\textbf{Physical Exam Finding(s): }

Elevated blood pressure

Regular heart rate

\textbf{Physical exam data reasons:}

Increased blood pressure can be a sign of sympathetic activation in acute coronary syndrome

Regular heart rate does not rule out myocardial infarction
\newline

\textbf{Diagnosis \#2: Unstable Angina}

\textbf{Historical Finding(s): }

Recurrent chest pain episodes over the past 3 months

Chest pain precipitated by exertion, heavy meals, and sexual activity

Partial relief of symptoms with antacids

\textbf{Historical reasons: }

Recurrent anginal episodes over time are suggestive of unstable angina

Chest pain triggered by exertion and relieved by rest is typical of angina

Partial relief of symptoms with antacids suggests an element of GERD may be present

\textbf{Physical Exam Finding(s): }

N/A

\textbf{Physical exam data reasons: }

The physical exam findings do not provide specific support for the diagnosis of unstable angina
\newline

\textbf{Diagnosis \#3: Gastroesophageal Reflux Disease (GERD) with Atypical Presentation}

\textbf{Historical Finding(s): }

Chest pain relieved by antacids in the past

History of GERD for 10 years

Cocaine use

\textbf{Historical reasons: }

Relief of chest pain with antacids suggests an element of GERD

Prior history of GERD increases the likelihood of GERD as a contributor to the current symptoms

Cocaine use can cause atypical chest pain presentations

\textbf{Physical Exam Finding(s):}

N/A

\textbf{Physical exam data reasons: }

The physical exam findings do not provide specific support for the diagnosis of GERD with atypical presentation\\
    \bottomrule
    \end{tabularx}
    \end{scriptsize}
\end{table*}

\end{document}